\documentclass[journal]{IEEEtran}
\usepackage{graphicx}
\usepackage{bm,hyperref,xcolor,enumitem,booktabs,arydshln,footnote,url}
\hypersetup{hypertex=true,
	colorlinks=true,
	linkcolor=green,
	anchorcolor=blue,
	citecolor=green}
\usepackage{array}
\usepackage[caption=false,font=normalsize,labelfont=sf,textfont=sf]{subfig}
\usepackage{textcomp}
\usepackage{url}
\usepackage{cite}
\usepackage{setspace}
\usepackage{epstopdf}
\usepackage[labelsep=period]{caption}  
\usepackage[british]{babel} 
\usepackage{listings}
\usepackage{bbding}

\usepackage{amssymb}
\usepackage{amsmath}
\usepackage{rotating} 
\usepackage{caption}
\usepackage{chen}

\definecolor{gold}{RGB}{255,215,0}  
\definecolor{dkgreen}{rgb}{0,0.6,0}
\definecolor{gray}{rgb}{0.5,0.5,0.5}
\definecolor{mauve}{rgb}{0.58,0,0.82}

\DeclareCaptionFormat{mylst}{\hrule#1#2#3}
\ifCLASSINFOpdf
\else
\fi

\begin{document}
%
\title{SegAssess: Panoramic quality mapping for robust and transferable unsupervised segmentation assessment}
%
%
%


\author{\IEEEauthorblockN{Bingnan Yang,
		Mi Zhang$^{\ast}$, 
		Zhili Zhang, Zhan Zhang, \\
		Yuanxin Zhao, Xiangyun Hu, and
		Jianya Gong}\\
	\thanks{*Corresponding author: Mi Zhang}
	\thanks{All authors except Zahili Zhang are with School of Remote Sensing and Information Engineering, Wuhan University, Wuhan 430079, China. (e-mail: \href{mailto:bingnan.yang@whu.edu.cn}{bingnan.yang@whu.edu.cn}; \href{mailto:mizhang@whu.edu.cn}{mizhang@whu.edu.cn};  \href{mailto:zhangzhili@whu.edu.cn}{zhangzhili@whu.edu.cn};
	\href{mailto:zhangzhanstep@whu.edu.cn}{zhangzhanstep@whu.edu.cn};
	\href{mailto:yuanxin.zhao@whu.edu.cn}{yuanxin.zhao@whu.edu.cn};
	\href{mailto:huxy@whu.edu.cn}{huxy@whu.edu.cn}; \href{mailto:gongjy@whu.edu.cn}{gongjy@whu.edu.cn})}
	\thanks{Zhili Zhang is with College of Electronic Science and Technology, National University of Defense Technology, Changsha, 410073, China.}
	\thanks{Zhan Zhang and Jianya Gong are also with State Key Laboratory of Information Enginnering in Surveying, Mapping and Remote Sensing, Wuhan University, Wuhan 430079, China.}
	\thanks{Mi Zhang, Xiangyun Hu and Jianya Gong are also with Hubei Luojia Laboratory, Wuhan 430079, China.}
}
\maketitle

\begin{abstract}
High-quality image segmentation is fundamental to pixel-level geospatial analysis in remote sensing, necessitating robust segmentation quality assessment (SQA), particularly in unsupervised settings lacking ground truth. Although recent deep learning (DL) based unsupervised SQA methods show potential, they often suffer from coarse evaluation granularity, incomplete assessments, and poor transferability. To overcome these limitations, this paper introduces Panoramic Quality Mapping (PQM) as a new paradigm for comprehensive, pixel-wise SQA, and presents SegAssess, a novel deep learning framework realizing this approach. SegAssess distinctively formulates SQA as a fine-grained, four-class panoramic segmentation task, classifying pixels within a segmentation mask under evaluation into true positive (TP), false positive (FP), true negative (TN), and false negative (FN) categories, thereby generating a complete quality map. Leveraging an enhanced Segment Anything Model (SAM) architecture, SegAssess uniquely employs the input mask as a prompt for effective feature integration via cross-attention. Key innovations include an Edge Guided Compaction (EGC) branch with an Aggregated Semantic Filter (ASF) module to refine predictions near challenging object edges, and an Augmented Mixup Sampling (AMS) training strategy integrating multi-source masks to significantly boost cross-domain robustness and zero-shot transferability. Comprehensive experiments demonstrate that SegAssess achieves state-of-the-art (SOTA) performance and exhibits remarkable zero-shot transferability to unseen masks. The code is available at \href{https://github.com/Yangbn97/SegAssess}{https://github.com/Yangbn97/SegAssess}
\end{abstract}


\section{Introduction}\label{intro}
The proliferation of high-resolution remote sensing imagery has revolutionized geospatial analysis, enabling unprecedented spatial detail for Earth observation\cite{gao2017novel,chen2018review}. Image segmentation (IS), which partitions pixels into semantically meaningful regions, serves as the foundational technique\cite{hay2008geographic, blaschke2014geographic, chen2018geographic}. As the basis of pixel-level geospatial analysis, the segmentation quality remarkably impacts the reliability and robustness of the entire workflow\cite{gao2017novel,chen2018review}. This critical dependency underscores the urgent demand for robust segmentation quality assessment (SQA) frameworks, especially in unsupervised settings where ground truth is unavailable.

\begin{figure*}[t]
	\centering{
		\includegraphics[scale=0.5]{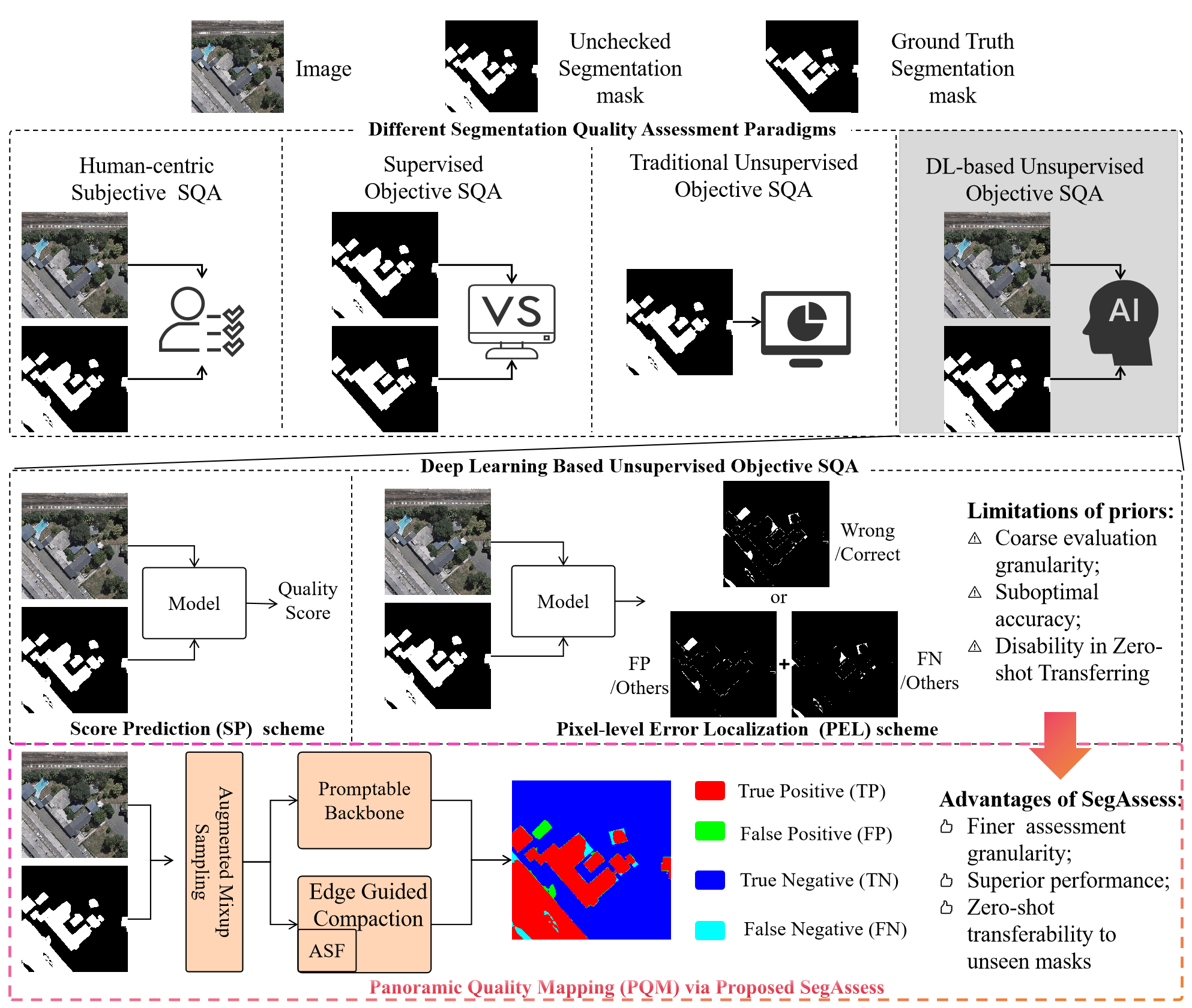}}
	\caption{Schematic diagrams of different segmentation quality assessment (SQA) paradigms (second row) and visual comparison between proposed Panoramic Quality Mapping scheme via SegAssess model (fourth row) and other deep learning based unsupervised SQA methods (third row). ASF: Aggregated Semantic Filter.
	}
	\label{Fig:SQA_methods}
	\vspace{-1.8em}
\end{figure*} 

As shown in Fig.~\ref{Fig:SQA_methods}, existing SQA methods broadly fall into two categories: human-centric subjective evaluation and algorithm-based objective evaluation. Human-centric assessment\cite{meinel2004comparison,chen2019visual}, while theoretically nuanced, is often economically impractical for large-scale applications. Supervised objective methods\cite{carleer2005assessment,clinton2010accuracy,yang2015discrepancy,costa2018supervised,cai2022adaptive,tetteh2023comparison} provide standardized evaluations by comparing outputs against ground truth (GT) references,  but their fundamental reliance on GT data restricts their applicability in many real-world scenarios. Traditional unsupervised approaches\cite{espindola2006parameter,ming2009evaluation,zhang2011scale} eliminate this dependency by analyzing intrinsic data patterns, but often struggle with transferability to new tasks or datasets.

With the emergence of deep learning, unsupervised SQA has advanced significantly, primarily following two paradigms: score prediction (SP)\cite{zhou2019robust,uslu2024robust,shi2025remote} and Pixel-level Error Localization (PEL)\cite{zhang2023automatic,zaman2023segmentation}. SP networks process images and segmentation masks to predict a predefined quality score in a regression manner (Fig.~\ref{Fig:SQA_methods}). Though effective in certain contexts, these approaches assess quality at a coarse image or object level, lacking the pixel-wise diagnostic capabilities needed to localize specific defects. 

\begin{figure*}[t]
	\centering{
		\includegraphics[scale=0.5]{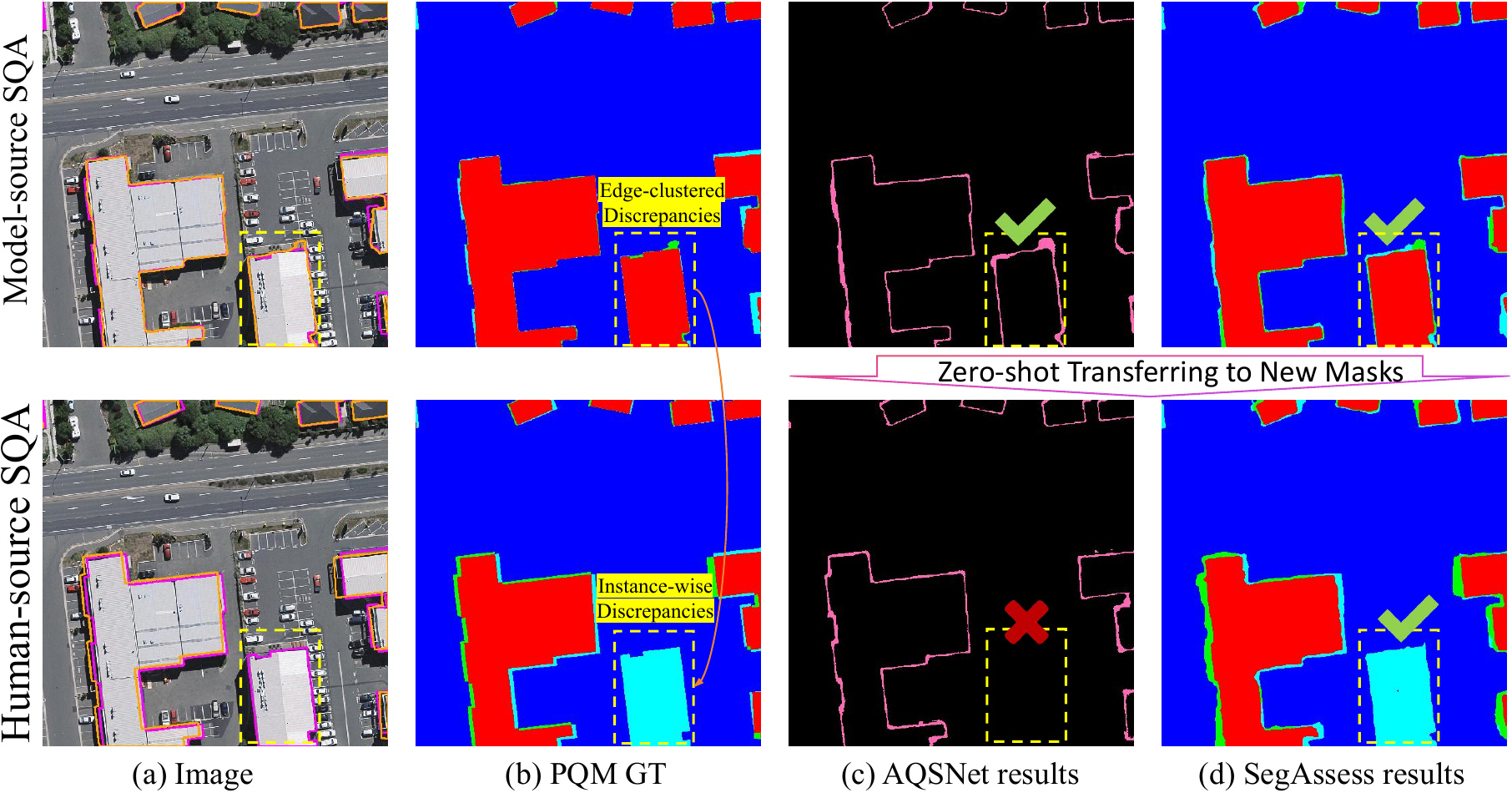}}
	\caption{Visual comparison between SegAssess (PQM approach) and prior work (AQSNet). SegAssess and AQSNet were firstly trained with model generated masks (first row), and then zero-shot applied to unseen masks from historial human annotations (second row). (a) Input image with mask under evaluation (\textcolor{orange}{Orange} contour) and ground truth  (\textcolor{magenta}{magenta} contour). (b) Ground truth PQM assessment map (4-class). (c) AQSNet result (FP/FN errors only). (d) SegAssess PQM result (4-class). Colors represent \textcolor{red}{TP}, \textcolor{green}{FP}, \textcolor{blue}{TN} and \textcolor{cyan}{FN}. \textcolor{pink}{Pink} overlapping FP/FN predictions by AQSNet, highlighting spatial ambiguity not present in the SegAssess PQM output. \textcolor{gold}{Golden} highlights AQSNet's failure in zero-shot transferring from model-source masks with edge-clustered discrepancies between unchecked and GT masks to unseen human-source masks with instance-wise discrepancies.
	}
	\label{Fig:intro}
	\vspace{-1.8em}
\end{figure*} 
To address the granularity limitations of SP models, recent research has explored PEL networks\cite{zhang2023automatic,zaman2023segmentation}. These methods attempt to obtain pixel-wise, objective assessment results by directly pinpointing per-pixel inaccuracies within the segmentation mask (Fig.~\ref{Fig:SQA_methods}). For example, one prior work\cite{zaman2023segmentation} employed "correct/incorrect" binary segmentation to simply identify erroneous pixels. Meanwhile, AQSNet\cite{zhang2023automatic} was designed to generate separate masks identifying false positive (FP, wrong foreground) and false negative (FN, missing foreground) pixels within the mask under evaluation. However, these PEL approaches, despite offering finer spatial detail, introduce their own set of limitations. Binary error classification (e.g., "correct/incorrect")  lacks the necessary nuance to distinguish between different types of mistakes. Even methods like AQSNet that differentiate FP and FN errors face significant drawbacks. As illustrated by our preliminary experiments(Fig~\ref{Fig:intro}), AQSNet provides an incomplete assessment by overlooking correct categories, i.e.true positive (TP) and true negative (TN). Furthermore, its strategy of predicting FP and FN independently can lead to spatially overlapping error regions (pink areas in Fig.~\ref{Fig:intro}), suggesting potential architectural weaknesses. Moreover, AQSNet's attempt at zero-shot transferring application clearly reveals its disability in tackling unseen segmentation masks, particularly when the error characteristics differ from the training data (dashed rectangles in Fig.~\ref{Fig:intro}). 

The limitations observed in existing DL-based SQA methods are fundamentally rooted in common data characteristics and inherent conceptual or architectural deficiencies. From a data perspective, erroneous pixels are often concentrated near complex object edges, and the four quality categories (TP, FP, TN, FN) are typically highly imbalanced, hindering a panoramic assessment (see Section~\ref{sec:data_stats}). Architecturally, prior paradigms exhibit flaws. Score prediction inherently lacks spatial detail, while the theoretical basis of error localization methods is problematic for comprehensive assessment. A complete quality evaluation arguably requires considering all four pixel categories, as they provide mutual contextual priors that can aid identification (see Section~\ref{train_val1}).

These fundamental challenges motivate the exploration of a third, distinct paradigm: Panoramic Quality Mapping (PQM). We define PQM as the reformulation of unsupervised SQA into a comprehensive, pixel-wise classification task, wherein every pixel of the mask under evaluation is classified into one of four fundamental quality categories—true positive (TP), false positive (FP), true negative (TN), or false negative (FN)—thereby generating a complete quality map. This paradigm aims for a holistic assessment that resolves the issues of incomplete categorization and spatial ambiguity found in prior methods.

To realize this PQM vision and alleviate the interconnected issues of prior approaches, we propose SegAssess—a novel deep learning framework for pixel-level unsupervised segmentation quality assessment designed to achieve finer categorization, improved accuracy, and robust transferability. SegAssess embodies the PQM paradigm by explicitly formulating the SQA task as a four-class panoramic segmentation problem. It is built upon an High-quality Segment Anything Model\cite{ke2023segment} (HQ-SAM) architecture, uniquely treating the input segmentation mask as a prompt to facilitate effective feature integration via cross-attention. Key innovations include an Edge Guided Compaction (EGC) branch containing an Aggregated Semantic Filter (ASF) module to specifically tackle challenges related to edge complexity and class imbalance. Furthermore, to directly combat poor cross-domain transferability, an Augmented Mixup Sampling (AMS) strategy is employed during training. 

To sum up, we provide contributions as follows:

\begin{itemize}
	\item [(1)] 
	We introduce SegAssess, a novel framework for pixel-level, unsupervised segmentation quality assessment based on the Panoramic Quality Mapping (PQM) paradigm. Compared to prior methods, SegAssess offers finer assessment granularity (TP/FP/TN/FN), superior performance, and significantly enhanced transferability.
	\item [(2)]      
	We design an Edge Guided Compaction (EGC) branch, incorporating a novel Aggregated Semantic Filter (ASF) module. These components specifically target complex edge regions, improving the model's ability to differentiate challenging classes more effectively.
	\item [(3)]
	We propose a dual strategy to achieve robust generalization and transferability. SegAssess leverages the domain-agnostic visual representations of a pre-trained SAM via an optimized prompt-based mechanism and is complemented by an Augmented Mixup Sampling (AMS) strategy during training. 
\end{itemize}

\section{Related work}\label{related_work}
As illustrated in Fig.~\ref{Fig:SQA_methods}, existing segmentation quality assessment (SQA) methods can be broadly classified into human-centric subjective evaluations and algorithm-based objective evaluations.

\subsection{Human-centric subjective segmentation quality assessment}
Human-centric subjective evaluation, often considered the benchmark, relies on visual inspection and qualitative judgment by domain experts, aligning assessment directly with human-defined quality criteria\cite{meinel2004comparison,chen2019visual}. This approach involves collaborative assessment by multiple domain experts who qualitatively evaluate full or sampled segmentation outputs based on their professional experience. Despite its superior credibility compared to automated methods, manual SQA faces inherent challenges stemming from evaluator expertise variations, cognitive biases, and inter-rater discrepancies\cite{gao2017novel, zhang2015segmentation, chen2018review}. Mitigating these limitations requires rigorous assessor training programs and standardized evaluation protocols.

However, the establishment of professional assessment teams and implementation of labour-intensive evaluation processes present significant scalability barriers, particularly for large-scale datasets. Several recent empirical studies\cite{zhang2025efficiently,radsch2024quality} on annotation workflow optimization have revealed critical insights: (1) quality control procedures exhibit diminishing returns beyond certain annotation volumes, and (2) there exists an optimal balance point between assessment granularity and resource allocation efficiency. These inherent limitations of manual SQA highlight the critical need for developing automated assessment frameworks that balance human-like evaluation accuracy with computational efficiency and large-scale applicability.

\subsection{Supervised objective segmentation quality assessment}
Supervised objective SQA paradigm\cite{carleer2005assessment,clinton2010accuracy,yang2015discrepancy,costa2018supervised,cai2022adaptive,tetteh2023comparison} evaluates segmentation quality by quantifying the discrepancy between segmentation results and corresponding ground truth (GT) reference data, typically derived from high-precision manual annotations. It comprises three sequential components, namely reference dataset construction, object correspondence  matching and discrepancy quantification. The reference datasets are mainly produced through aforementioned human-centric segmentation approaches for the sake of accuracy assurance. The object correspondence phase addresses the critical challenge of establishing optimal pairings between generated segments and reference annotations. While an ideal one-to-one correspondence is theoretically desirable, practical scenarios frequently exhibit complex many-to-one or one-to-many relationships. This has motivated the development of specialized matching strategies such as Maximum Overlapping Area Method\cite{zhao2015segmentation}, Object-Fate Matching Method\cite{schopfer2006object} and Two-Sided 50$\%$ Method\cite{yang2015discrepancy}. 

After obtaining paired generated and reference segments, discrepancy quantification is conducted to measure their differences via diverse representative metrics. In the conventional remote sensing SQA domain, the geometric relationship between generated and reference segments are primarily categorized as three basic types, namely overlapping segmentation, under-segmentation and over-segmentation\cite{bowyer2000validation,blaschke2008object,marpu2010enhanced,liu2012discrepancy,chen2018review}. Area-based methods\cite{clinton2010accuracy,zhang2015segmentation}, such as Fraction of Correctly Segmented Pixels (FCSP) index\cite{fram1975quantitative,chen2006segmentation}, Area Fit Index and SimSize Index\cite{zhan2005quality}, quantify these geometric discrepancies through single or composite relationship assessments. Alternatively, discrepancy quantification can also be implemented based on differences in object locations\cite{zhan2005quality,moller2007comparison}, boundaries\cite{yu2010optimal}, segment numbers\cite{liu2012discrepancy} between generated and reference data. 

In the deep learning field, quality assessment typically employs confusion matrix analysis at pixel level based on generated and GT segmentation masks, and then yield true positive (TP), false positive (FP), true negative (TN) and false negative (FN) counts. These fundamental counts are then used to compute standard metrics including Precision, Recall, F1-score, and Intersection over Union (IoU) for more holistic performance evaluation. Beyond that, some scholars also implement SQA by focusing on boundaries\cite{cheng2021boundary}, directly validating topological correctness\cite{van2018spacenet, biagioni2012inferring} or selectively combining various metrics\cite{xu2023rngdet++,yang2023topdig} to enhance evaluation rigour. 

Despite providing objective and automated quality measures, the fundamental reliance of supervised SQA methods on GT data severely restricts their applicability in real-world scenarios where reference annotations are often unavailable or prohibitively expensive to create. Furthermore, the optimal choice of metrics can be dataset-dependent, often requiring task-specific customization and limiting the generalizability of these approaches. These limitations necessitate unsupervised SQA methods.   

\subsection{Unsupervised objective segmentation quality assessment}    
Unsupervised objective SQA methods eliminate the dependency on reference data, evaluating segmentation quality based on intrinsic properties of the segments and their relationship with the image content. 
\subsubsection{Traditional methods}
Traditional approaches\cite{weszka1978threshold,levine1985dynamic,woodcock1987factor,zhang1996survey,espindola2006parameter,ming2009evaluation,zhang2011scale} quantify segmentation quality through expert-designed metrics that analyse intra-segment homogeneity and inter-segment heterogeneity. These goodness indexes are calculated based on texture complexity\cite{weszka1978threshold,levine1985dynamic,lin2024no}, spectral difference at the level of local pixels\cite{weszka1978threshold} or spectral variance at segment level\cite{woodcock1987factor,liu1994multiresolution,zhang1996survey,druaguct2014automated}. In order to evaluate intra-segmentation homogeneity, some researchers\cite{chen2004use} introduced Intra-Region Visual Error that calculates the value difference between each pixel and the corresponding segment average. The approach in \cite{liu1994multiresolution} evaluated the variances of pixel spectral values in each segment. As to the inter-segment heterogeneity, $\Delta$GL index\cite{zhang2011scale} compares the mean spectral value of a segment to that of all its adjacent segments. Variance Contrast Across Region Measure\cite{wang2017high} quantifies discrepancy of grey value variances between each segment pair. While useful, these methods often involve subjective metric design choices and may lack robustness across diverse datasets.

\subsubsection{Deep learning based methods}
The advent of deep learning has spurred innovations in unsupervised SQA frameworks\cite{zhou2019robust,zhang2023automatic,zaman2023segmentation,uslu2024robust,shi2025remote}. These typically involve training a model (often supervised by GT during development) to assess segmentation quality using the image and the segmentation mask as inputs, enabling reference-free assessment during deployment. Current methodologies fall into two categories, namely score prediction (SP) networks\cite{zhou2019robust,uslu2024robust,shi2025remote} and pixel-wise error localization (PEL) networks\cite{zhang2023automatic,zaman2023segmentation}. The former paradigm aims at mapping input images and segmentation masks to quality scores. For instance, the method proposed in \cite{uslu2024robust} rotated an input medical image with four angles and then design four parallel decoders to generate corresponding segmentation masks, which are utilized to collaboratively predict the final quality score through a regression model. RS-SQA model\cite{shi2025remote} predicts the quality score of input segmentation masks based on a vision language model (VLM). While these SP models enable reference-free and automatic execution, they face two critical limitations. on the one hand, most of them adopt quantitative metrics that defined according to expert perception, thus there are still implicit uncertainty caused by the subjective design bias. On the other hand, score based assessment is implemented at image or object levels, restricting the analysis granularity. Alternatively, some studies pioneered the PEL paradigm that is introduced to address the granularity limitations of SP methods. A previous work\cite{zaman2023segmentation} employed a U-net based network to process concatenated medical images and segmentation masks to identify wrong pixels. AQSNet\cite{zhang2023automatic} adopts a HRNet based architecture to process remote sensing images and segmentation masks to segment wrong and missing pixels. Though making processing, these PEL methods still suffer from significant drawbacks: they provide an incomplete assessment by typically ignoring correctly classified pixels (TP, TN), can exhibit spatial ambiguity (e.g., overlapping FP/FN predictions), and often demonstrate inadequate performance and poor transferability, especially to unseen mask sources. 

The persistent limitations inherent in both the score prediction and pixel-level error localization paradigms, ranging from coarse granularity and subjective biases in SP, to incomplete assessment and poor generalization in PEL, underscore the need for a fundamentally different approach. This motivates the Panoramic Quality Mapping (PQM) paradigm proposed in this work, implemented via the SegAssess framework, which aims to provide a comprehensive, robust, and transferable pixel-level assessment covering all quality categories (TP, FP, TN, FN).

\section{Methodology}\label{method}
\begin{figure*}
	\centering
	\includegraphics[width=1\linewidth]{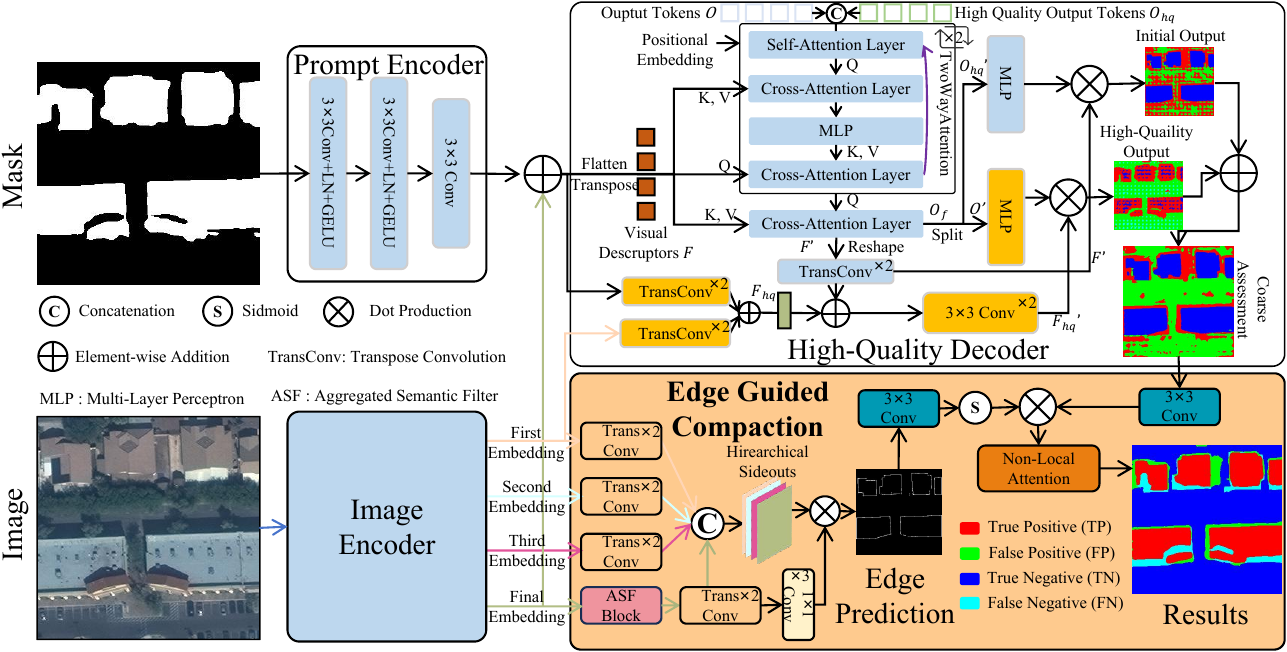}
	\caption{Overview of the SegAssess framework architecture for Panoramic Quality Mapping. SegAssess employs a promptable HQ-SAM backbone for coarse PQM assessment, followed by an Edge Guided Compaction (EGC) branch (orange box) incorporating an Aggregated Semantic Filter (ASF) module (pink rectangle) for result refinement.}
	\label{Fig:overview}
\end{figure*}

SegAssess implements unsupervised, pixel-wise segmentation quality assessment following the Panoramic Quality Mapping (PQM) paradigm. It classifies each pixel in the input segmentation mask (the mask under evaluation) into one of four categories: true positive (TP), false positive (FP), true negative (TN), or false negative (FN), generating a comprehensive quality map. As illustrated in Fig~\ref{Fig:overview}, the SegAssess framework is built upon an adjusted HQ-SAM\cite{ke2023segment} architecture, chosen for its strong foundation model capabilities and promptable nature, enabling effective integration of image and mask information (Section \ref{sec:backbone}). To achieve fine-grained PQM results and robust transferability, SegAssess incorporates two key innovations beyond the backbone: an Edge Guided Compaction (EGC) branch (Section \ref{sec:EGC})with an Aggregated Semantic Filter (ASF) module (Section \ref{sec:A3}) for refined predictions, particularly near boundaries, and an Augmented Mixup Sampling (AMS) strategy during training to enhance generalization across diverse mask sources (Section \ref{sec:AMS}).

\subsection{Promptable HQ-SAM backbone} \label{sec:backbone} 
A core challenge in designing models for SQA is effectively integrating the image $I^{3 \times H \times W}$ and the segmentation mask under evaluation $S^{H \times W}$ , where $H$ and $W$ are height and width. Prior methods often resort to simple channel-wise concatenation, which can hinder feature interpretation. To solve this problem, SegAssess employs a promptable backbone adapted from HQ-SAM\cite{ke2023segment}. Our key adaptation is not a re-design of the backbone's internal layers, but its novel application to the SQA task, where we uniquely treat the input mask as a dense prompt, enabling sophisticated feature integration with the image representation via cross-attention mechanisms. This approach is motivated by the powerful multi-modal fusion capabilities inherent in large foundation models like SAM. By leveraging the model's pre-trained weights, we treat the image and the mask as two distinct modalities and capitalize on the backbone's ability to fuse them effectively via cross-attention. This strategy has been shown to be more robust and effective than adapting a smaller, single-modality network to multi-model domain, allowing for a more nuanced integration of information for the final assessment. The backbone consists of an image encoder, a prompt encoder, and a high-quality decoder.

\textbf{Image encoder} follows the standard Vision Transformer (ViT) architecture used in SAM to process the input image $I^{3 \times H \times W}$ through several attention stages and consequently produces multi-scale feature embeddings ${F_i}^{{D_{im}} \times \frac{H}{16} \times \frac{W}{16}}$ where $i \in \{1,2,3,4\}$, $D_{im}$ means dimension number. The final high-level image embedding from the last stage ${F_4}^{{D_{im}} \times \frac{H}{16} \times \frac{W}{16}}$ is dimensionally compressed to obtain ${F_{im}}^{{D_{pr}} \times \frac{H}{16} \times \frac{W}{16}}$, which matches the prompt embedding channel number $D_{pr}$. Concurrently, the first-stage embedding ${F_1}^{{D_{im}} \times \frac{H}{16} \times \frac{W}{16}}$ is retained to provide essential low-level details to the decoder.

\textbf{Prompt encoder} in other SAM-like models face multi-modal demands (points, boxes, text, masks) while requires only a dense prompt encoder optimized for tokenizing the input segmentation mask $S^{H \times W}$. This encoder sequentially comprises two staked blocks of convolution layer with $3 \times 3$ kernel size ($3 \times 3$ Conv), layer normalization (LN), and Gaussian Error Linear Unit (GELU), followed by an extra $3 \times 3$ Conv layer. It processes the unchecked segmentation mask $S^{H \times W}$ to the initial prompt feature map ${F_{pr}}^{{D_{pr}} \times \frac{H}{4} \times \frac{W}{4}}$. This feature map is then downsampled to match the spatial resolution of the image embedding ${F_{im}}^{{D_{pr}} \times \frac{H}{16} \times \frac{W}{16}}$, resulting in the final prompt embedding  ${F_{pr}}^{{D_{pr}} \times \frac{H}{16} \times \frac{W}{16}}$.

\textbf{High-quality decoder} is employed to fuse tokenized images and segmentation masks to produce the final four-class PQM assessment map. 

Specifically, the image embedding ${F_{im}}^{{D_{pr}} \times \frac{H}{16} \times \frac{W}{16}}$ and prompt features ${F_{pr}}^{{D_{pr}} \times \frac{H}{16} \times \frac{W}{16}}$ are firstly added element-wise. Positional embeddings are incorporated, and the result is then flattened and transposed to form the initial visual descriptors ${F}^{(\frac{H}{16} \times \frac{W}{16}) \times {D_{pr}}}$. Additionally, to enhance the low-level perception, both the feature map ${F_1}^{{D_{im}} \times \frac{H}{16} \times \frac{W}{16}}$ from the first image encoder stage and the final image representation ${F_{im}}^{{D_{pr}} \times \frac{H}{16} \times \frac{W}{16}}$ are upsampled by two transposed convolution layers, and then sum up to obtain high-quality image representations ${F_{hq}}^{{D_{pr}} \times \frac{H}{4} \times \frac{W}{4}}$. 

Subsequently, Two sets of learnable output tokens are initialized: standard output tokens ${O}^{4 \times {D_{pr}}}$ representing the four PQM classes (TP, FP, TN, FN), and high-quality output tokens ${O_{hq}}^{4 \times {D_{pr}}}$ following the HQ-SAM \cite{ke2023segment} design to facilitate refined predictions. Mathematically, the final output tokens ${O_f}^{8 \times {D_{pr}}}$ are yielded obeying $O_f=PE[O, O_{hq}]$ where $PE$ and $[\cdot]$ refer to the positional embedding and concatenation operation, respectively. 

After obtaining ${F}^{(\frac{H}{16} \times \frac{W}{16}) \times {D_{pr}}}$ and ${O_f}^{8 \times {D_{pr}}}$, features are propagated between them using two iterations of a Two Way Attention block. Each block consists of self-attention (SA) on the tokens, cross-attention (CA) from tokens to image features, an multi-layer perception (MLP), and another cross-attention from image features back to tokens (using the standard scaled dot-product attention mechanism, Eq.\ref{Eq:attention}):

\begin{align}\label{Eq:attention}
	\text{Attention}({Q}, {K}, {V}) = \text{softmax}\left( \frac{{Q} {K}^\top}{\sqrt{d_k}} \right) {V}.
\end{align}

where T means transpose operation, $\sqrt{d_k}$ refers to the dimensions of keys, ${Q}$, ${K}$ and ${V}$
are query (${Q}$), key (${K}$) and values (${V}$), respectively. Concretely, in the SA layer, ${Q}$, ${K}$ and ${V}$ are ${O_f}$ processed by three parallel linear projections, and in the first CA layer, the ${Q}$ is linearly projected ${O_f}$ while ${K}$ and ${V}$ are projected from $F$. The settings in the second CA layer is opposite to the former one. The outputs of the first TwoWayAttention are then received by the the second one as inputs. This process updates the tokens ${O_f}^{8 \times {D_{pr}}}$ to ${{O_f}^{\prime}}^{8 \times {D_{pr}}}$. Afterwards, a final cross-attention layer performs one last interaction between ${O_f}^{\prime}$ and visual features $F$, yielding the feature-infused output tokens ${O_m}^{8 \times {D_{pr}}}$ and updated visual descriptors ${F^{\prime}}^{(\frac{H}{16} \times \frac{W}{16}) \times {D_{pr}}}$. 

Then the updated tokens ${O_m}^{8 \times {D_{pr}}}$ are split back into ${O^{\prime}}^{4 \times {D_{pr}}}$ and ${{O_{hq}}^{\prime}}^{4 \times {D_{pr}}}$. They are subsequently processed by two different MLP blocks for initial and high-quality assessment predictions, respectively. In terms of initial assessment, the visual descriptors ${F^{\prime}}^{(\frac{H}{16} \times \frac{W}{16}) \times {D_{pr}}}$ is upsampled with ratio of 4 to ${F^{\prime}}^{(\frac{H}{4} \times \frac{W}{4}) \times {D_{pr}}}$ via two transposed convolution layers and element-wise multiplied with the processed standard tokens ${O^{\prime}}^{4 \times {D_{pr}}}$. The result is reshaped and upsampled to the full resolution initial assessment mask ${A_{init}}^{4 \times H\times W}$. As to high-quality assessment, high-quality image representations ${F_{hq}}^{{D_{pr}} \times \frac{H}{4} \times \frac{W}{4}}$ and reshaped visual descriptors ${F^{\prime}}^{{D_{pr}} \times \frac{H}{4} \times \frac{W}{4}}$ are added and merged by two convolution layers, and then flattened to obtain refined features ${{F_{hq}}^{\prime}}^{(\frac{H}{4} \times \frac{W}{4}) \times {D_{pr}}}$, which fuses high and low level features. After that, ${{F_{hq}}^{\prime}}^{(\frac{H}{4} \times \frac{W}{4}) \times {D_{pr}}}$ and ${{O_{hq}}^{\prime}}^{4 \times {D_{pr}}}$ are multiplied, reshaped and upsampled to the high-quality assessment mask ${A_{hq}}^{4 \times H\times W}$. Finally the ${A_{init}}^{4 \times H \times W}$ and ${A_{hq}}^{4 \times H\times W}$ are summed up to produce the coarse PQM output from the backbone, marked as ${A_{1}}^{4 \times H\times W}$.

\subsection{Edge guided compaction branch} \label{sec:EGC}
While the promptable backbone described above provides a coarse PQM assessment ${A_{1}}^{H\times W}$, further refinement is necessary. This is because, unlike standard SAM-based segmentation where prompts guide output generation, SegAssess evaluates the input mask itself. Besides, among four targeted classes in PQM, pixels belonging to FP, FN (i.e. wrong pixels) are much rarer than TP, TN (i.e. correct pixels) elements, and these erroneous regions prominent cluster in the narrow neighbourhood of object edges (see Section.~\ref{sec:data_stats}). On the one hand, this pattern provides potential prior information to guild the identification of FP/FN classes. On the other hand, it can also lead to overly diffuse error predictions near boundaries. Therefore, SegAssess incorporates an Edge Guided Compaction (EGC) branch (Fig.~\ref{Fig:overview}, orange box). This branch specifically aims to predict precise, one-pixel-wide object edges ${E}^{H \times W}$ derived from the image features, which are then used to refine the coarse assessment ${A_{1}}$.

The edge guided compaction branch leverages the multi-scale image embeddings ${F_1}$, ${F_2}$, ${F_3}$ and ${F_4}$ generated by the four stages of the image encoder. Specifically, the lower-level features  ${F_1}^{{D_{im}} \times \frac{H}{16} \times \frac{W}{16}}$, ${F_2}^{{D_{im}} \times \frac{H}{16} \times \frac{W}{16}}$, and ${F_3}^{{D_{im}} \times \frac{H}{16} \times \frac{W}{16}}$ are independently upsampled by two transposed convolutions to preliminary edge prediction sideouts ${E_1}^{H \times W}$, ${E_2}^{H \times W}$ and ${E_3}^{H \times W}$, respectively. The highest-level image features ${F_4}^{{D_{im}} \times \frac{H}{16} \times \frac{W}{16}}$ are first processed by the Aggregated Semantic Filter (ASF) module (see Section~\ref{sec:A3}) to enhance relevant semantic context for edge detection, and then similarly upsampled to produce the fourth edge sideout ${E_4}^{H \times W}$ via transpose convolutions. These four sideouts are then concatenated channel-wise into hierarchical sideouts ${E_{ms}}^{4 \times H \times W}$. Simultaneously, three convolutions with $1 \times 1$ kernel size are utilized to process ${E_4}^{H \times W}$ and produce a weighting map ${W}^{4 \times H \times W}$. The ${E_{ms}}^{4 \times H \times W}$ and ${W}^{4 \times H \times W}$ are multiplied and added up along the channel dimension to generate the final edge prediction ${E}^{H \times W}$. Given ${E}^{H \times W}$ and coarse assessment ${A_{1}}^{4 \times H\times W}$, then final SQA results ${A}^{4 \times H\times W}$ is calculated following Eq.~\ref{Eq:refine}:
\begin{align}\label{Eq:refine}
	A = NL(sigmoid(Conv(E)) \cdot Conv(A_1)),
\end{align}
where $Conv$ is a convolution operation with $3 \times 3$ kernel size, $\cdot$ refers to dot production and $NL$ denotes the non-local attention following the classic practice proposed in\cite{wang2018non}.

\begin{figure*}
	\centering
	\includegraphics[width=1\linewidth]{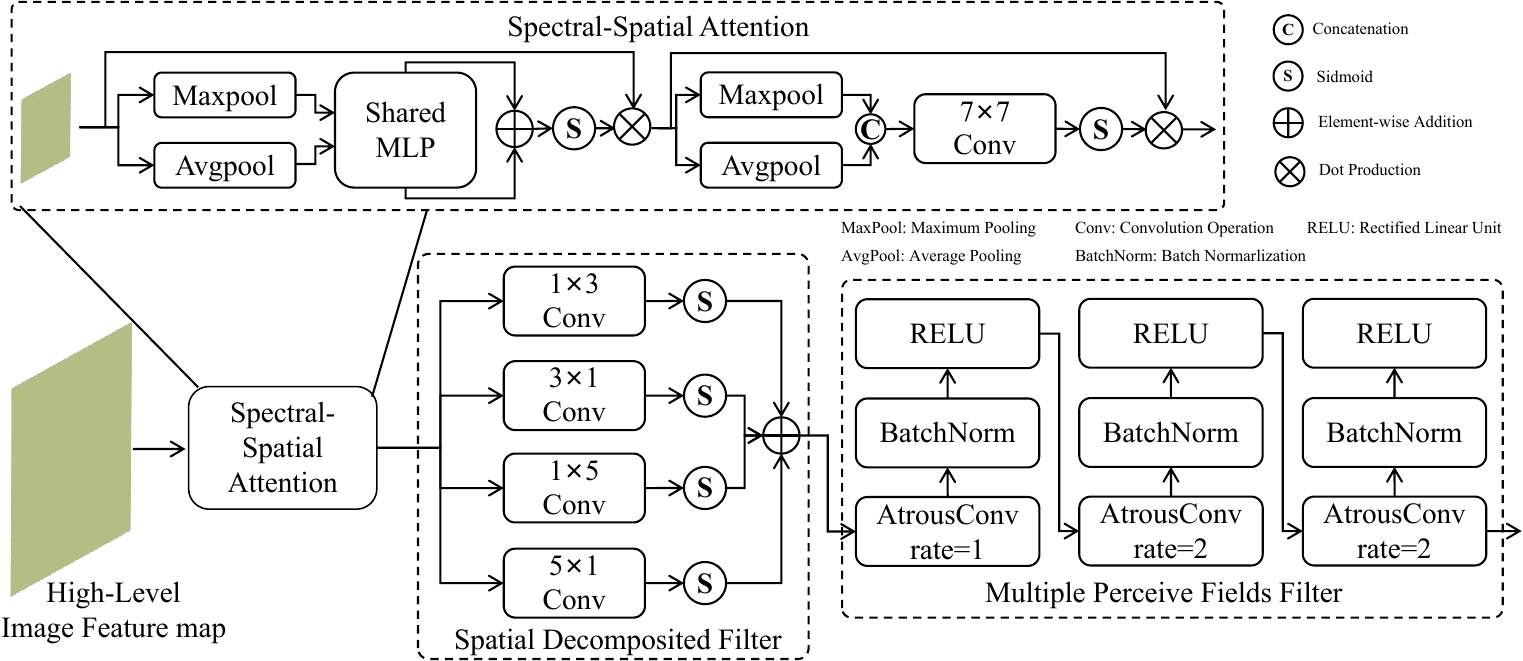}
	\caption{Detailed architecture of the Aggregated Semantic Filter (ASF) module. It integrates three perception mechanisms: spectral-spatial attention, a spatial decomposed filter, and a multiple perceive fields filter using atrous convolutions.}
	\label{Fig:ASF}
\end{figure*}

\subsection{Aggregated semantic filter module}\label{sec:A3}
The aggregated semantic filter (ASF) module processes the high-level image features ${F_4}^{{D_{im}} \times \frac{H}{16} \times \frac{W}{16}}$ from the image encoder. Its purpose, within the EGC branch, is to enhance semantic discrimination and focus on edge-relevant context, thereby improving the accuracy and compactness of the final edge prediction used for PQM refinement. As shown in Fig.~\ref{Fig:ASF}, the ASF achieves this by integrating three distinct perception mechanisms: spectral-spatial attention, spatial decomposited filter and multiple perceive fields filter. 

\textbf{Spectral-spatial attention:} Inspired by \cite{woo2018cbam}, this component first applies channel-wise (spectral) attention followed by spatial attention. The spectral attention receives ${F_4}^{{D_{im}} \times \frac{H}{16} \times \frac{W}{16}}$, and then parallelly execute maximum pooling (MaxPool) and average pooling (AvgPool). Both pooled outputs are fed through a shared MLP, summed element-wise, passed through a sigmoid activation, and the resulting channel attention weights are multiplied back with the input ${F_4}^{{D_{im}} \times \frac{H}{16} \times \frac{W}{16}}$ to produce spectrally refined features  ${F_{spr}}^{{D_{im}} \times \frac{H}{16} \times \frac{W}{16}}$. Afterwards, two parallel MaxPool and AvgPool are applied to ${F_{spr}}^{{D_{im}} \times \frac{H}{16} \times \frac{W}{16}}$ and outcomes are concatenated along the dimension channel. This concatenated map is then passed through a convolution layer with $7 \times 7$ kernel size and $3 \times 3$ padding size ($Conv^{7 \times 7}$). The output is activated and multiplied with ${F_{spr}}^{{D_{im}} \times \frac{H}{16} \times \frac{W}{16}}$ to yield the spectrally and spatially attended features ${F_{ss}}^{{D_{im}} \times \frac{H}{16} \times \frac{W}{16}}$, which is the output of the entire spectral-spatial module. The aforementioned procedures can be formulated as Eq.~\ref{Eq:CBAM}:
\begin{equation}\label{Eq:CBAM}
	\begin{aligned}
	F_{spr} &= {F_4} \cdot (sigmoid(MLP(MaxPool(F_4)) \\&+ MLP(AvgPool(F_4)))),\\ 
	F_{ss} &= F_{spr} \cdot sigmoid(Conv^{7 \times 7}([MaxPool(F_{spr}), \\& AvgPool(F_{spr})])),
\end{aligned}
\end{equation}
where $[\cdot]$ means the concatenation operations.

\textbf{Spatial decomposited filter:} This filter applies convolutions decomposed into horizontal and vertical directions to the attention-refined features ${F_{ss}}^{{D_{im}} \times \frac{H}{16} \times \frac{W}{16}}$. As formulated in Eq.~\ref{Eq:SDF}, four parallel asymmetric convolutions followed with sigmoid activation are setted to process ${F_{ss}}^{{D_{im}} \times \frac{H}{16} \times \frac{W}{16}}$. Kernel sizes and padding sizes of these convolutions are $1 \times 3$, $3 \times 1$, $1 \times 5$, $5 \times 1$ and $0 \times 1$, $1 \times 0$, $0 \times 2$, $2 \times 0$, respectively. Outcomes of these convolutions are added up as the ${F_{sdf}}^{{D_{im}} \times \frac{H}{16} \times \frac{W}{16}}$.
\begin{equation}\label{Eq:SDF}
\begin{split}
	F_{sdf} &= sigmoid(Conv^{1 \times 3}(F_{ss})) \\&+ sigmoid(Conv^{3 \times 1}(F_{ss})) +sigmoid(Conv^{1 \times 5}(F_{ss})) \\&+ sigmoid(Conv^{5 \times 1}(F_{ss})).
\end{split}
\end{equation}

\textbf{Multiple perceive fields filter:} Finally, to capture contextual information at varying scales, three Atrous Convolution blocks are applied sequentially to ${F_{sdf}}^{{D_{im}} \times \frac{H}{16} \times \frac{W}{16}}$, obtaining the final results of the entire ASF module. Each block consists of an Atrous Convolution layer (using sequential dilation rates of 1, 2, and 3 for the three blocks, respectively), followed by Batch Normalization (BatchNorm) and a Rectified Linear Unit (RELU) activation. The output of the third Atrous Convolution block represents the final output of the entire ASF module. This refined high-level feature representation is then passed back to the EGC branch (Section \ref{sec:EGC}) for generating the ${E_4}$ edge sideout.

\begin{figure*}
	\centering
	\includegraphics[width=1\linewidth]{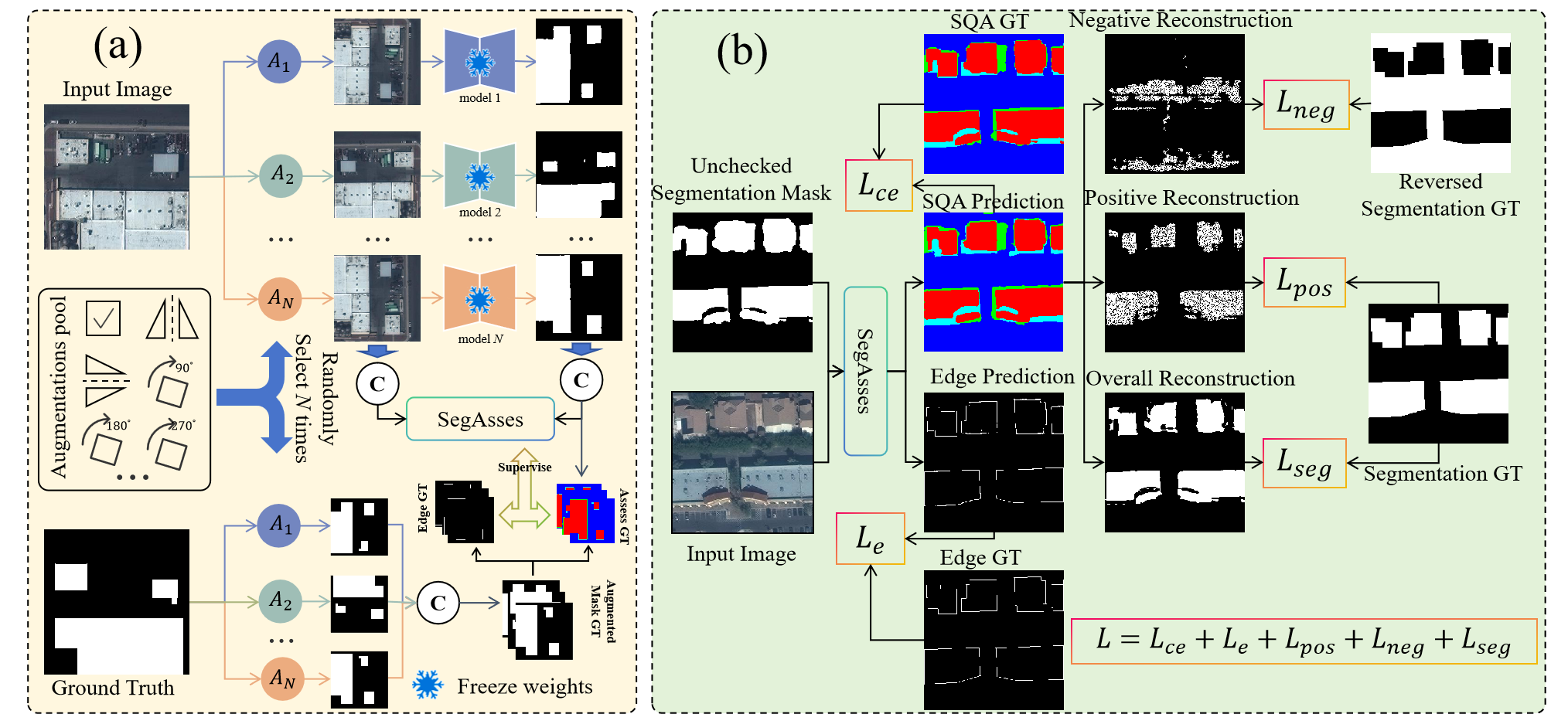}
	\caption{Supporting strategies for SegAssess training. (a) The Augmented Mixup Sampling (AMS) strategy used during training to enhance robustness by generating diverse masks online using multiple frozen segmentation models. (b) Overview of the multi-component loss function calculation used for supervision.}
	\label{Fig:AMS}
\end{figure*}

\subsection{Augmented mixup sampling}\label{sec:AMS}
Real-world SQA applications must handle segmentation masks originating from diverse sources, including manual annotations and outputs from various segmentation models. Traditional DL-based SQA models, often trained on masks from a single source (e.g., one specific model or manual annotations), typically struggle to generalize across these different mask characteristics, limiting their practical transferability. Inspired by successful data mixing strategies in foundation model training \cite{kirillov2023segment, guo2024skysense,bountos2025fomo}, we propose an Augmented Mixup Sampling (AMS) strategy to explicitly enhance the cross-domain robustness and transferability of SegAssess. AMS enriches training data diversity by dynamically generating and incorporating masks from multiple different segmentation models online. 

As illustrated in Fig.~\ref{Fig:AMS}(a), in training, random transformations are executed for each original training sample pair of RS image $I^{3 \times H \times W}$ and corresponding binary segmentation mask ground truth (GT) $\bar{S}^{H \times W}$. Specifically, $N$ random geometric transformations (e.g., rotation, flipping), denoted as ${Z_i} \in {Z_N}$, are selected from an augmentation pool and applied to the input image $I$, creating $N$ augmented images ${I_i}^{3 \times H \times W} \in {I_{aug}}^{N \times 3 \times H \times W}$. The same corresponding transformations are applied to the ground truth mask $\bar{S}^{H \times W}$ to produce consistent augmented ground truth masks ${\bar{S}_i}^{H \times W} \in {\bar{S}_{aug}}^{N \times H \times W}$. Subsequently, corresponding one-pixel-wide edge ground truth maps ${\bar{E}_i}^{H \times W} \in {\bar{E}_{aug}}^{N \times H \times W}$ are generated from the augmented masks ${\bar{S}_{aug}}^{N \times H \times W}$ using standard edge detection algorithms. After obtaining augmented images ${I_i}^{3 \times H \times W} \in {I_{aug}}^{N \times 3 \times H \times W}$, $N$ trained segmentation models ${M_i} \in {M_N}$ are employed to perform inference on the corresponding augmented image ${I_i} \in {I_{aug}}$ to generate unchecked segmentation masks, denoted as ${S_i}^{H \times W} \in {S_{aug}}^{N \times H \times W}$. This step dynamically creates a diverse set of $N$ unchecked segmentation masks, reflecting variations inherent to different segmentation architectures and qualities. Finally, the resultant augmented images ${I_{aug}}^{N \times 3 \times H \times W}$ and predicted segmentation masks ${S_i}^{H \times W} \in {S_{aug}}^{N \times H \times W}$ are received by the SegAssess as inputs while the augmented mask GT ${\bar{S}_{aug}}^{N \times H \times W}$ and edge GT ${\bar{E}_{aug}}^{N \times H \times W}$ are utilized to supervise the model learning (See Section~\ref{sec:loss}). 

By exposing SegAssess to masks generated by multiple diverse models during training, AMS forces the model to learn features that are robust to variations in segmentation style and quality, significantly enhancing its ability to generalize to unseen mask sources at inference time. It is important to note that during inference, AMS is not used; SegAssess processes the original input image and the single segmentation mask provided for evaluation.

\subsection{Loss Functions}\label{sec:loss}
SegAssess is trained end-to-end in a supervised manner using a composite loss function designed to optimize both the primary PQM task and the auxiliary edge prediction task, while also enforcing prediction consistency (see Fig.~\ref{Fig:AMS}(b)). This section is going to provide mathematical introduction to these loss functions. Notably, Although training utilizes the augmented samples generated via AMS (Section \ref{sec:AMS}) the loss functions are described below assuming a single, non-augmented sample for notational simplicity. The total loss $\mathcal{L}$ comprises five components:

\textbf{Weighted Cross-Entropy Loss ($\mathcal{L}_{ce}$): }This is the primary loss for the four-class PQM task. It measures the discrepancy between the predicted PQM probability map ${A}^{H\times W}$ and the ground truth assessment map $\bar{A}$, following Eq.~\ref{Eq:ce}:
\begin{align}\label{Eq:ce}
	\mathcal{L}_{ce} = \mathcal{M}{(w \cdot {\bar{A}} \log(A))},
\end{align} 
where $\mathcal{M}(\cdot)$ represents the arithmetic mean, $\bar{A}$ is the ground truth (GT) of assessment mask and $w$ is the weighting matrix handling the inherent imbalance among the TP, FP, TN, and FN categories. 

\textbf{Edge Prediction Loss ($\mathcal{L}_{e}$): }This loss supervises the auxiliary edge prediction task from the EGC branch. Due to the severe imbalance between the sparse one-pixel-wide edge foreground and the extensive background in the ground truth edge map, $\mathcal{L}_{e}$ combines a weighted binary cross-entropy (BCE) term and a Dice loss \cite{milletari2016v} term, as formulated in Eq.~\ref{Eq:edge_mse}:
\begin{equation}\label{Eq:edge_mse}
\begin{split}
	\mathcal{L}_e &= \mathcal{M}{(-\gamma{((1 - \bar{E})\log(1 - {E}) + \bar{E}\log({E}))})} \\&+ \mathcal{M}(1 - \frac{2 E \bar{E} + \epsilon}{sum(E) + sum(\bar{E}) + \epsilon})
\end{split}
\end{equation} 
where $E$ is the predicted edge probability map, $\epsilon$ is a small constant for numerical stability, and $\gamma$ is a pixel-wise weight for the BCE term obeying Eq.~\ref{Eq:rcf}: 
\begin{equation}\label{Eq:rcf}
	\gamma = 
	\begin{cases}
		\frac{\left|\bar{E}^{-}\right|}{\left|\bar{E}^{+}\right| + \left|\bar{E}^{-}\right|},  & {p=1} \\
		\lambda\ {\frac{\left|\bar{E}^{+}\right|}{\left|\bar{E}^{+}\right| + \left|\bar{E}^{-}\right|}},  & {p = 0}		
	\end{cases},
\end{equation}
where $p$ represents each pixel value in $\bar{E}$, with $\left| \bar{E}^{+} \right|$ and $\left| \bar{E}^{-} \right|$ tallying positive and negative pixels in GT edge masks, respectively. The parameter $\lambda$, which dictates the penalty for negatives, is fixed at 1.1.

\textbf{Reconstruction Losses ($\mathcal{L}_{pos}$, $\mathcal{L}_{neg}$ and $\mathcal{L}_{seg}$): }Since the predicted PQM categories (TP, FP, TN, FN) implicitly define a reconstruction of the original segmentation ground truth $\bar{S}$, three additional losses are introduced to enforce this consistency. These losses, denoted as $\mathcal{L}_{pos}$, $\mathcal{L}_{neg}$ and $\mathcal{L}_{seg}$, are formulated as Eq.~\ref{Eq:reconstruction}:  
\begin{subequations}\label{Eq:reconstruction}
	\begin{align}
		\mathcal{L}_{pos} &= \mathcal{M}{({TP + FN - \bar{S})^2}}, \label{Eq:L_pos} \\
		\mathcal{L}_{neg} &= \mathcal{M}{({FP + TN - (1 - \bar{S}))^2}}, \label{Eq:L_neg} \\
		\mathcal{L}_{seg} &= \mathcal{M}{({A + FN - FP - \bar{S})^2}}. \label{Eq:L_seg}
	\end{align}
\end{subequations} 

Finally, the total loss of SegAssess is the linear addition of aforementioned five components, which is $\mathcal{L} = \mathcal{L}_{ce} + \mathcal{L}_e + \mathcal{L}_{pos} + \mathcal{L}_{neg} + \mathcal{L}_{seg}$.

\section{Experimental settings}\label{settings}

\subsection{Implement details}\label{details}
SegAssess adopts HQ-SAM base version as its backbone, where numbers of Transformer layers in four image encoder stages are 2, 5, 8, 11, respectively. The image and prompt dimension numbers are set as $D_{im}$=768, $D_{pr}$=256, respectively. All parameters are trainable. As to AMS, we apply $N$=4 random transformations on training samples. The augmentations pool contains clockwise rotations (90$^{\circ}$, 180$^{\circ}$, 270$^{\circ}$), vertical flipping and horizontal flipping while the trained segmentation models used to produce unchecked masks include DeepLabv3+ \cite{chen2018encoder}, HRNet \cite{wang2020deep}, TransUNet \cite{chen2021transunet}, UnetFormer \cite{wang2022unetformer}. These models represent CNN-based, Transformer-based, and hybrid architectures, ensuring exposure to varied mask characteristics during training. The training stage is executed utilizing Adam optimizer, 10$^{-4}$ learning rate and early stopping strategy. Based on empirical analysis of class distributions across all datasets, we empirically set class weights $w$ in $\mathcal{L}_{e}$ for TP, FP, TN and FN as 0.5,5.0,0.1 and 5.0, respectively. The computing platform is equipped with NVIDIA H800 GPU and Intel Xeon Platinum 8463B CPU.

\subsection{Datasets}\label{data}
The development and evaluation of SegAssess utilize a total of 32 distinct SQA datasets derived from 6 primary source datasets. These are detailed below.
 
\subsubsection{Derivatives from image segmentation datasets}\label{sec:model_source}
To generate realistic SQA scenarios with masks produced by diverse models, we first utilized four publicly available semantic segmentation datasets: Inria Aerial Image Labeling dataset (Inria)\cite{maggiori2017Inria} and CrowdAI Map Challenge dataset (CrowdAI)\cite{mohanty2020crowdai}, DeepGlobe Road Extraction Challenge dataset\cite{DeepGlobe18} and five-classes Gaofen Image Dataset (GID)\cite{tong2020GID}. Each dataset underwent specific preprocessing as follows:

\textbf{Inria} dataset contains 180 aerial RGB images (0.3m resolution, 5000$\times$5000 pixels) covering 5 cities, with binary building annotations. We designated the first and last image from each city for validation (10 images) and the rest for training (170 images). All images were then cropped into non-overlapping 320$\times$320 pixel tiles, and tiles containing no buildings were discarded.

\textbf{CrowdAI} dataset comprises 300$\times$300 WorldView-3 satellite RGB images (0.3m resolution) with building annotations. The small version of validation set, which holds 1820 samples is adopted in this work. Furthermore, as suggested in\cite{yang2025diffvector}, we filtered the sets to remove duplicate and leaked samples\cite{adimoolam2023efficient}, resulting in 67440 training and 1480 validation samples.

\textbf{DeepGlobe} dataset holds 1024$\times$1024 pixel RGB satellite images (0.5m resolution) with road annotations, originally split into training and testing sets (totaling 6234 images). These images were cropped into 512$\times$512 pixel tiles. Tiles containing no roads were filtered out. The resulting tiles were then randomly split into training and validation sets, yielding 34551 training and 4966 validation image-GT tile pairs (ensuring tiles from the original test images remained in the validation set). 

\begin{figure*}[!t]
	\centering
	
	\subfloat[]{
		\centering
		\includegraphics[width=0.33\linewidth]{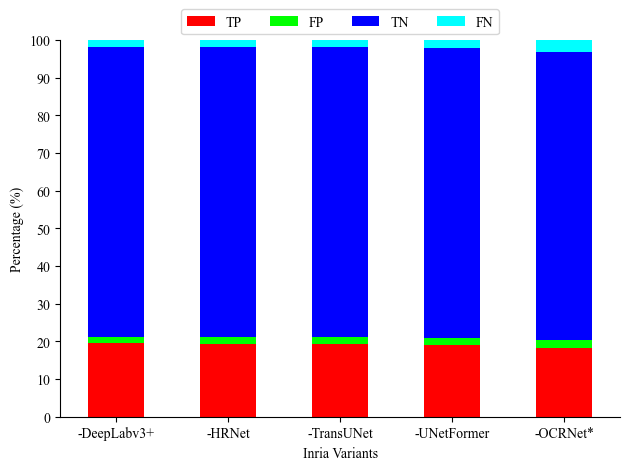}
	}
	\subfloat[]{
		\centering
		\includegraphics[width=0.33\linewidth]{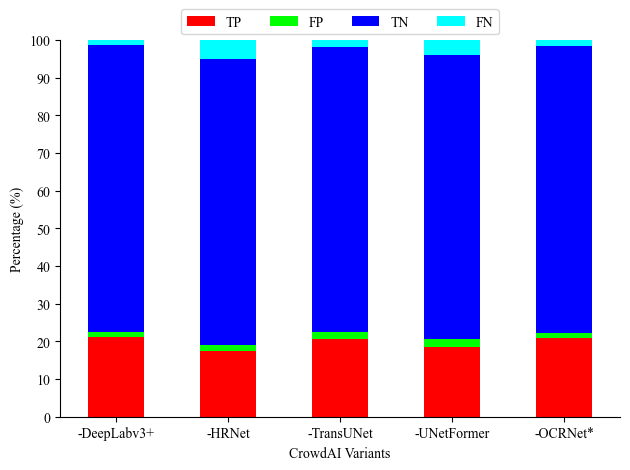}
	}
	\subfloat[]{
		\centering
		\includegraphics[width=0.33\linewidth]{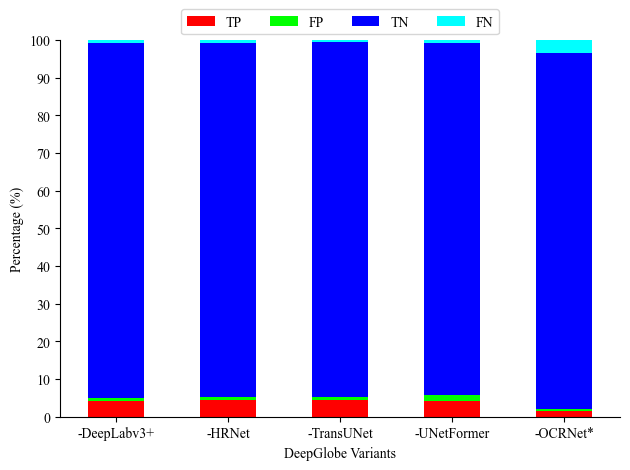}
	}
	
	\vspace{-0.7em}
	
	\subfloat[]{
		\centering
		\includegraphics[width=0.33\linewidth]{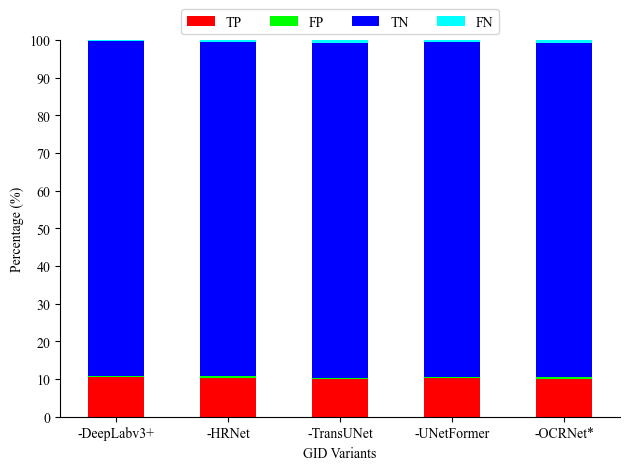}
	}
	\subfloat[]{
		\centering
		\includegraphics[width=0.33\linewidth]{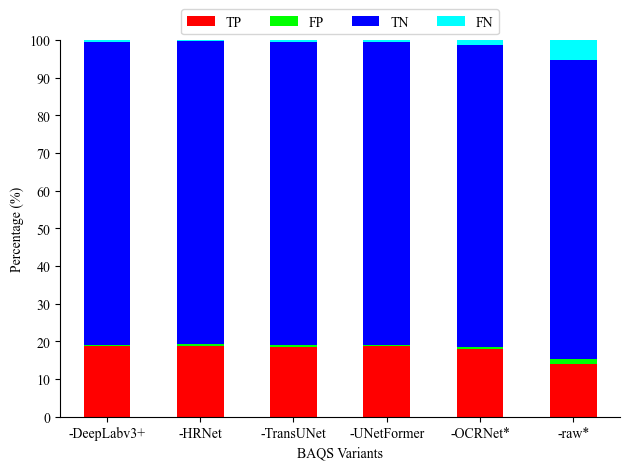}
	}
	\subfloat[]{
		\centering
		\includegraphics[width=0.33\linewidth]{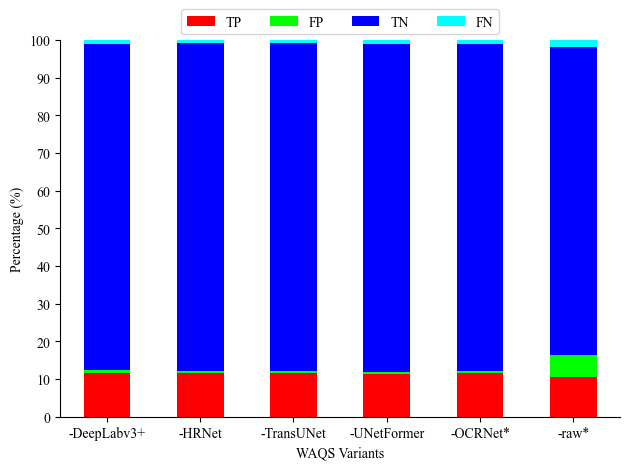}
	}
	\vspace{-0.7em}
	\subfloat[]{
		\centering
		\includegraphics[width=0.49\linewidth]{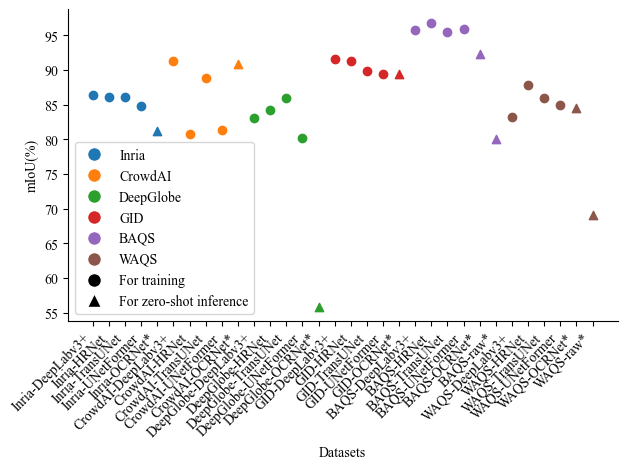}
	}
	\subfloat[]{
		\centering
		\includegraphics[width=0.49\linewidth]{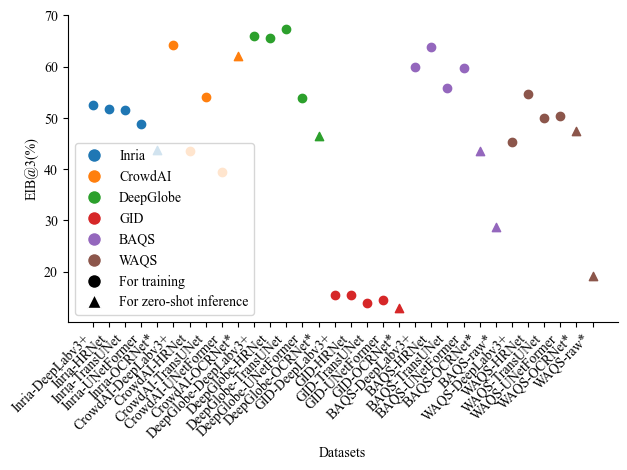}
	}
	
	\caption{Statistics of datasets. Panels (a)-(f) depict pixel percentage distribution for each PQM class (TP, FP, TN, FN) across all Inria, CrowdAI, DeepGlobe, GID, BAQS and WAQS variants, respectively. (g) and (f) illustrates the mIoU and EIB@3. Dataset names marked by superscript $^{\ast}$ are utilized to evaluate the zero-shot transferability of SegAssess, while others are used for the standard train-validation protocol (including AMS training)}  
	\label{fig:data}
\end{figure*}

\textbf{GID} dataset provides Gaofen-2 satellite images with the spatial resolution of 4m contains fifteen-classes and five-classes raster annotations. We employed the latter set and relabelled its raw mask labels to water/no-water binary maps. The original large images were cropped into non-overlapping 512$\times$512 pixel patches without filtering. These patches were then split into training and validation sets following the radio of 8:2, resulting in 25500 training and 6000 validation samples. 

\textbf{SQA Variant Generation}: After preprocessing these four base datasets, five different deep learning segmentation models were independently trained on the training split of each dataset for the respective binary segmentation task (buildings, roads, or water). The models used were: DeepLabv3+\cite{chen2018encoder}, HRNet \cite{wang2020deep}, TransUNet \cite{chen2021transunet}, UNetFormer\cite{wang2022unetformer} and OCRNet\cite{yuan2020object}. Subsequently, for every image tile in both the training and validation sets of each base dataset, each of these five trained models was used to generate a segmentation mask. This process resulted in multiple SQA dataset variants for each base dataset, where each variant contains pairs of (image tile, generated mask, ground truth mask). These derived datasets are denoted using the convention 'X-Y', where 'X' is the source dataset name (Inria, CrowdAI, DeepGlobe, GID) and 'Y' is the name of the model used to generate the segmentation masks under evaluation. 


\subsubsection{Derivatives from segmentation quality assessment datasets}
In addition to generating SQA variants from standard segmentation datasets, we also utilized two existing benchmark datasets specifically designed for SQA, which inherently provide human-sourced segmentation masks requiring assessment alongside curated ground truth. These two SQA datasets are Building AQS dataset (BAQS) and Water-body AQS dataset (WAQS)\cite{zhang2023automatic}.

\textbf{BAQS} dataset is adapted from the Wuhan University Building Change 
Detection (WHU-BCD) dataset\cite {ji2018fully}. For the SQA task, the original 2012 building annotations serve as the 'masks under evaluation' (representing potentially outdated or imperfect human annotations), while the 2016 annotations were manually inspected and corrected to serve as the high-quality ground truth (GT) masks. The dataset consists of 512$\times$512 pixel image tiles (0.2m resolution), providing triplets of (image, mask under evaluation, GT mask). The dataset contains 6,120 training and 448 validation samples.

\textbf{WAQS} dataset is sampled from the four-class land cover dataset, entitled Hubei Dataset (HBD4), whose annotations are partially inconsistent with images. WAQS selected 44 images of 5000$\times$5000 pixels from HBD4 to manually correct the water annotations, and then cropped these images and corresponding raw and corrected annotations to 512$\times$512 pixels. Subsequently, all resultant tiles are split to 14400 and 968 samples for training and validation, respectively. Each sample include an RGB Gaofen-1 satellite image with a spatial resolution of 2m, a raw water binary mask for evaluation and a corrected water annotation as the segmentation GT.    

\textbf{SQA Variant Generation from BAQS \& WAQS:} These two datasets provide valuable human-sourced masks for evaluation ('BAQS-raw' and 'WAQS-raw' variants). To ensure comparability with the datasets derived in Section 4.2.1 and to further enrich the diversity for training and testing, we also generated model-sourced variants from BAQS and WAQS. Using the same five pre-trained segmentation models (DeepLabv3+, HRNet, TransUNet, UNetFormer, OCRNet), we generated corresponding segmentation masks for each image tile in BAQS and WAQS. This created additional SQA dataset variants named 'BAQS-Y' and 'WAQS-Y' (where Y is the model name), analogous to the variants described in Section \ref{sec:model_source}. 

In total, the experimental validation of SegAssess utilizes 32 distinct SQA dataset variants derived from the six base datasets described above (Inria, CrowdAI, DeepGlobe, GID, BAQS, WAQS). These 32 variants are strategically employed to evaluate SegAssess under different conditions, simulating key real-world application scenarios:\\
(1) \textbf{Model Development Training/Validation Sets (24 variants):} For developing SegAssess using the Augmented Mixup Sampling strategy (Section \ref{sec:AMS}), we utilize the variants generated by four specific models (DeepLabv3+, HRNet, TransUNet, UNetFormer) across all six base datasets (6 sources $\times$ 4 models = 24 datasets).\\ (2) \textbf{Zero-Shot Model-Source (ZS-MS) Test Sets (6 variants)}: To evaluate SegAssess's zero-shot transferability to masks generated by an unseen model architecture, we reserve the variants generated by the fifth model (OCRNet) across all six base datasets (6 sources $\times$ 1 model = 6 datasets). SegAssess is not trained on any OCRNet-generated masks.\\ (3) \textbf{Zero-Shot Human-Source (ZS-HS) Test Sets (2 variants):} To evaluate SegAssess's zero-shot transferability to unseen human-sourced masks with potentially different characteristics, we use the 'BAQS-raw' and 'WAQS-raw' datasets, which contain the original human annotations as masks under evaluation. SegAssess is not explicitly trained on these raw human-sourced masks during the AMS phase.

This experimental design allows for a comprehensive assessment of SegAssess's performance under standard training/validation conditions (within the distribution seen training) and its crucial zero-shot generalization capabilities to both unseen model types and unseen human annotations.

\subsubsection{Statistics of datasets}\label{sec:data_stats}
To better understand the characteristics influencing the SQA task, we analyzed several statistical properties across the 32 derived dataset variants, including pixel percentage distribution among the four PQM categories (i.e. TP, FP, TN, FN), percentage of error pixels (i.e. FP, FN) within a 3-pixel buffer zone of object edges (EIB@3) and mean IoU (mIoU, see Eq.~\ref{Eq:miou}) of unchecked binary segmentation mask against GT. Resultant statistics are depicted in Fig.~\ref{fig:data} and several key findings can be concluded. 

A prominent observation across nearly all datasets is the severe class imbalance, where error pixels (FP and FN) are significantly rarer than correctly classified pixels (TP and especially TN). This imbalance underscores the necessity for the class weighting scheme employed in the primary loss function $\mathcal{L}_{ce}$ to ensure sufficient learning focus on the minority error classes. Secondly, the analysis highlights a distinct spatial pattern of errors, particularly for model-generated masks. In most such variants (except GID), a high percentage of errors cluster near object boundaries, as indicated by large EIB@3 values. This finding reinforces the rationale for incorporating the Edge Guided Compaction (EGC) branch, specifically designed to improve prediction accuracy in these challenging edge regions. Conversely, the human-sourced masks (BAQS-raw, WAQS-raw) exhibited generally lower EIB@3 values, suggesting potentially different underlying error mechanisms. Finally, the statistics reveal significant source-dependent characteristics. While masks generated by different models for the same base dataset often share similar statistical profiles, these profiles differ markedly from those of the human-sourced masks in terms of class balance, edge error concentration, and mIoU, consequently challenging the transferability of SegAssess.

\subsection{Evaluation metrics}\label{metrics}
Since SegAssess performs segmentation quality assessment by generating a four-class Panoramic Quality Map (predicting TP, FP, TN, and FN pixels), its performance is evaluated using standard accuracy metrics commonly employed in multi-class semantic segmentation tasks. This approach treats the PQM output as a segmentation map and measures its pixel-wise agreement with the ground truth assessment map. For each class, a two-class (foreground and background) confusion matrix (CM) is yielded from the one-shot predicted and GT mask. In a CM, the true positive (TP) sum up the diagonal values, the false positive (FP) and false negative (FN) are the summation of column and row non-diagonal values, respectively. After that, we compute the standard per-class  F1-score (F1) and interaction over union (IoU), can be calculated following Eq.~\ref{Eq:metrics}. To assess overall performance across all four classes, Eq.~\ref{Eq:mf1} and Eq.~\ref{Eq:miou} average the F1$^c$ and IoU$^c$ of all $C$ classes , denoted as mF1 and mIoU. 
\begin{subequations}\label{Eq:metrics}
	\begin{align}
		&\text{TN} = sum(\text{CM}) - (\text{TP}+\text{FN}+\text{FP}),\\
		&\text{Precision} = \frac{\text{TP}}{\text{TP}+\text{FP}},\\
		&\text{Recall} = \frac{\text{TP}}{\text{TP}+\text{FN}},\\
		&\text{F1} = \frac{2 \times \text{Precision} \times \text{Recall}}{\text{Precision}+\text{Recall}},\\
		&{\text{IoU}} = \frac{\text{TP}}{\text{TP}+\text{FP}+\text{FN}},\\
		&{\text{mF1}} = \frac{1}{C}\sum_{c=1}^{C}\text{F1}^c,\label{Eq:mf1}\\
		&{\text{mIoU}} = \frac{1}{C}\sum_{c=1}^{C}\text{IoU}^c.\label{Eq:miou}
	\end{align}
\end{subequations} 


%
\section{Results and discussion}\label{results}
\subsection{Ablation study}\label{ablation}
\begin{figure*}[htbp]
	\centering
	
	\subfloat[F1 on BAQS-UNetFormer]{
		\centering
		\includegraphics[width=0.49\linewidth]{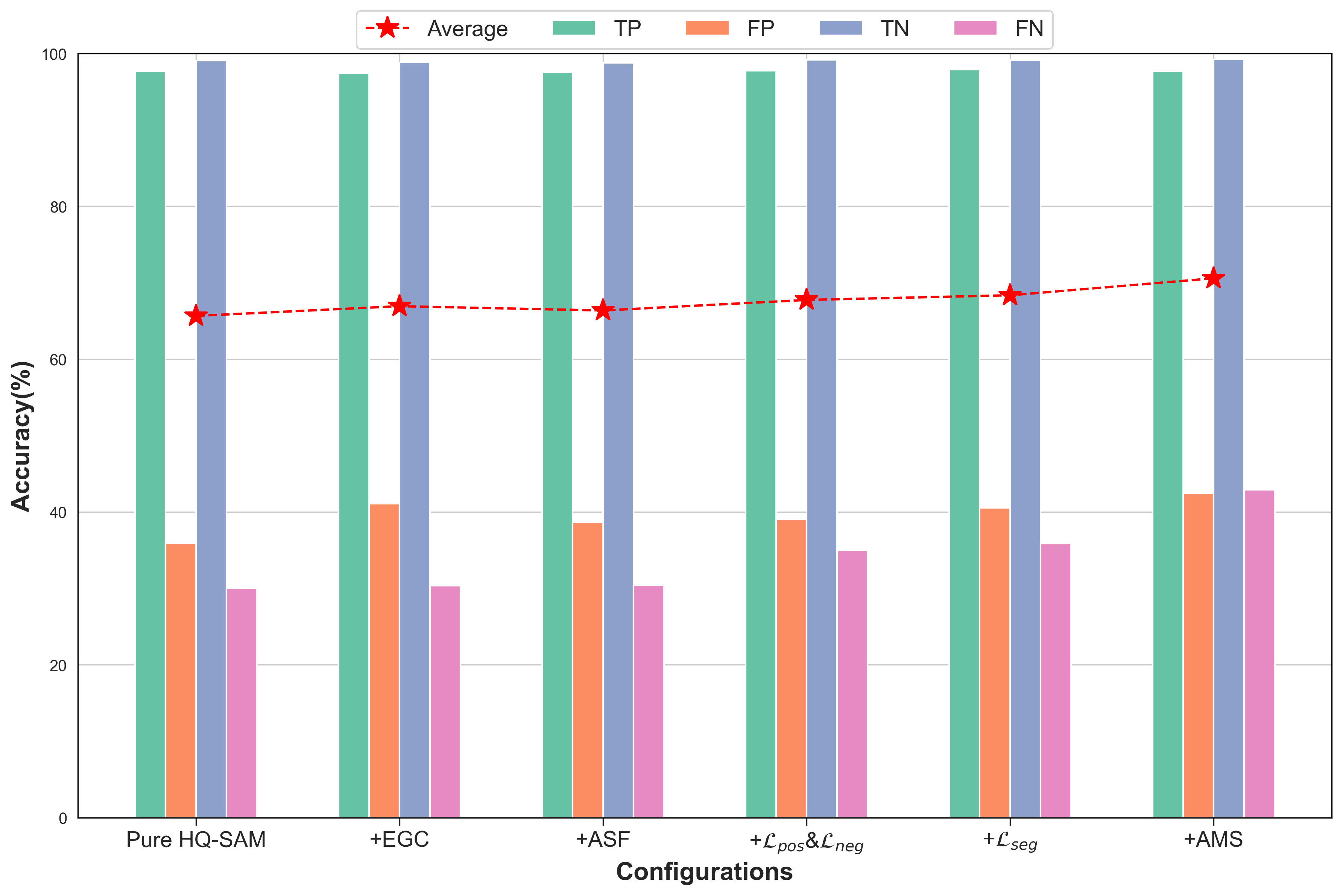}
	}
	\subfloat[IoU on BAQS-UNetFormer]{
		\centering
		\includegraphics[width=0.49\linewidth]{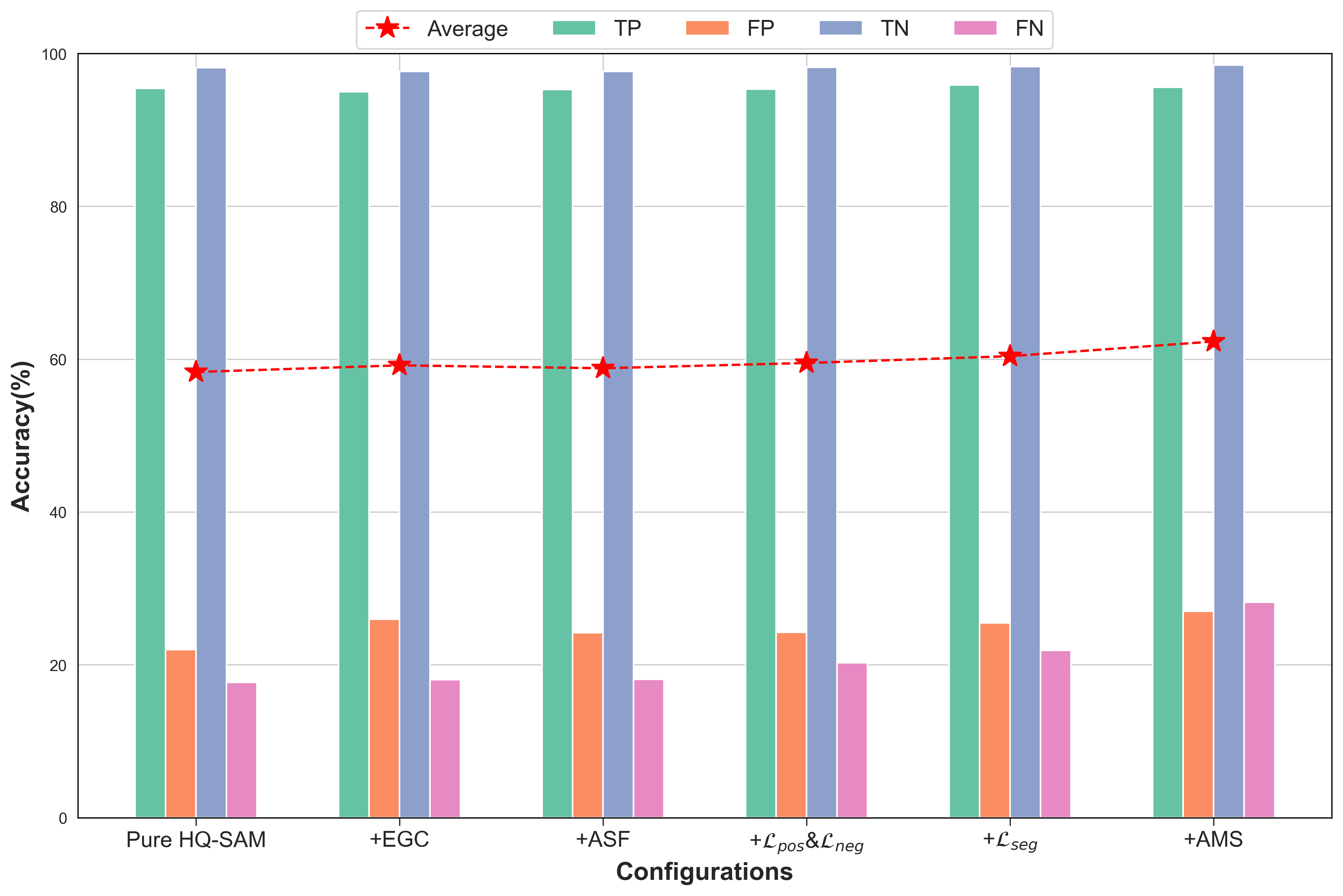}
	}
	
	\vspace{-0.7em}
	
	\subfloat[F1 on BAQS-OCRNet]{
		\centering
		\includegraphics[width=0.49\linewidth]{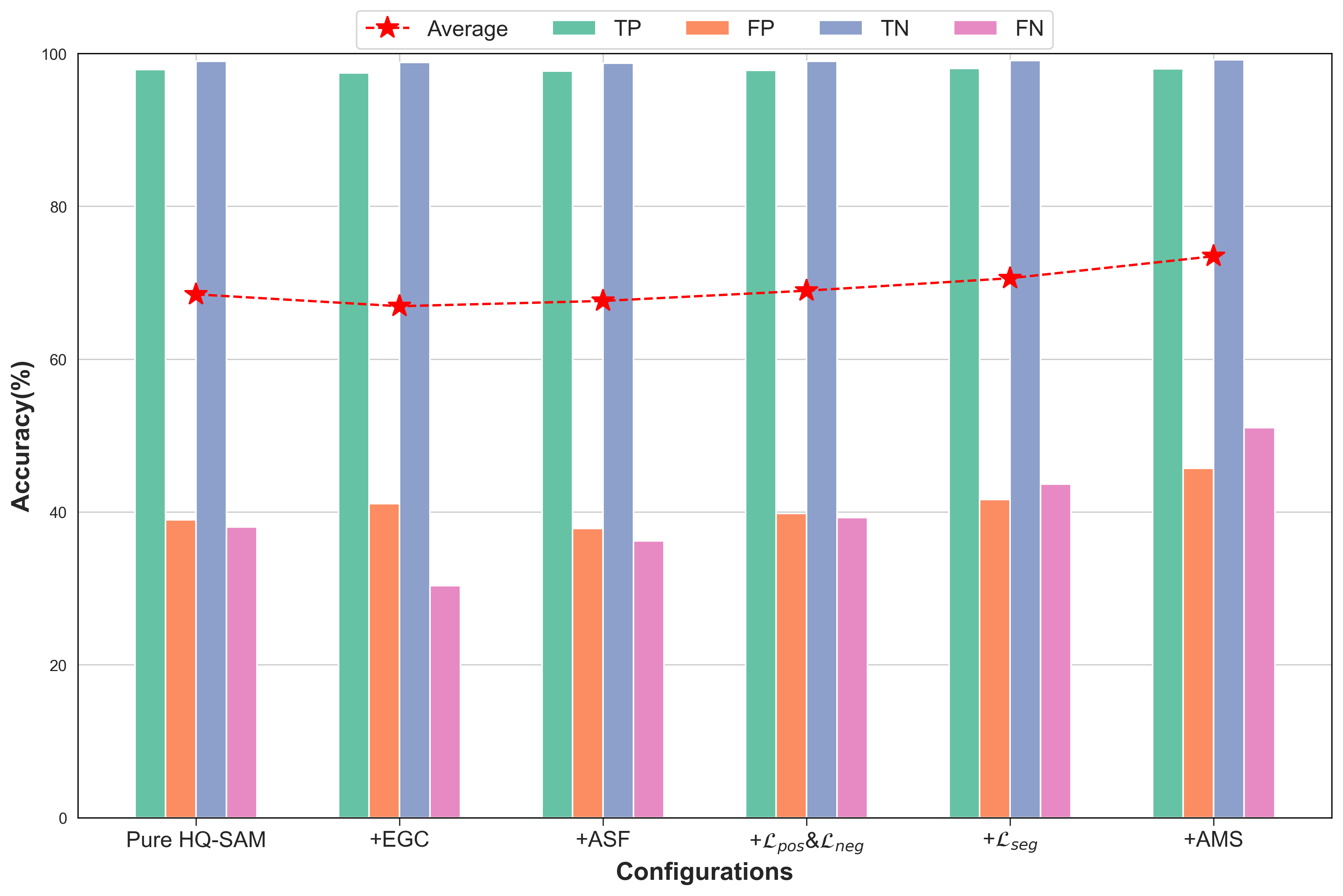}
	}
	\subfloat[IoU on BAQS-OCRNet]{
		\centering
		\includegraphics[width=0.49\linewidth]{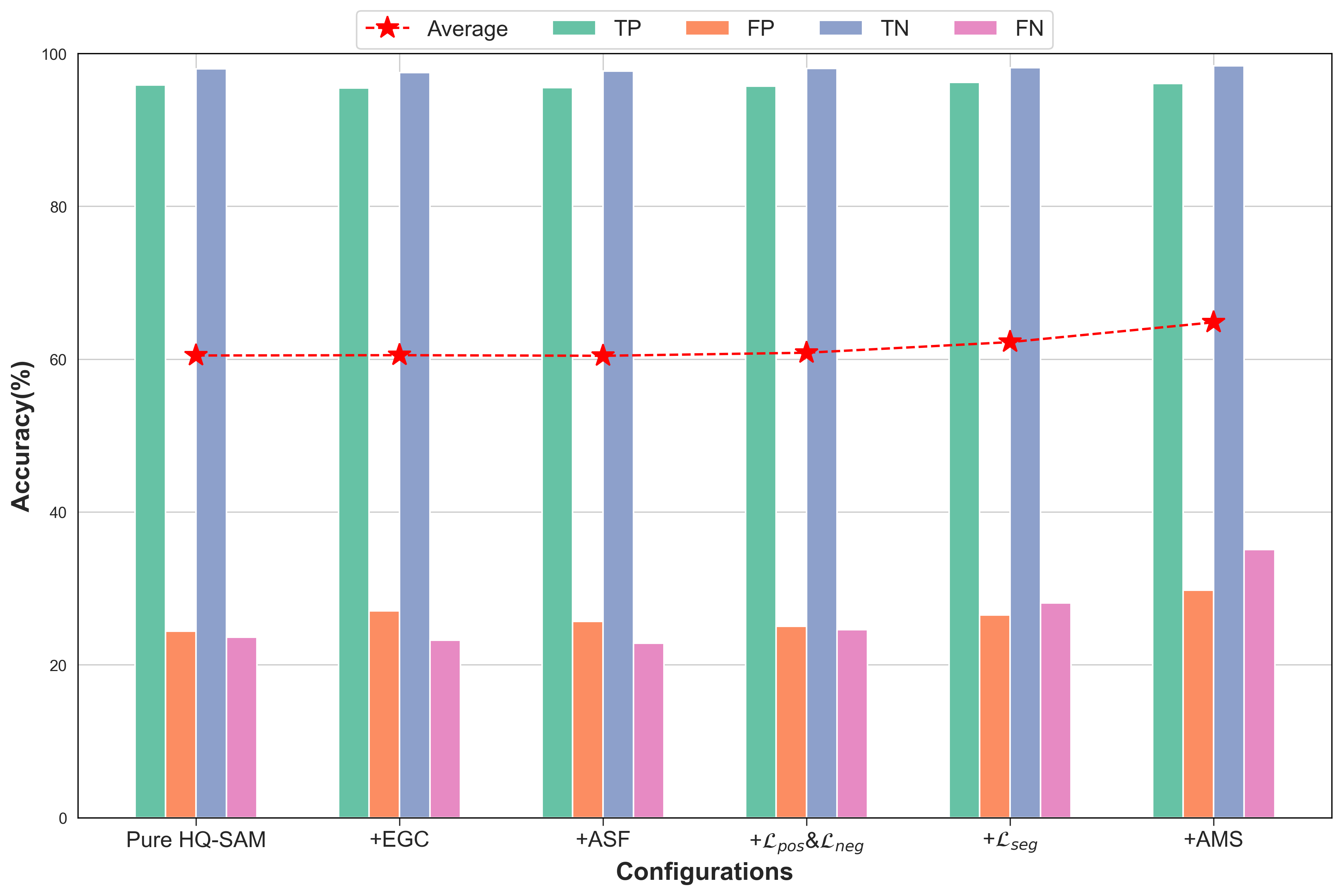}
	}
	
	\vspace{-0.7em}
	
	\subfloat[F1 on BAQS-raw]{
		\centering
		\includegraphics[width=0.49\linewidth]{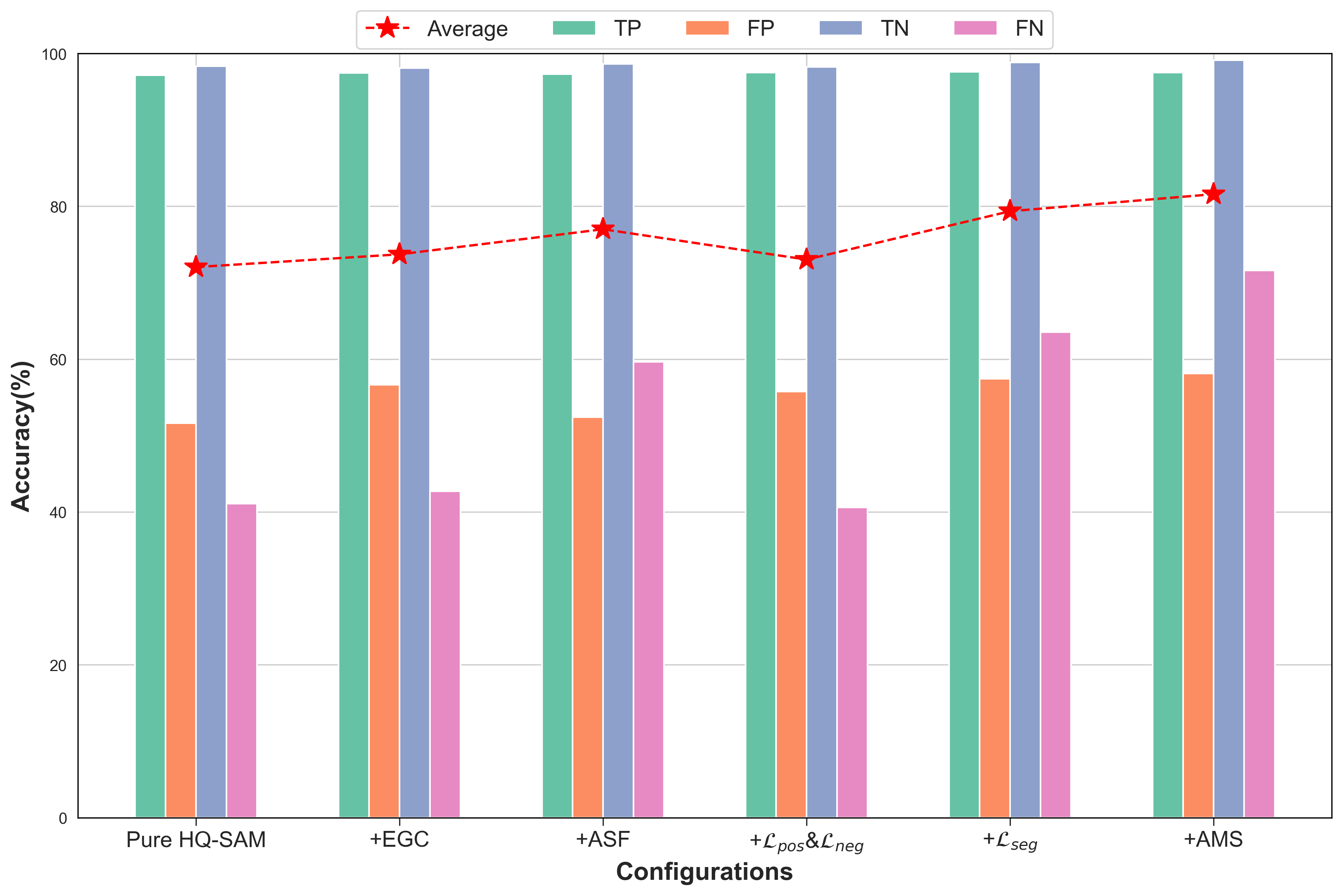}
	}
	\subfloat[IoU on BAQS-raw]{
		\centering
		\includegraphics[width=0.49\linewidth]{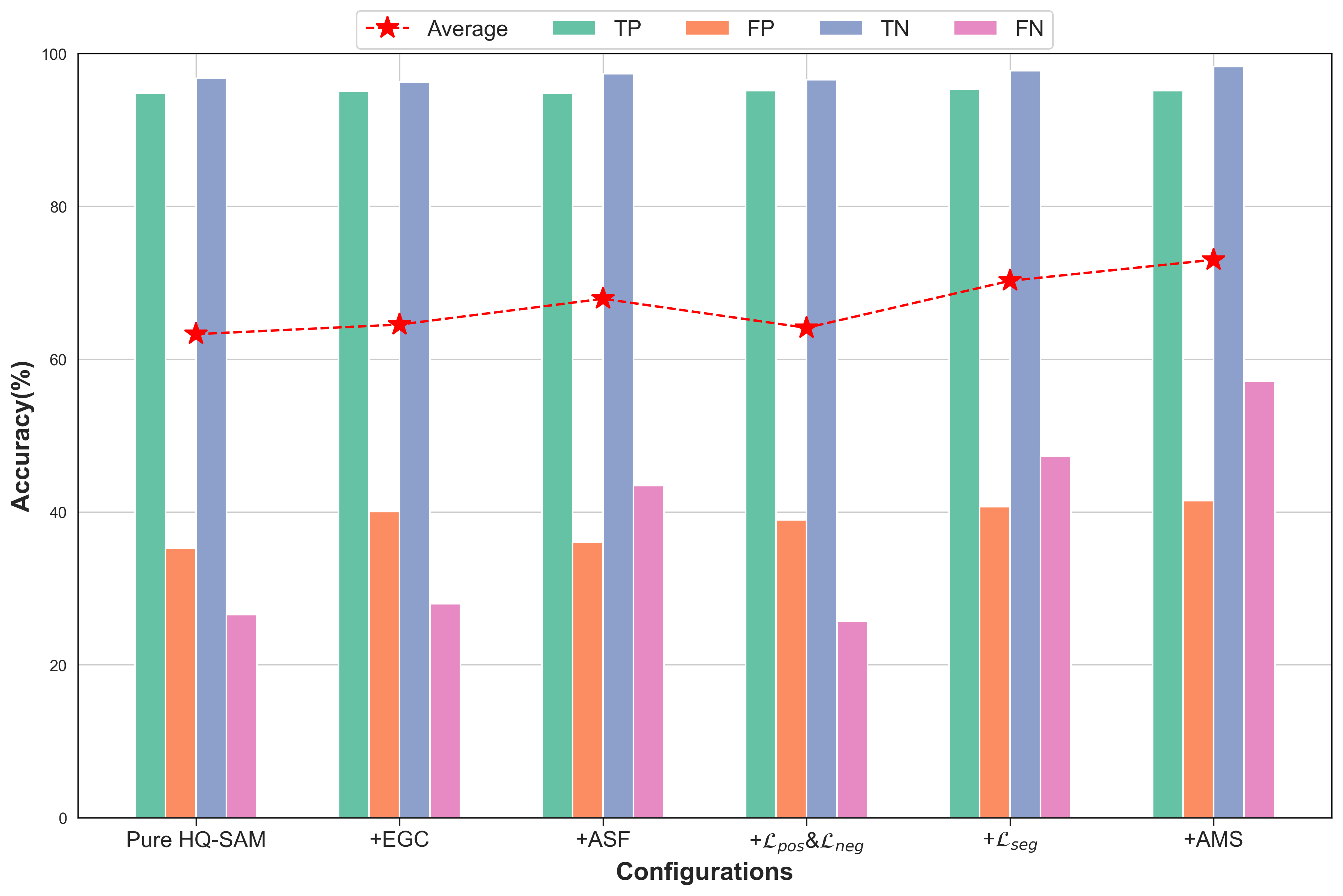}
	}
	
	\caption{F1 and IoU results of ablation studies on the effectiveness of different components. Components (EGC, ASF, $\mathcal{L}_{pos}$\&$\mathcal{L}_{neg}$, $\mathcal{L}_{seg}$ and AMS) are sequentially added are added sequentially to the HQ-SAM baseline (abscissa). Evaluations are shown for (a, b) Train-Val (BAQS-UNetFormer), (c, d) Zero-Shot Model-Source (ZS-MS, BAQS-OCRNet), and (e, f) Zero-Shot Human-Source (ZS-HS, BAQS-raw). Bars represent per-class accuracy (TP/FP/TN/FN); red stars linked by dashed lines show mean scores (mF1/mIoU)}  
	\label{fig:designs}
\end{figure*}

To rigorously validate the effectiveness of the proposed components within SegAssess, we conducted comprehensive ablation experiments. Given the goal of practical, transferable SQA applications, we selected three specific variants of the BAQS dataset for evaluation, each representing a distinct scenario: BAQS-UNetFormer simulates standard model development using a typical training-validation (Train-Val) protocol; BAQS-OCRNet evaluates zero-shot transferability to unseen model-source (ZS-MS) segmentation masks; and BAQS-raw assesses zero-shot transferability to unseen human-source (ZS-HS) masks. This section investigates the impact of individual architectural and training strategy components (Section~\ref{sec:designs}) and the influence of different backbone model sizes (Section~\ref{sec:size}).

 \begin{figure*}[!t]
	\centering
	
	\subfloat{%
		\makebox[0pt][r]{\rotatebox{90}{\scriptsize{\textbf{BAQS-UNetFormer}}}\hspace{3pt}}%
		\includegraphics[width=0.117\linewidth]{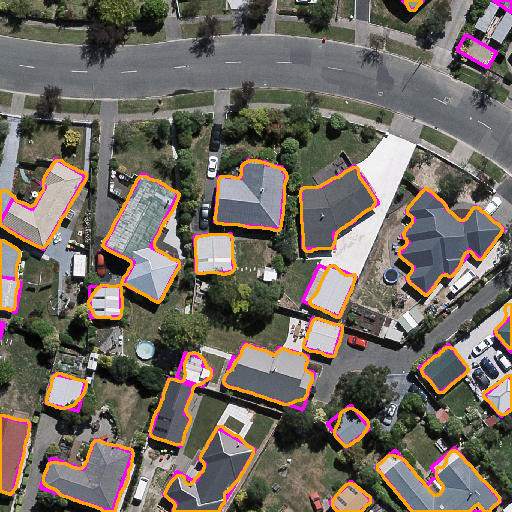}%
	}\hspace{0.3em}%
	\subfloat{%
		\includegraphics[width=0.117\linewidth]{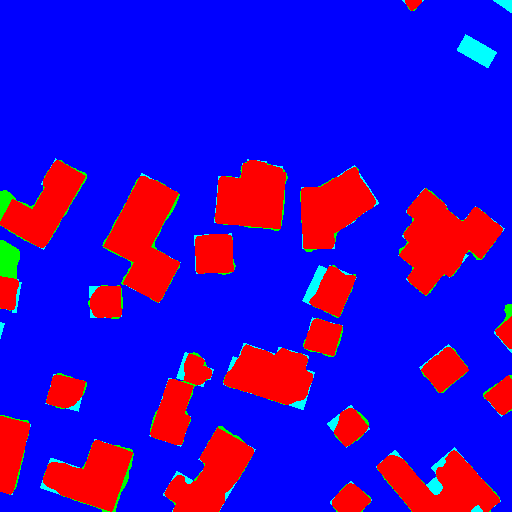}%
	}\hspace{0.3em}%
	\subfloat{%
		\includegraphics[width=0.117\linewidth]{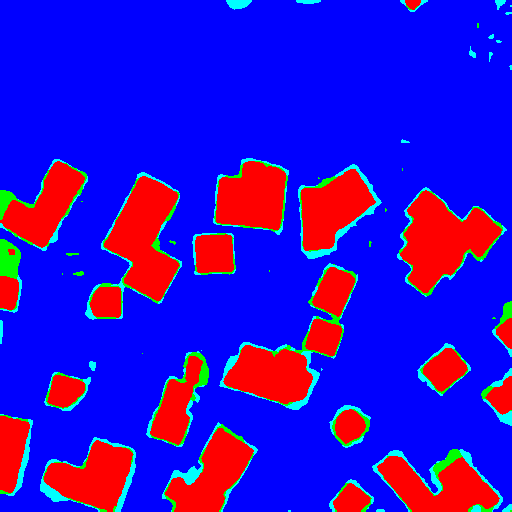}%
	}\hspace{0.3em}%
	\subfloat{%
		\includegraphics[width=0.117\linewidth]{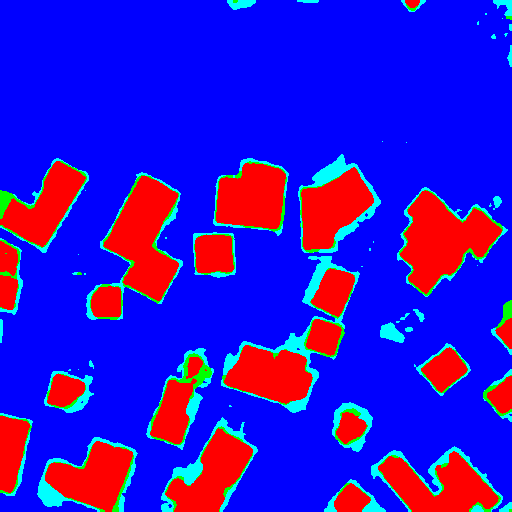}%
	}\hspace{0.3em}%
	\subfloat{%
		\includegraphics[width=0.117\linewidth]{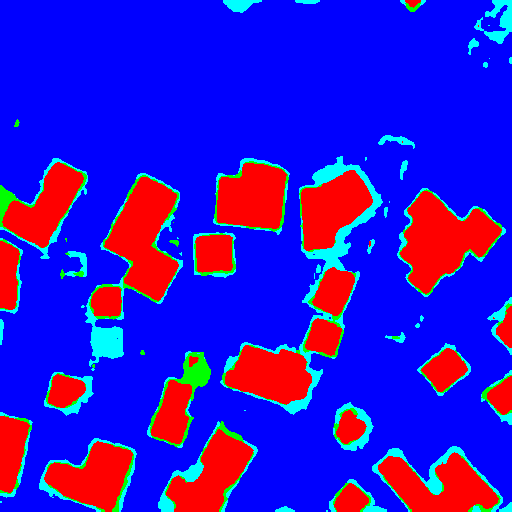}%
	}\hspace{0.3em}%
	\subfloat{%
		\includegraphics[width=0.117\linewidth]{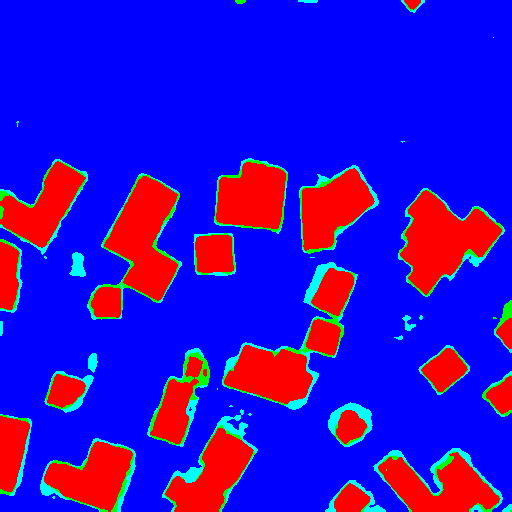}%
	}\hspace{0.3em}%
	\subfloat{%
		\includegraphics[width=0.117\linewidth]{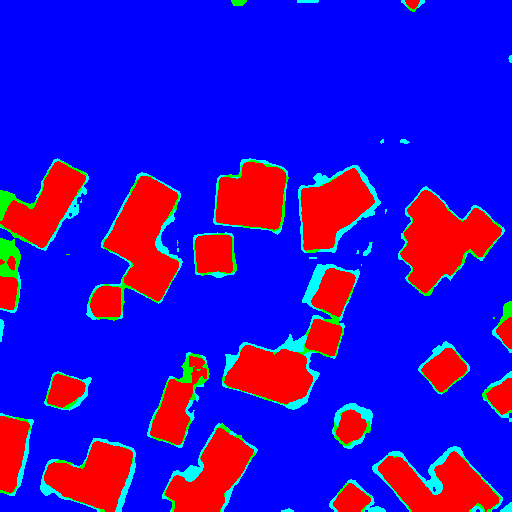}%
	}\hspace{0.3em}%
	\subfloat{%
		\includegraphics[width=0.117\linewidth]{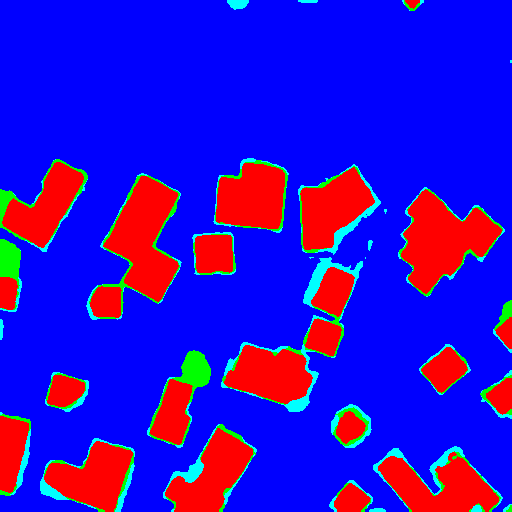}%
	}%
	
	\vspace{-0.7em}
	\setcounter{subfigure}{0}
	
	\subfloat{%
		\makebox[0pt][r]{\rotatebox{90}{\scriptsize{\textbf{~BAQS-OCRNet}}}\hspace{3pt}}%
		\includegraphics[width=0.117\linewidth]{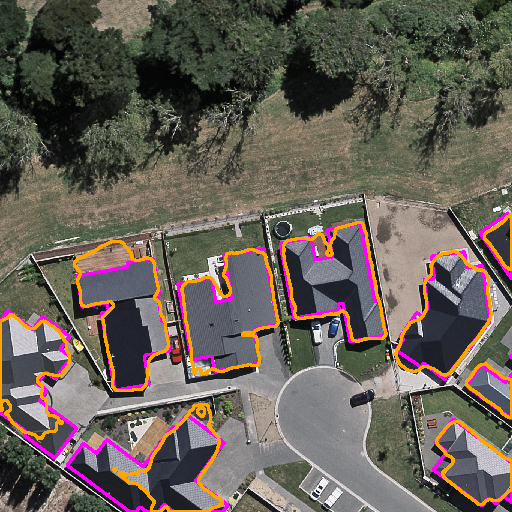}%
	}\hspace{0.3em}%
	\subfloat{%
		\includegraphics[width=0.117\linewidth]{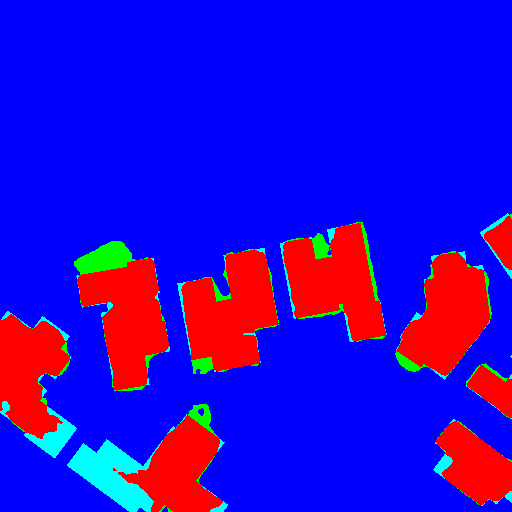}%
	}\hspace{0.3em}%
	\subfloat{%
		\includegraphics[width=0.117\linewidth]{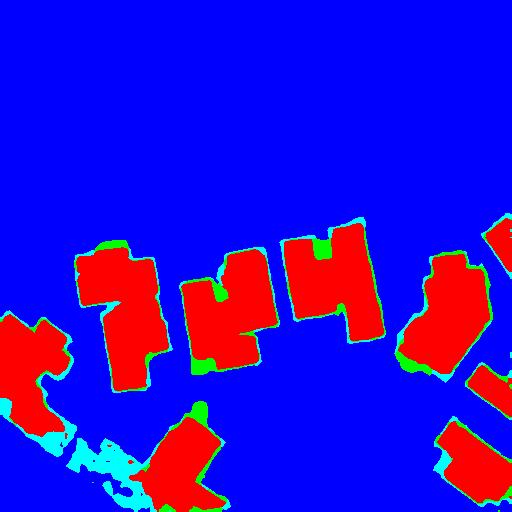}%
	}\hspace{0.3em}%
	\subfloat{%
		\includegraphics[width=0.117\linewidth]{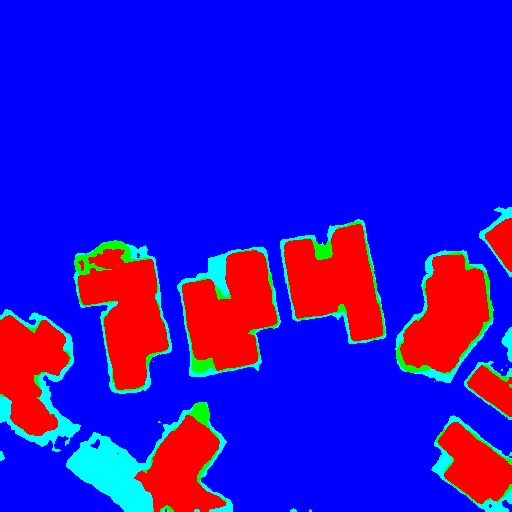}%
	}\hspace{0.3em}%
	\subfloat{%
		\includegraphics[width=0.117\linewidth]{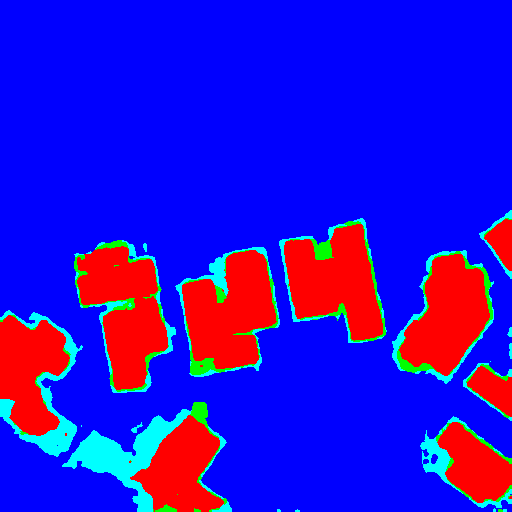}%
	}\hspace{0.3em}%
	\subfloat{%
		\includegraphics[width=0.117\linewidth]{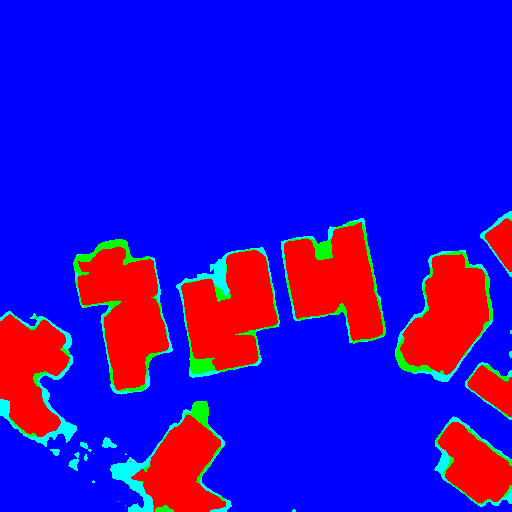}%
	}\hspace{0.3em}%
	\subfloat{%
		\includegraphics[width=0.117\linewidth]{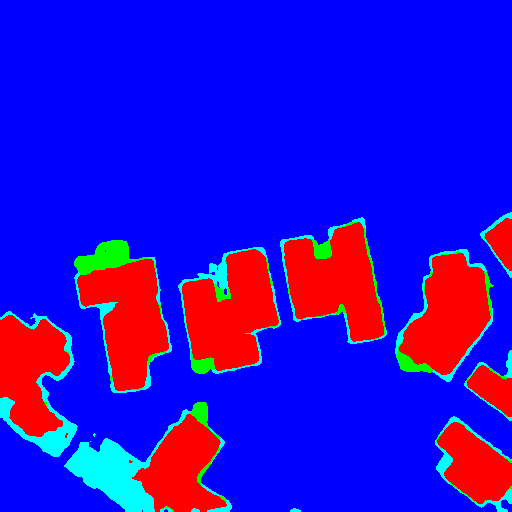}%
	}\hspace{0.3em}%
	\subfloat{%
		\includegraphics[width=0.117\linewidth]{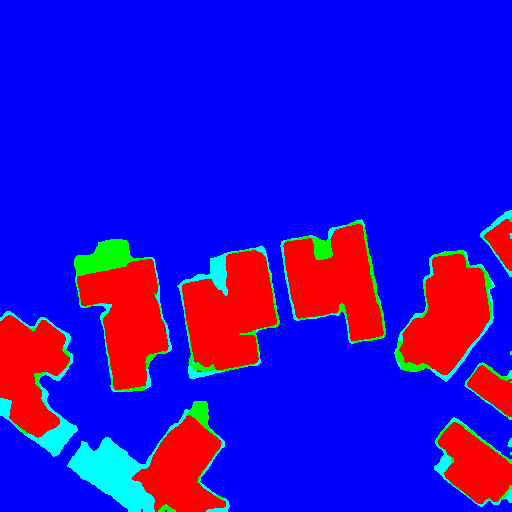}%
	}%
	
	\vspace{-0.7em}
	\setcounter{subfigure}{0}
	
	\subfloat[Image]{%
		\makebox[0pt][r]{\rotatebox{90}{\scriptsize{\textbf{~~~~~BAQS-raw}}}\hspace{3pt}}%
		\includegraphics[width=0.117\linewidth]{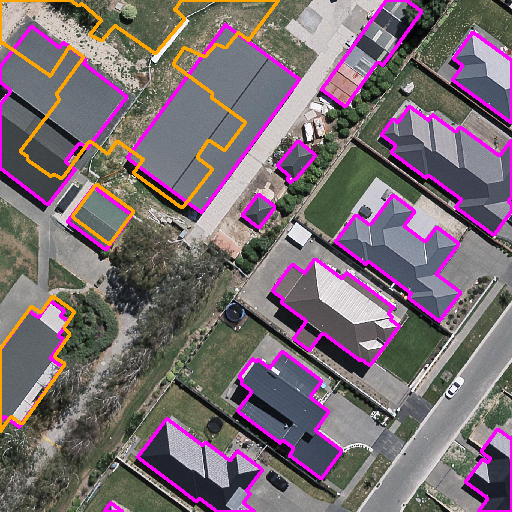}%
	}\hspace{0.3em}%
	\subfloat[GT]{%
		\includegraphics[width=0.117\linewidth]{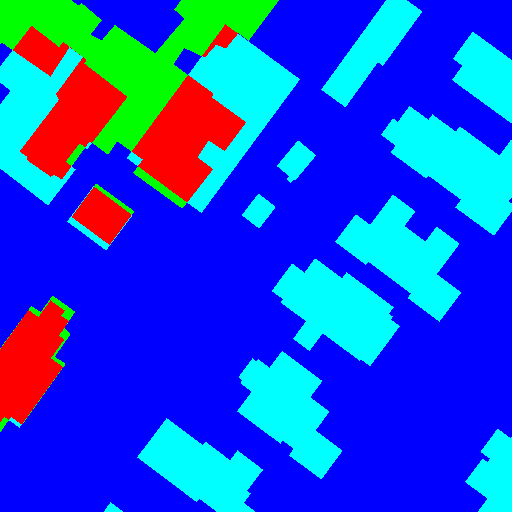}%
	}\hspace{0.3em}%
	\subfloat[HQ-SAM]{%
		\includegraphics[width=0.117\linewidth]{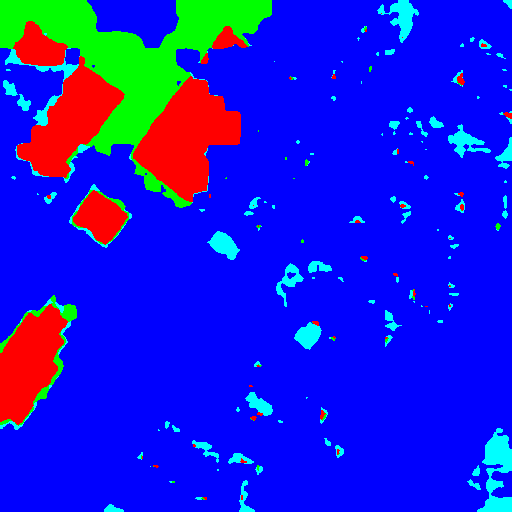}%
	}\hspace{0.3em}%
	\subfloat[+EGC]{%
		\includegraphics[width=0.117\linewidth]{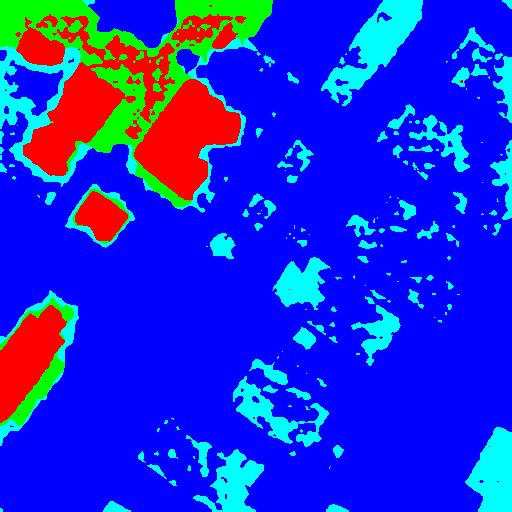}%
	}\hspace{0.3em}%
	\subfloat[+ASF]{%
		\includegraphics[width=0.117\linewidth]{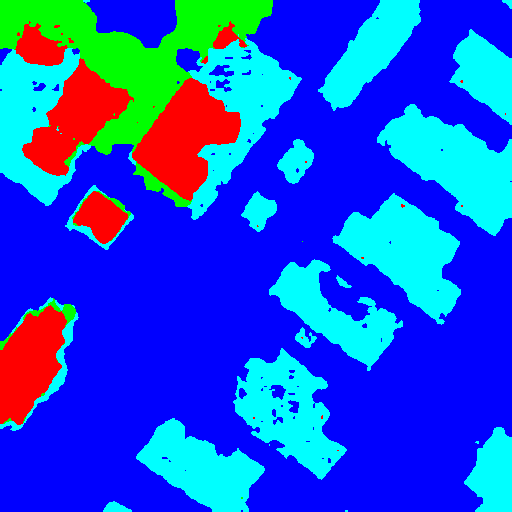}%
	}\hspace{0.3em}%
	\subfloat[+$\mathcal{L}_{pos}$\&$\mathcal{L}_{neg}$]{%
		\includegraphics[width=0.117\linewidth]{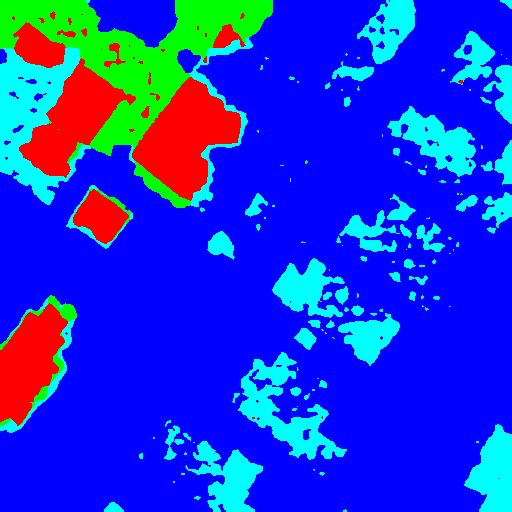}%
	}\hspace{0.3em}%
	\subfloat[+$\mathcal{L}_{seg}$]{%
		\includegraphics[width=0.117\linewidth]{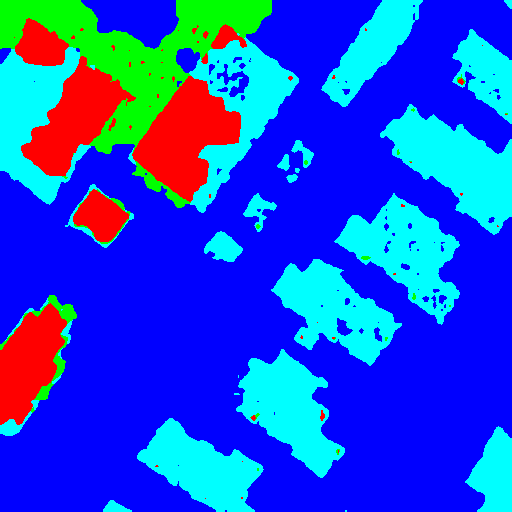}%
	}\hspace{0.3em}%
	\subfloat[+AMS]{%
		\includegraphics[width=0.117\linewidth]{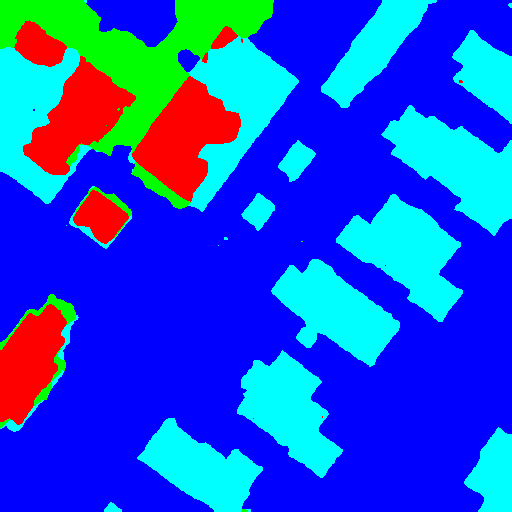}%
	}%

	\caption{Qualitative visual comparisons illustrating the impact of sequentially adding components during the ablation study. (a) Input image with mask under evaluation (\textcolor{orange}{Orange} contour) and ground truth  (\textcolor{magenta}{magenta} contour). (b) Ground truth PQM assessment map. (c)-(h) PQM predictions from SegAssess configurations with progressively added components (EGC, ASF, $\mathcal{L}_{pos}$\&$\mathcal{L}_{neg}$, $\mathcal{L}_{seg}$ and AMS). Colors represent \textcolor{red}{TP}, \textcolor{green}{FP}, \textcolor{blue}{TN} and \textcolor{cyan}{FN}.}  
	\label{fig:vis_ablation}
\end{figure*}

\subsubsection{Effectiveness of different components}\label{sec:designs}
We systematically evaluated the contribution of key designs by incrementally adding them to a baseline HQ-SAM architecture: the Edge Guided Compaction (EGC) branch, the Aggregated Semantic Filter (ASF) module, the reconstruction loss terms \(\mathcal{L}_{pos}\)\&\(\mathcal{L}_{neg}\), followed by \(\mathcal{L}_{seg}\), and finally the Augmented Mixup Sampling (AMS) strategy. Commencing from the baseline HQ-SAM architecture, we incrementally incorporate these components one by one to meticulously observe the resultant performance variations. For configurations without AMS, models were trained solely on the BAQS-UNetFormer dataset. When AMS was included, training utilized a mix of four variants (BAQS-DeepLabv3+, BAQS-HRNet, BAQS-TransUNet, BAQS-UNetFormer) to expose the model to diverse mask sources. All configurations were evaluated on BAQS-UNetFormer (Train-Val), BAQS-OCRNet (ZS-MS), and BAQS-raw (ZS-HS). The quantitative results are presented in Fig.~\ref{fig:designs}, with corresponding qualitative examples in Fig.~\ref{fig:vis_ablation}.

Incorporating the EGC component provides the first level of refinement over the baseline. As shown in Fig.~\ref{fig:designs}, this addition notably improves the identification of false positive (FP) pixels across all three evaluation datasets, with F1$^{FP}$ and IoU$^{FP}$ increases by at least 2.07$\%$ and 2.67$\%$ ,respectively. While the effect on false negative (FN) detection varied slightly across datasets, the overall mean F1 and IoU scores remained stable or showed a slight upward trend, indicating a positive net contribution from explicitly guiding the assessment with edge information.

Adding the ASF module to the EGC branch yields further significant benefits, particularly enhancing the model's transferability to human-sourced masks (ZS-HS scenario on BAQS-raw). On this dataset, ASF dramatically boosted F1$^{FN}$ and IoU$^{FN}$ with +16.95$\%$ and +15.46$\%$ incremental increases compared to the configuration with only EGC (see 3$^{rd}$ row 4$^{th}$ column example in Fig.~\ref{fig:vis_ablation}). This improvement is likely due to ASF's enhanced semantic filtering of high-level features, leading to better discrimination, particularly aiding the correct classification of false negative (FN) pixels in challenging cases.

Three reconstruction loss functions, namely $\mathcal{L}_{pos}$, $\mathcal{L}_{neg}$ and $\mathcal{L}_{seg}$ also play a crucial role in optimizing the model learning process, thus bring improvements in performance across multiple categories and datasets. The addition of $\mathcal{L}_{pos}$ and $\mathcal{L}_{neg}$ benefits almost all metrics but leads to a noticeable drop in F1$^{FN}$ and IoU$^{FN}$ on the BAQS-raw dataset (3$^{rd}$ row 6$^{th}$ column of Fig.~\ref{fig:vis_ablation}). We hypothesize this occurred because the contribution of the extremely rare FN pixels to $\mathcal{L}_{pos}$ might be overshadowed by the contribution of the much more prevalent TP pixels (see Eq.~\ref{Eq:L_pos}). This issue is resolved by subsequently  incorporating $\mathcal{L}_{seg}$ which highlights FP and FN regions and gains improvements in metrics of all categories and datasets (Fig.~\ref{fig:vis_ablation}(g) column).

Finally, integrating the AMS strategy during training provided a transformative improvement. With AMS enabled, all metrics—including class-specific F1/IoU and overall mF1/mIoU—reached their peak values across all three evaluation scenarios (Train-Val, ZS-MS, ZS-HS), as clearly shown in Fig.~\ref{fig:designs} and Fig.~\ref{fig:vis_ablation}. This substantial gain underscores the effectiveness of AMS in enriching training data diversity by exposing the model to masks generated by various architectures. This forces SegAssess to learn features robust to different segmentation characteristics, greatly enhancing both its overall effectiveness and its crucial generalization ability to unseen mask sources.

In conclusion, each of these components-EGC, ASF, $\mathcal{L}_{pos} \& \mathcal{L}_{neg}$, $\mathcal{L}_{seg}$, and AMS-makes a distinct and valuable contribution to the model's performance in SQA task. Their synergistic integration enables SegAssess to perform the PQM task with high accuracy and excellent zero-shot transferability.

\begin{table*}[!t]
	\centering
	\tiny
	\caption{Impacts of different backbone sizes. FZ means freezing image encoder during training. This table compares configurations using SAM-B (FZ: frozen encoder, EA: adapter tuning, Base: fully fine-tuned) and fully fine-tuned SAM-L (Large) and SAM-H (Huge). Metrics include trainable parameters (\#Param.), inference speed (FPS), and accuracy (mF1/mIoU) on Train-Val, ZS-MS, and ZS-HS scenarios.}
	\resizebox{\textwidth}{!}
	{
			\renewcommand\arraystretch{1.5}
					\begin{tabular}{ll|cc|cc|cc|cc}
							\hline			
							\multirow{2}{*}{Version}&\multirow{2}{*}{Backbone}&\multirow{2}{*}{\#Param.} &\multirow{2}{*}{FPS}&\multicolumn{2}{c|}{BAQS-UNetFormer (Train-Val)}&\multicolumn{2}{c|}{BAQS-OCRNet (ZS-MS)}&\multicolumn{2}{c}{BAQS-raw (ZS-HS)}\\
							&&&&mF1($\%$)&mIoU($\%$)&mF1($\%$)&mIoU($\%$)&mF1($\%$)&mIoU($\%$)\\
							\hline
							
							SegAssess-FZ&SAM-FZ &30.10 M&32.13&64.72&57.53&65.59&58.13&66.11&58.31\\
							SegAssess-EA&SAM-EA &40.96 M&29.57&65.58&58.20&67.41&59.61&71.83&63.06\\
							SegAssess-Base&SAM-B &127.78 M&31.91&70.38&62.31&73.47&64.82&81.59&73.01\\
							SegAssess-Large&SAM-L &371.95 M&18.81&66.72&59.11&68.10&60.18&69.88&61.45\\
							SegAssess-Huge&SAM-H &733.48 M&11.62&71.55&63.15&73.12&64.53&78.31&69.67\\
							\hline
							
						\end{tabular}
				}
			
			\label{table:size}
\end{table*}
	
\subsubsection{Influences of different backbone sizes}\label{sec:size}
We also analyze the model performance and efficiency with difference backbone sizes. The original SAM and HQ-SAM offers different scales, primarily SAM-B, SAM-L, and SAM-H, based on ViT-Base, ViT-Large, and ViT-Huge image encoders, respectively. Common fine-tuning approaches include freezing the pre-trained image encoder entirely (SAM-FZ) or using parameter-efficient adapter modules like SAM-Adapter \cite{chen2023sam} to adapt the frozen encoder (SAM-EA).  to inject new domain knowledge to frozen original encoder weights (SAM-EA). Therefore, we compared five SegAssess configurations: SegAssess-FZ (SAM-B backbone, frozen encoder), SegAssess-EA (SAM-B backbone, frozen encoder with adapter tuning), SegAssess-Base (SAM-B backbone, fully fine-tuned), SegAssess-Large (SAM-L backbone, fully fine-tuned), and SegAssess-Huge (SAM-H backbone, fully fine-tuned). We measured performance using the total number of trainable model parameters (\#Param.) for complexity, frame per second (FPS) for inference speed, and mF1/mIoU for accuracy, evaluated again on the BAQS-UNetFormer (Train-Val), BAQS-OCRNet (ZS-MS), and BAQS-raw (ZS-HS) datasets.

As illustrated in Table.~\ref{table:size}, SegAssess-FZ reports the smallest model size and highest inference speed, at the cost of the worst accuracy scores. For fine-tuning towards current RS task, SegAssess-EA employs MLP to inject new domain knowledge to the outcomes of each Transformer layer in image encoder. It gains improvements in F1 and IoU across all datasets with the expense of 40.96M \#Param. increase in model scale and average 2.56 FPS decrease in running speed. A significant performance leap occurred with SegAssess-Base, where fully fine-tuning all parameters of the SAM-B backbone, resulting in remarkable accuracy gains with at least 4.80$\%$ mF1 and 4.11$\%$ mIoU increases. Though the total parameter size jumps from 40.96 M to 127.78 M, SegAssess-Base version report 2.34 FPS accelerated inference speed comparing to SegAssess-EA. 

However, further increasing the backbone size to SAM-L (SegAssess-Large) and SAM-H (SegAssess-Huge) did not lead to consistent performance improvements and demonstrated diminishing returns. While these larger models outperformed the FZ and EA versions, they were generally surpassed by the smaller SegAssess-Base in the crucial zero-shot transfer scenarios (BAQS-OCRNet and BAQS-raw). Although SegAssess-Huge achieved slightly higher accuracy than SegAssess-Base on the Train-Val dataset (BAQS-UNetFormer), this came at the cost of a drastically reduced inference speed (approximately one-third of SegAssess-Base). This lack of scaling benefit for the larger models might suggest that the available training data volume, while substantial, may not be sufficient to fully leverage the capacity of the ViT-Large and ViT-Huge backbones for this specific PQM task. 

Considering these results, the SegAssess-Base configuration, utilizing a fully fine-tuned SAM-B backbone, achieves the best overall balance between high accuracy, robust zero-shot transferability, model complexity, and inference speed. Therefore, we adopt this configuration for subsequent experiments.
  
\begin{table*}[t]
	\centering
	\caption{Quantitative comparison for the standard Training-Validation protocol. This table compares SegAssess (trained with AMS using 4 model types) against AQSNet (trained conventionally) on the 24 model-source SQA datasets (X-Y, where Y $\in$ {DeepLabv3+, HRNet, TransUNet, UNetFormer}). SegAssess reports metrics for all 4 PQM classes (TP/FP/TN/FN), while AQSNet reports only FP/FN metrics (highlighted in grey for direct comparison). - indicates scores below 1e-4.}
	\resizebox{\textwidth}{!}{
		\renewcommand\arraystretch{1.5}
		\begin{tabular}{l 
				>{\columncolor{gray!20}}c   
				>{\columncolor{gray!40}}c   
				>{\columncolor{gray!20}}c   
				>{\columncolor{gray!40}}c   
				c c 
				>{\columncolor{gray!20}}c   
				>{\columncolor{gray!40}}c   
				c c 
				>{\columncolor{gray!20}}c   
				>{\columncolor{gray!40}}c   
				c c }
			\toprule
			\multirow{2}{*}{Dataset} & \multicolumn{4}{c}{AQSNet (Train-Val)} & \multicolumn{10}{c}{SegAssess (Train-Val)} \\
			\cmidrule(lr){2-5} \cmidrule(lr){6-15}
			& F1$^{FP}$ & IoU$^{FP}$ & F1$^{FN}$ & IoU$^{FN}$ & F1$^{TP}$ & IoU$^{TP}$ & F1$^{FP}$ & IoU$^{FP}$ & F1$^{TN}$ & IoU$^{TN}$ & F1$^{FN}$ & IoU$^{FN}$ & mF1 & mIoU \\
			\midrule
			Inria-DeepLabv3+ &25.05&14.38&20.89&11.72&91.91&85.08&42.92&27.82&97.48&95.10&38.36&23.95&67.68&57.99\\
			Inria-HRNet &27.44&15.94&24.39&13.96&91.76&84.82&41.67&26.62&97.48&95.10&39.14&24.50&67.51&57.76\\
			Inria-TransUNet &27.01&15.67&24.99&14.33&92.08&85.37&42.91&27.51&97.41&94.97&39.92&25.11&68.08&58.24\\
			Inria-UNetFormer &26.68&15.46&25.07&14.38&92.31&85.77&46.60&30.51&97.45&95.05&44.26&28.59&70.16&59.98\\
			CrowdAI-DeepLabv3+ &28.30&16.50&28.20&16.43&95.07&90.63&43.57&27.95&98.22&96.50&48.51&32.20&71.34&61.82\\
			CrowdAI-HRNet &27.92&16.60&62.55&45.97&95.32&90.82&54.49&37.79&98.06&96.19&69.78&53.87&79.38&69.67\\
			CrowdAI-TransUNet &31.22&18.52&25.44&14.63&94.99&90.48&48.60&32.25&98.21&96.50&52.83&36.13&73.66&63.84\\
			CrowdAI-UNetFormer &43.76&28.50&57.63&40.93&94.85&90.22&60.57&43.74&97.98&96.04&69.23&53.23&80.66&70.81\\
			DeepGlobe-DeepLabv3+ &24.43&14.01&24.35&13.99&80.06&67.02&46.25&30.27&99.02&98.06&38.91&24.44&66.06 &54.95\\
			DeepGlobe-HRNet &26.40&15.29&21.17&11.93&79.79&66.61&48.54&32.36&99.04&98.11&34.71&21.19&65.52&54.54\\
			DeepGlobe-TransUNet &33.56&20.28&21.21&11.93&79.26&65.89&49.43&33.01&99.03&98.09&31.35&18.74&64.77&53.93\\
			DeepGlobe-UNetFormer &42.97&27.53&16.83&9.25&78.46&64.83&56.56&39.63&99.05&98.12&36.27&22.44&67.58&56.25 \\
			GID-DeepLabv3+ &2.40&1.60&2.33&1.56&83.80&78.71&40.36&27.52&99.26&98.54&27.80&17.87&62.80&55.66\\
			GID-HRNet &0.89&0.45&0.93&0.47&83.03&77.85&39.28&27.79&99.23&98.48&31.53&20.93&63.17&56.29\\
			GID-TransUNet &0.70&0.36&1.76&0.91&82.76&78.10&35.27&24.56&99.22&98.46&37.47&25.63&63.68&56.69\\
			GID-UNetFormer &2.13&1.42&2.13&1.42&83.06&78.42&41.31&29.37&99.14&98.30&33.91&22.44&64.35&57.12\\
			BAQS-DeepLabv3+ &28.63&16.76&24.54&14.02 &95.95&93.22&38.55&24.31&99.15&98.32&35.44&22.70&67.27&59.64\\
			BAQS-HRNet &0.81&0.41&1.03&0.52 &95.81&92.96&35.56&22.04&99.23&98.48&31.21&19.28&65.45&58.19 \\
			BAQS-TransUNet &28.42&16.62&27.34&15.91&96.05&93.41&36.66&23.05&99.20&98.41&34.03&21.02&66.49&59.97\\
			BAQS-UNetFormer &-&-&-&- &97.72&95.59&42.46&27.00&99.23&98.48&42.92&28.20&70.58&62.31\\
			WAQS-DeepLabv3+ &40.88&25.82&15.88&8.67&93.92&88.71&52.30&35.72&99.21&98.43&31.01&18.76&69.11&60.41\\
			WAQS-HRNet &0.95&0.48&0.95&0.48&93.85&88.60&38.97&24.66&99.19&98.39&28.13&16.58&65.04&57.06\\
			WAQS-TransUNet &1.16&0.59&0.87&0.44 &93.60&88.16&44.94&29.35&99.21&98.44&26.86&15.88&66.15&57.96\\
			WAQS-UNetFormer &-&-&-&- &93.81&88.51&43.87&28.46&99.20&98.41&28.89&17.17&66.44&58.14\\
			
			\bottomrule
	\end{tabular}}
	\label{Tab: train_val}
	\hspace{-1.8em}
\end{table*}

\subsection{Performance Comparison with the State-of-the-Art PEL methods}\label{train_val1}
\begin{figure*}[!t]
	\centering
	\vspace{-0.8em}
	\subfloat{%
		\makebox[0pt][r]{\rotatebox{90}{\scriptsize{\textbf{~~~~~CrowdAI-DeepLabv3+}}}\hspace{3pt}}%
		\includegraphics[width=0.2\linewidth]{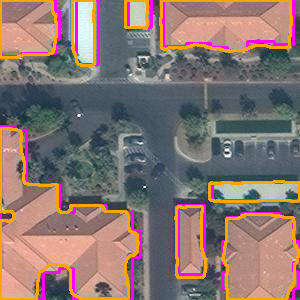}%
	}
	\subfloat{%
		\includegraphics[width=0.2\linewidth]{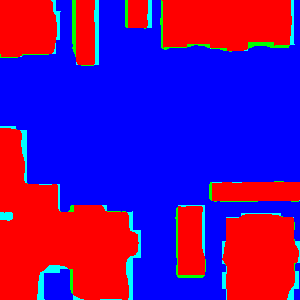}%
	}
	\subfloat{%
		\includegraphics[width=0.2\linewidth]{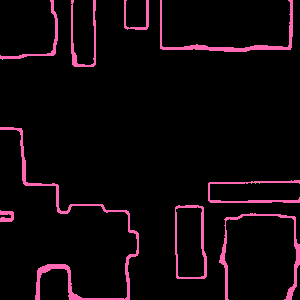}%
	}
	\subfloat{%
		\includegraphics[width=0.2\linewidth]{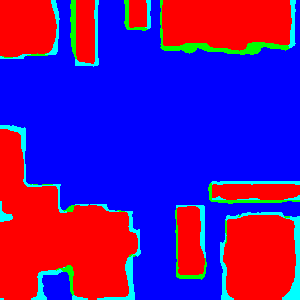}%
	}
	
	\vspace{-1.0em}
	\setcounter{subfigure}{0}
	
	\subfloat{%
		\makebox[0pt][r]{\rotatebox{90}{\scriptsize{\textbf{~~~~~~CrowdAI-UNetFormer}}}\hspace{3pt}}%
		\includegraphics[width=0.2\linewidth]{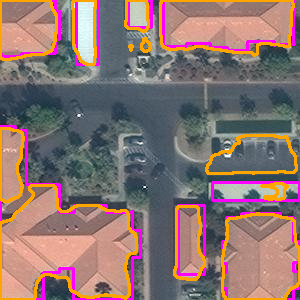}%
	} 
	\subfloat{%
		\includegraphics[width=0.2\linewidth]{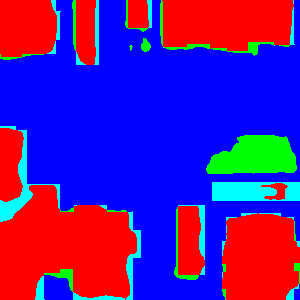}%
	}
	\subfloat{%
		\includegraphics[width=0.2\linewidth]{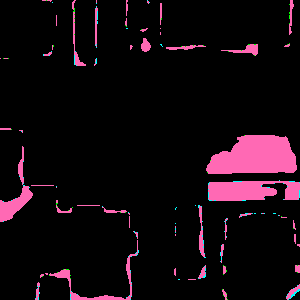}%
	}
	\subfloat{%
		\includegraphics[width=0.2\linewidth]{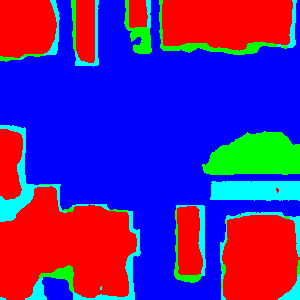}%
	}
	
	\vspace{-1.0em}
	\setcounter{subfigure}{0}
	
	\subfloat{%
		\makebox[0pt][r]{\rotatebox{90}{\scriptsize{\textbf{~~~~DeepGlobe-DeepLabv3+}}}\hspace{3pt}}%
		\includegraphics[width=0.2\linewidth]{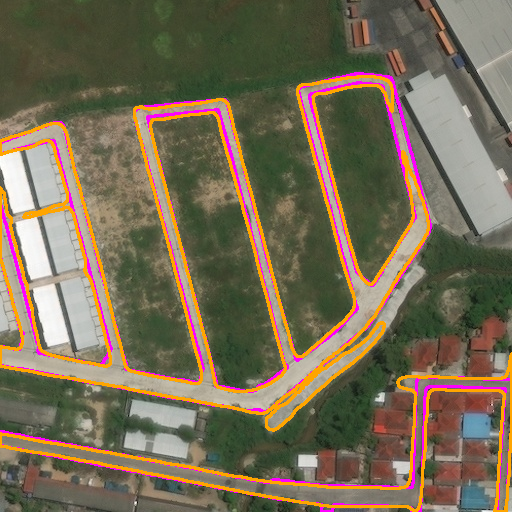}%
	} 
	\subfloat{%
		\includegraphics[width=0.2\linewidth]{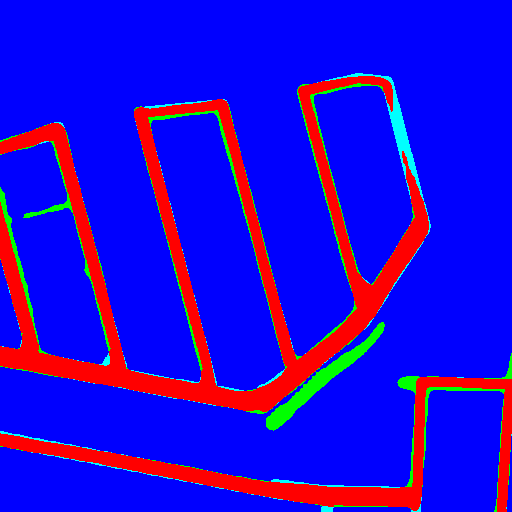}%
	}
	\subfloat{%
		\includegraphics[width=0.2\linewidth]{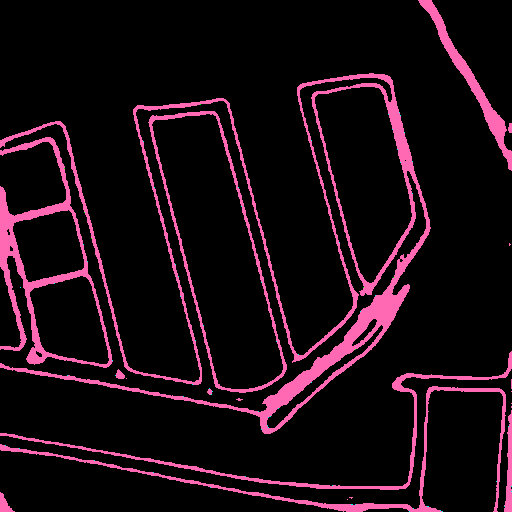}%
	}
	\subfloat{%
		\includegraphics[width=0.2\linewidth]{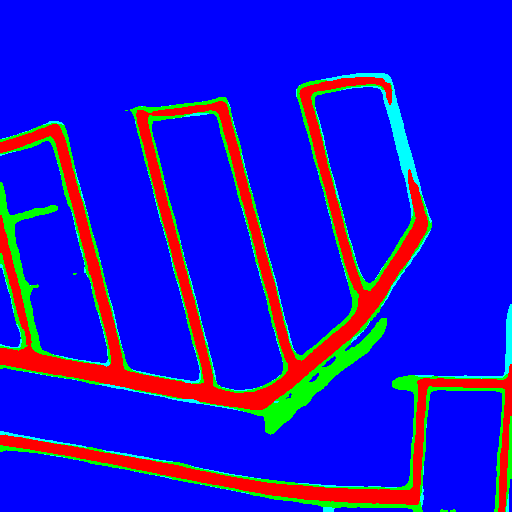}%
	}
	
	\vspace{-1.0em}
	\setcounter{subfigure}{0}
	
	\subfloat{%
		\makebox[0pt][r]{\rotatebox{90}{\scriptsize{\textbf{~~~~DeepGlobe-UNetFormer}}}\hspace{3pt}}%
		\includegraphics[width=0.2\linewidth]{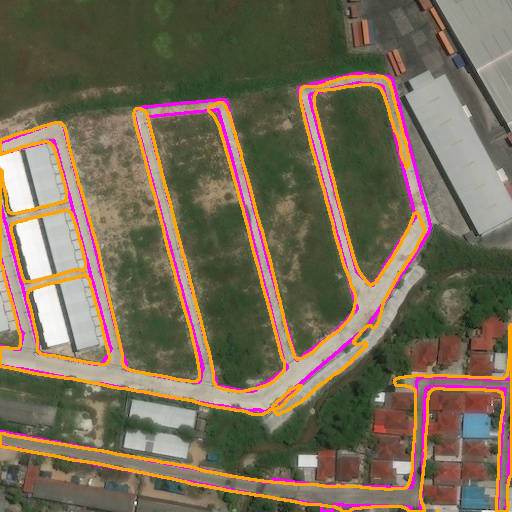}%
	} 
	\subfloat{%
		\includegraphics[width=0.2\linewidth]{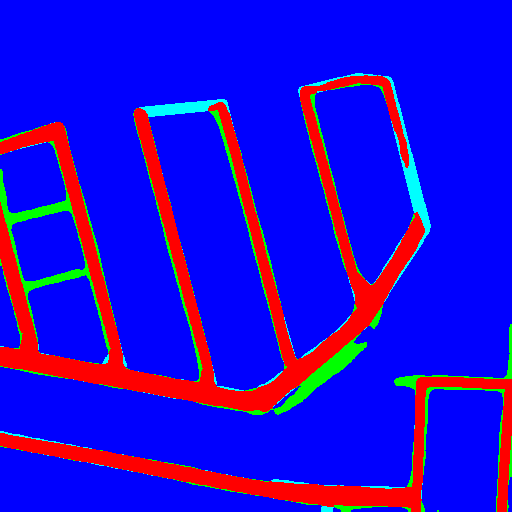}%
	}
	\subfloat{%
		\includegraphics[width=0.2\linewidth]{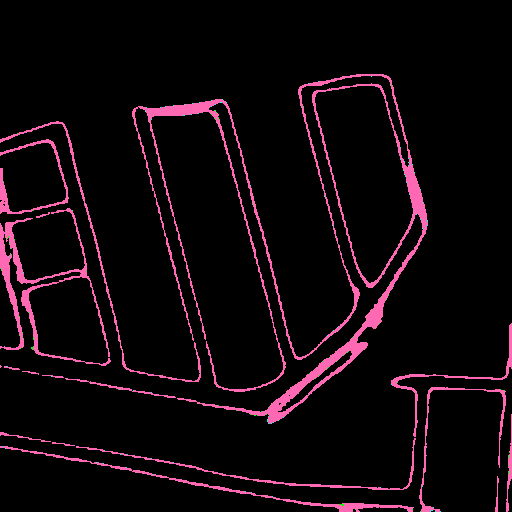}%
	}
	\subfloat{%
		\includegraphics[width=0.2\linewidth]{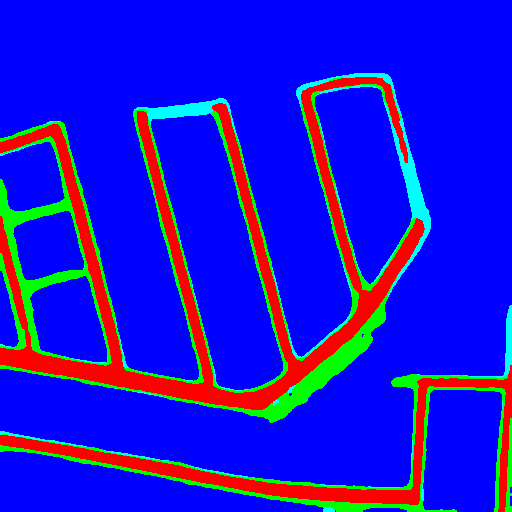}%
	}
	
	\vspace{-1.0em}
	\setcounter{subfigure}{0}
	
	\subfloat{%
		\makebox[0pt][r]{\rotatebox{90}{\scriptsize{\textbf{~~~~~~~WAQS-DeepLabv3+}}}\hspace{3pt}}%
		\includegraphics[width=0.2\linewidth]{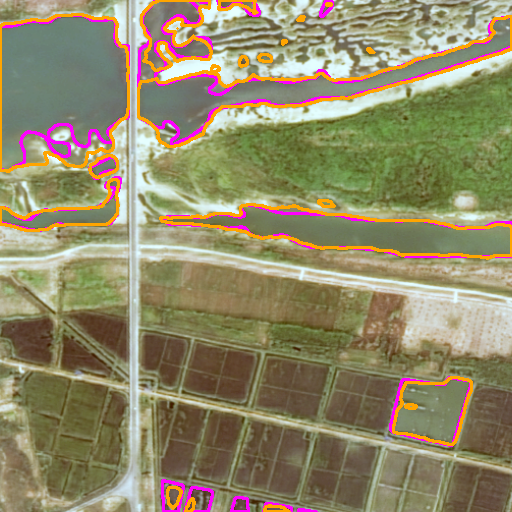}%
	} 
	\subfloat{%
		\includegraphics[width=0.2\linewidth]{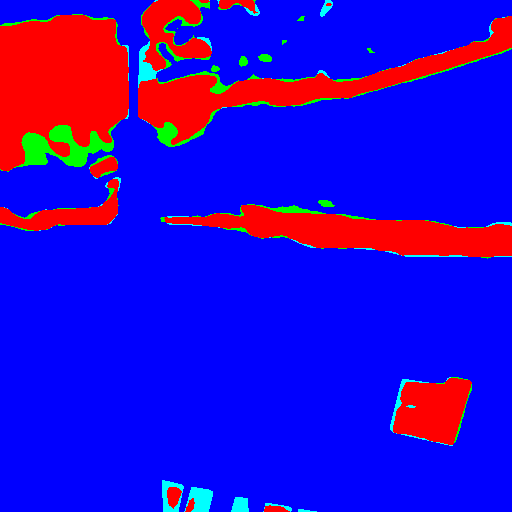}%
	}
	\subfloat{%
		\includegraphics[width=0.2\linewidth]{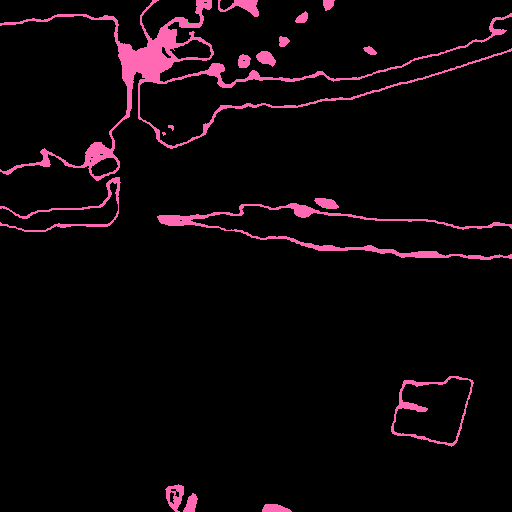}%
	}
	\subfloat{%
		\includegraphics[width=0.2\linewidth]{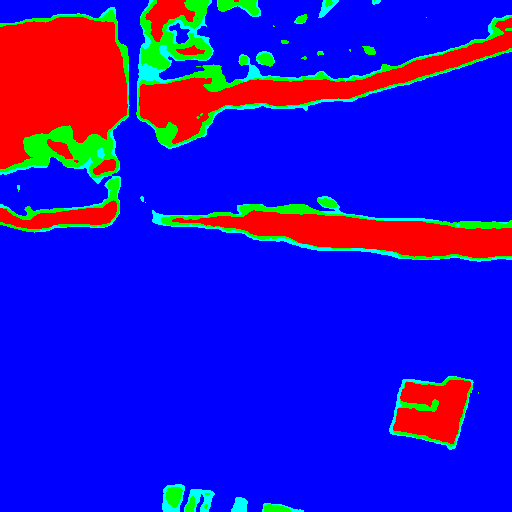}%
	}
	
	\vspace{-1.0em}
	\setcounter{subfigure}{0}
	
	\subfloat{%
		\makebox[0pt][r]{\rotatebox{90}{\scriptsize{\textbf{~~~~~~~~WAQS-UNetFormer}}}\hspace{3pt}}%
		\includegraphics[width=0.2\linewidth]{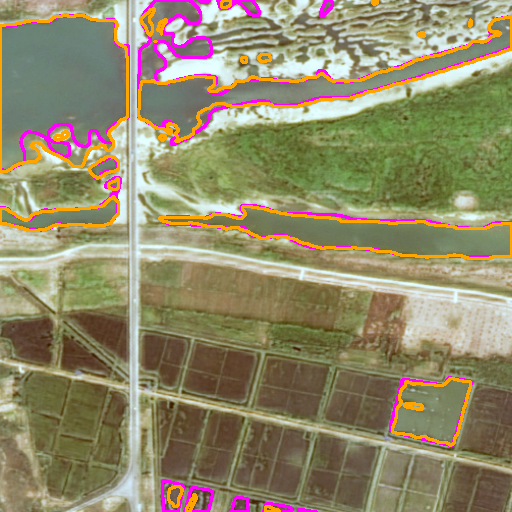}%
	} 
	\subfloat{%
		\includegraphics[width=0.2\linewidth]{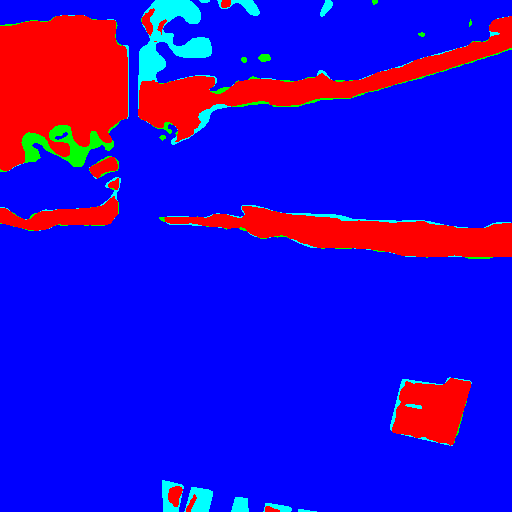}%
	}
	\subfloat{%
		\includegraphics[width=0.2\linewidth]{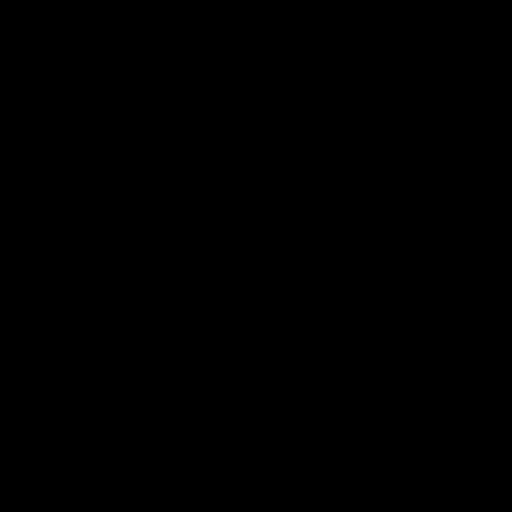}%
	}
	\subfloat{%
		\includegraphics[width=0.2\linewidth]{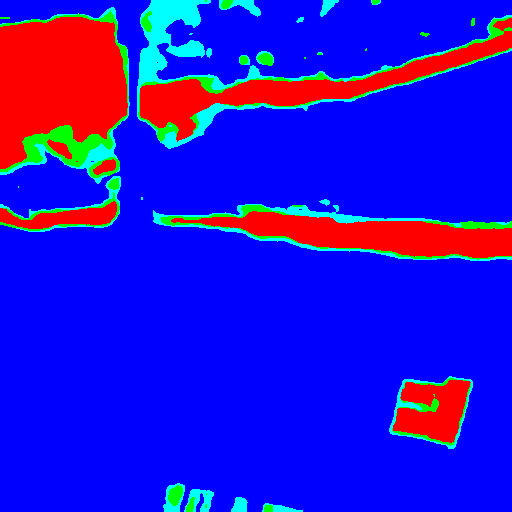}%
	}
	
	\caption{Visual comparison of SegAssess and AQSNet performance under the standard Train-Val protocol. (Left column: Image with \textcolor{orange}{mask under evaluation}/\textcolor{magenta}{GT} contours. Middle columns: GT PQM, AQSNet FP/FN prediction. Right column: SegAssess 4-class PQM prediction). SegAssess provides a comprehensive map without the FP/FN overlap (\textcolor{pink}{pink} areas) seen in AQSNet. Colors represent \textcolor{red}{TP}, \textcolor{green}{FP}, \textcolor{blue}{TN} and \textcolor{cyan}{FN}.}  
	\label{fig:compare}
\end{figure*}

\begin{figure*}[!t]
	\centering
	\vspace{-1.8em}
	\subfloat{%
		\includegraphics[width=0.49\linewidth]{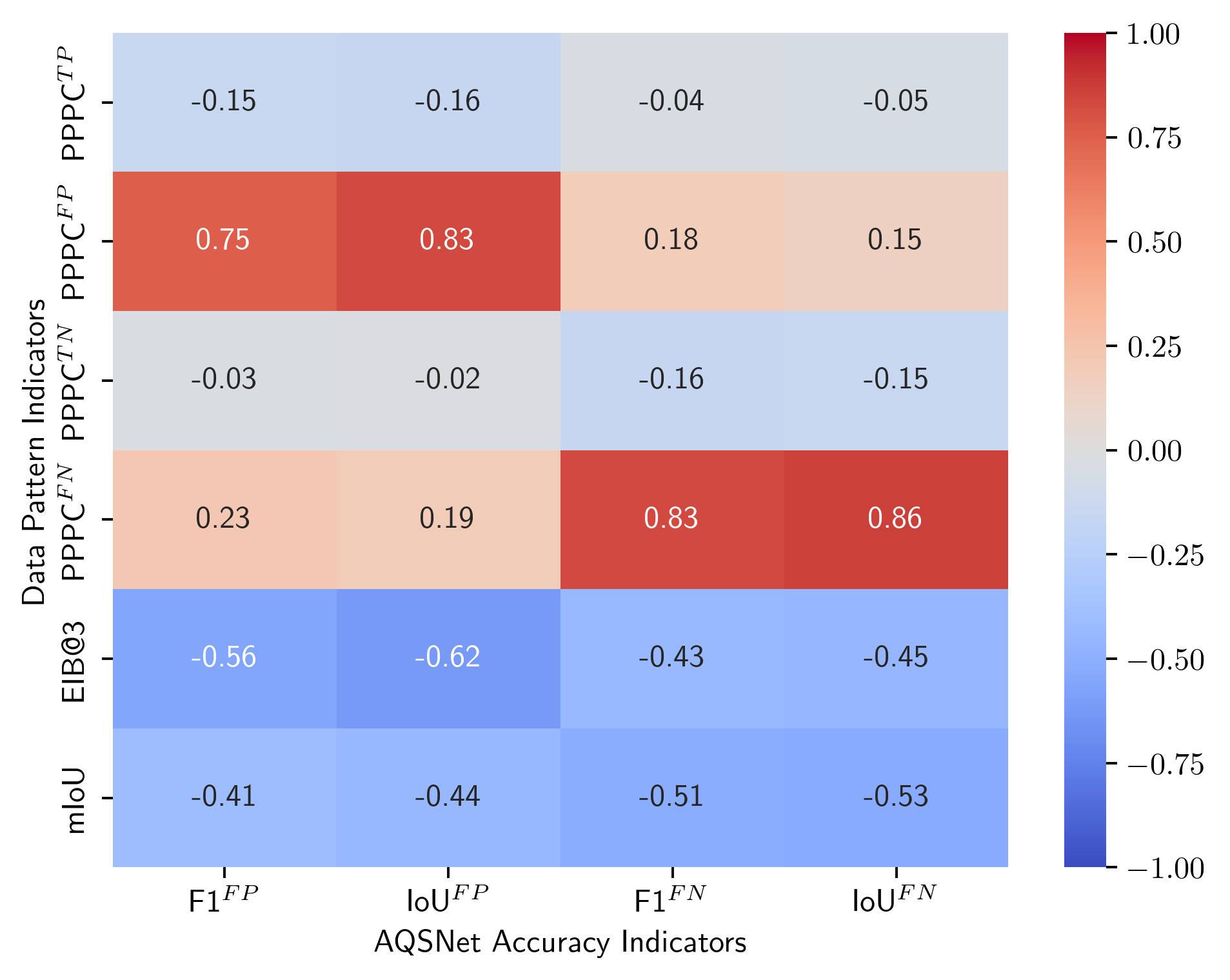}%
	}
	\subfloat{%
		\includegraphics[width=0.49\linewidth]{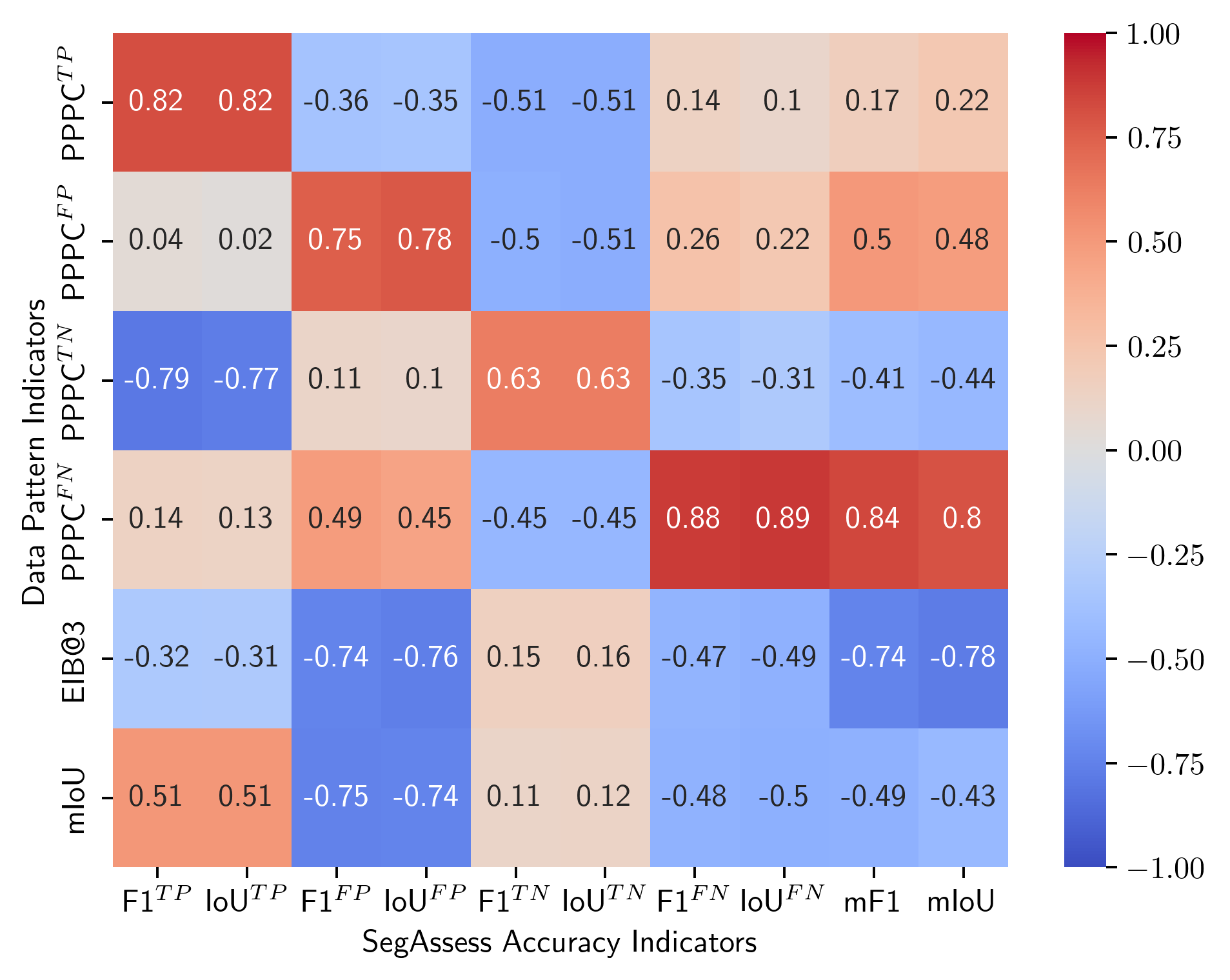}%
	}\
	
	\caption{Pearson correlation analysis between dataset statistics and model performance. Heatmaps show correlations for AQSNet (left) and SegAssess (right), considering data characteristics (e.g., class percentages, EIB@3, input mask mIoU) and accuracy metrics (per-class F1/IoU). Calculations were based on results from Table\ref{Tab: train_val} and Table \ref{Tab: zero_shot}.}  
	\label{fig:correlation}
\end{figure*}

To validate the fundamental effectiveness of our proposed Panoramic Quality Mapping (PQM) paradigm, it is imperative to benchmark it against the state-of-the-art Pixel-level Error Localization (PEL) paradigm. Given that SegAssess implements the four-class Panoramic Quality Mapping (PQM) paradigm (TP/FP/TN/FN), simpler binary error localization methods, such as \cite{zaman2023segmentation}, are not directly comparable due to the incompatibility in both label space and evaluation granularity. Among priors, ASQNet \cite{zhang2023automatic}, which distinguishes FP and FN errors, represents the closest benchmark, although it lacks the comprehensive assessment capabilities. Therefore, in this section, we conduct extensive experiments to systematically evaluates SegAssess through: (1) intra-dataset performance analysis with F1 and IoU scores following the standard training-validation protocol on 24 benchmark datasets (described in Section.~\ref{data}), i.e. X-DeepLabv3+, X-HRNet, X-TransUNet, X-UNetFormer, and (2) comparison against ASQNet under identical experimental conditions.

The quantitative results, presented in Table ~\ref{Tab: train_val}, demonstrate SegAssess's prominent superiority over AQSNet in standard Train-Val accuracy evaluations. On the whole, SegAssess not only enables finer-grained assessment categorization with TP, FP, TN and FN classes, but also consistently achieves stable and reliable performance across diverse mask generation models, image sources (base datasets), and object types, with mF1 and mIoU scores consistently above 62.80$\%$ and 53.93$\%$. This stability contrasts sharply with AQSNet, which exhibited fragility and failed to produce meaningful results (indicated by near-zero FP/FN scores) on over a third of the datasets (9 out of 24), particularly struggling with GID and several BAQS/WAQS variants. This suggests that SegAssess's PQM approach, combined with design elements like AMS, leads to a more robust assessment framework.  

From the perspective of assessment classificatory, SegAssess effectively handles all four PQM categories. As expected due to their prevalence, the accuracy of TN class ranks first against others, and all F1$^{TN}$ and IoU$^{TN}$ impressively exceed 95$\%$. Another correctness category, true positive (TP), obtains the second best metrics scores with at least 78.46$\%$ F1$^{TP}$ and 64.83$\%$ IoU$^{TP}$. Crucially, SegAssess also significantly outperforms AQSNet in localizing the error categories. F1$^{FP}$ and IoU$^{FP}$ scores for show improvements of at least +8.24 $\%$ and +7.55 $\%$, respectively, over AQSNet across all datasets. For FN pixel identification, SegAssess keeps surpassing AQSNet with respect to all F1$^{FN}$ and IoU$^{FN}$ evaluations. For both SegAssess and AQSNet, localizing FN pixels is more difficult than other classes due to its typical rarity. 

Further analysis using Pearson correlation (Fig.~\ref{fig:correlation}), between dataset statistics (Section \ref{sec:data_stats}) and model accuracy reinforces these findings. While accuracy for both models correlates positively with class pixel percentage, SegAssess uniquely shows non-negligible negative correlations between a class's accuracy and the proportion of its opposite class (e.g., FP accuracy vs TP proportion), supporting the hypothesis that the PQM approach leverages inter-class relationships in Section \ref{intro}. The analysis also confirms that error concentration near edges (EIB@3) negatively impacts the assessment accuracy of most classes (except TN), validating the focus on edge regions. The accuracy of inputted uncheck segmentation mask itself positively contributes to correctness categories while suppress the identification of FP and FN pixels. The GID dataset results exemplify the need for comprehensive analysis: its low EIB@3 (see Fig.~\ref{fig:data}(h)) might seem beneficial, but its extremely low FN proportion severely challenged AQSNet, leading to failure, while SegAssess maintained stable performance.

Qualitative comparisons presented in Fig.~\ref{fig:compare} visually corroborate the quantitative results. SegAssess consistently produces accurate PQM maps across different datasets and mask sources, whereas AQSNet shows instability (2$^{nd}$ row 3$^{rd}$ column in Fig.~\ref{fig:compare}) and frequent failures (the last row 3$^{rd}$ column in Fig.~\ref{fig:compare}). Furthermore, AQSNet's independent FP and FN predictions often result in significant spatial overlap (Fig.~\ref{fig:compare} pink areas), indicating difficulty in distinguishing error types, an issue largely resolved by SegAssess's unified four-class PQM approach. Overall, under standard training and validation conditions, SegAssess demonstrates superior accuracy, stability, and assessment granularity compared to the prior state-of-the-art.

\subsection{Fair Comparison across Different Pixel-wise SQA Granularities}\label{train_val2}
Following the direct comparison with the advanced PEL method AQSNet, we now broaden the analysis to facilitate a fair comparison with PEL methods of varying granularity. Specifically, we introduce a common 'correct/wrong' binary evaluation framework to directly compare against the paradigm of \cite{zaman2023segmentation}. By converting the outputs of SegAssess (TP/TN \textit{vs.} FP/FN) and AQSNet (TP/TN \textit{vs.} the rest) into this shared binary space, we can conduct a fair and direct side-by-side evaluation across three distinct pixel-wise SQA paradigms.

The quantitative results presented in Table~\ref{Tab: train_val2} clearly validate our hypothesis: finer assessment granularity translates to superior performance, even when evaluated under a simplified binary scheme. The method from \cite{zaman2023segmentation}, representing the coarsest 'correct/wrong' paradigm, serves as a foundational baseline and, as expected, yields the lowest accuracy. AQSNet, by introducing a more nuanced FP/FN distinction, generally achieves significantly better error localization. However, as noted previously, its dual-mask prediction strategy is prone to model collapse on challenging datasets (e.g., GID variants). Ultimately, SegAssess, operating under the most comprehensive four-class PQM paradigm, consistently and stably outperforms both PEL networks across all datasets. This demonstrates that by learning to distinguish all four quality categories, the model develops a more robust and accurate underlying representation of segmentation quality, a superiority that persists even when its output is simplified. These findings verify the discussion in Section~\ref{train_val1}.   

\begin{table}[t]
	\centering
	\caption{Quantitative comparison across different pixel-wise SQA paradigms with the standard Training-Validation protocol. SegAssess and AQSNet results are converted to 'correct/wrong' binary formant to measure accuracy.}
	\resizebox{0.5\textwidth}{!}{
		\renewcommand\arraystretch{1.5}
		\begin{tabular}{lcccccc}
			\toprule
			\multirow{2}{*}{Dataset} & \multicolumn{2}{c}{\cite{zaman2023segmentation} (Train-Val)} &
			\multicolumn{2}{c}{AQSNet (Train-Val)} & \multicolumn{2}{c}{SegAssess (Train-Val)} \\
			\cmidrule(lr){2-3} \cmidrule(lr){4-5} \cmidrule(lr){6-7}
			& mF1 & mIoU & mF1 & mIoU & mF1 & mIoU\\
			\midrule
			Inria-DeepLabv3+ &60.31&54.46&66.52 &58.03 &71.85&61.92\\
			Inria-HRNet &63.77&56.75&69.17&60.18&91.76&84.82\\
			Inria-TransUNet & 64.62 & 68.81 &59.92&59.90&71.39&61.73\\
			Inria-UNetFormer &63.95&56.78&69.39&60.09&73.22&63.35\\
			CrowdAI-DeepLabv3+ &70.71&62.21&71.25&62.25&75.41&65.81\\
			CrowdAI-HRNet &80.43&71.03& 86.56&78.31 &83.40&74.13\\
			CrowdAI-TransUNet &64.22&57.15& 71.47&62.16 &77.19&67.52\\
			CrowdAI-UNetFormer &82.72&73.59&86.59 &78.34 &83.48&74.23\\
			DeepGlobe-DeepLabv3+ & 64.64& 58.00 &70.17 &61.77 &73.15&64.21\\
			DeepGlobe-HRNet & 69.82 & 61.86 &68.80 &60.69 &73.03&64.09\\
			DeepGlobe-TransUNet & 69.23 & 61.43 & 70.79&63.20 &73.10&64.12\\
			DeepGlobe-UNetFormer & 71.06& 62.78 & 74.65&65.64 &76.36&67.07\\
			GID-DeepLabv3+ & 55.97 & 52.99 & 49.83&49.60 &67.75&60.83\\
			GID-HRNet & 61.05& 55.90& 0.91&0.46 &68.71&61.76\\
			GID-TransUNet & 55.91& 52.62 &1.23 &0.63 &70.72&63.51\\
			GID-UNetFormer & 52.38 & 50.88 &49.75 &49.50 &69.42&62.15\\
			BAQS-DeepLabv3+ & 67.69 & 60.52 &70.68 &62.65 &72.26&64.01\\
			BAQS-HRNet & 64.98 & 58.55 & 0.94&0.46 &70.20&62.23\\
			BAQS-TransUNet & 69.72 & 62.04 & 71.53& 63.37&71.02&62.83\\
			BAQS-UNetFormer & 67.80 & 60.56 & 49.70& 49.41&71.46&65.79\\
			WAQS-DeepLabv3+ & 71.58 & 63.48 &72.79 &64.39 &73.84&64.84\\
			WAQS-HRNet & 61.05& 55.90& 0.95&0.48 &69.78&61.70\\
			WAQS-TransUNet & 63.66&57.59 &1.01 &0.51 &70.73&62.47\\
			WAQS-UNetFormer & 65.79& 59.07& 49.74&49.49 &70.57&62.32\\
			
			\bottomrule
	\end{tabular}}
	\label{Tab: train_val2}
	\hspace{-1.8em}
\end{table}

\subsection{Generalization and Zero-Shot Transferability Analysis}\label{ZS}
\begin{table*}[!t]
	\caption{Quantitative comparison between SegAssess zero-shot inference (Zero-shot) and AQSNet standard training-validation (Train-Val). Comparable metrics are highlighted by grey.}
	\resizebox{\textwidth}{!}{
		\renewcommand\arraystretch{1.5}
		\begin{tabular}{l 
				>{\columncolor{gray!20}}c   
				>{\columncolor{gray!40}}c   
				>{\columncolor{gray!20}}c   
				>{\columncolor{gray!40}}c   
				c c 
				>{\columncolor{gray!20}}c   
				>{\columncolor{gray!40}}c   
				c c 
				>{\columncolor{gray!20}}c   
				>{\columncolor{gray!40}}c   
				c c }
			\toprule
			\multirow{2}{*}{Dataset} & \multicolumn{4}{c}{AQSNet (Train-Val)} & \multicolumn{10}{c}{SegAssess(Zero-shot)} \\
			\cmidrule(lr){2-5} \cmidrule(lr){6-15}
			& F1$^{FP}$ & IoU$^{FP}$ & F1$^{FN}$ & IoU$^{FN}$ & F1$^{TP}$ & IoU$^{TP}$ & F1$^{FP}$ & IoU$^{FP}$ & F1$^{TN}$ & IoU$^{TN}$ & F1$^{FN}$ & IoU$^{FN}$ & mF1 & mIoU \\
			\midrule
			\textit{Model-source:} &&&&&&&&&&&&&&\\
			Inria-OCRNet &24.15&13.81&26.35&15.26 &92.06&85.34&50.94&34.97&97.33&94.82&49.63&33.32&72.49&62.11\\
			CrowdAI-OCRNet &2.94&1.50&3.37&1.71&95.10&90.68&42.61&27.18&98.23&96.53&48.34&32.06& 71.07&61.61    \\
			DeepGlobe-OCRNet &16.93&9.79&62.11&45.72&78.22&65.24&56.48&41.01&98.90&97.83&72.57&57.29&76.54&65.34 \\
			GID-OCRNet &1.03&0.53&1.65&0.85&82.49&77.81&48.55&36.43&99.10&98.22&37.81&25.85&66.99&59.58\\
			BAQS-OCRNet &23.80&13.55&31.51&18.89 &98.00&96.08&45.70&29.75&99.18&98.38&51.01&35.08&73.47&64.82\\
			WAQS-OCRNet &36.27&22.27&13.03&7.00&94.04&88.90&47.42&31.47&99.22&98.45&30.73&18.54&67.85&59.34\\		
			\textit{Human-source:} &&&&&&&&&&&&&&\\
			BAQS-raw &34.51&21.80&59.67 &45.21&97.50&95.13&58.10&41.50&99.14&98.30&71.60&57.11&81.59&73.01\\
			WAQS-raw &84.79&74.19&19.70&11.44&93.39&87.89&79.54&66.27&98.55&97.13&45.20&30.25&79.17&70.39\\
			\bottomrule
	\end{tabular}}
	\label{Tab: zero_shot}
\end{table*}

\begin{table*}[!t]
	\caption{Quantitative evaluation of SegAssess's zero-shot transferability to unseen image/object domains.}
	\resizebox{\textwidth}{!}{
		\renewcommand\arraystretch{1.5}
		\begin{tabular}{llcccccccccc}
			\toprule
			Source Domain & Target datasets
			& F1$^{TP}$ & IoU$^{TP}$ & F1$^{FP}$ & IoU$^{FP}$ & F1$^{TN}$ & IoU$^{TN}$ & F1$^{FN}$ & IoU$^{FN}$ & mF1 & mIoU \\
			\midrule
			\multicolumn{12}{l}{\textit{Cross image domain (Inria building $\rightarrow$ CrowdAI building):}}\\
			\multirow{5}{*}{Inria} & CrowdAI-DeepLabv3+ &79.99&66.72&17.77&9.76&93.11&87.12&15.59&8.46&51.62&43.02\\
			& CrowdAI-HRNet &80.87&67.96&22.53&12.76&93.05&87.01&33.44&20.14&57.47&46.97\\
			& CrowdAI-TransUNet &80.21&67.04&19.90&11.06&93.18&87.24&16.74&9.15&52.51&43.62\\
			& CrowdAI-UNetFormer &80.86&67.94&27.91&16.24&93.07&87.05&32.98&19.78&58.70&47.75\\
			& CrowdAI-OCRNet &79.87&66.56&17.41&9.55&93.10&87.10&15.61&8.48&51.50&42.92\\
			\multicolumn{12}{l}{\textit{Cross image and object domain (BAQS building $\rightarrow$ WAQS water):}} \\
			\multirow{6}{*}{BAQS} & WAQS-DeepLabv3+ &77.29&64.65&26.07&15.06&98.61&97.25&4.29&2.20&51.56&44.79\\
			& WAQS-HRNet &76.42&63.62&18.85&10.49&98.63&97.29&4.67&2.41&49.64&43.45\\
			& WAQS-TransUNet &76.55&63.70&22.15&12.53&98.60&97.24&4.02&2.06&50.33&43.88\\
			& WAQS-UNetFormer &76.88&64.06&20.28&11.36&98.60&97.25&4.22&2.18&50.00&43.71\\
			& WAQS-OCRNet &77.48&64.79&23.06&13.12&98.61&97.28&3.81&1.96&50.74&44.28\\
			& WAQS-raw &77.17&64.20&59.71&42.78&98.00&96.09&7.05&3.72&60.48&51.70\\
			\bottomrule
	\end{tabular}}
	\label{Tab: zero_shot2}
\end{table*}

\begin{figure*}[!t]
	\centering
	\vspace{-0.8em}
	\subfloat{%
		\includegraphics[width=0.24\linewidth]{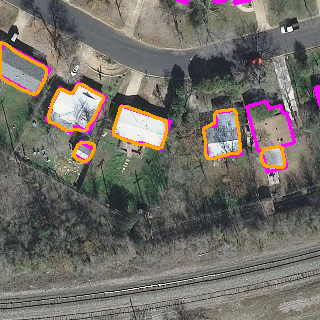}%
	}
	\subfloat{%
		\includegraphics[width=0.24\linewidth]{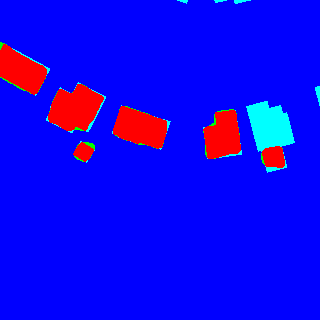}%
	}
	\subfloat{%
		\includegraphics[width=0.24\linewidth]{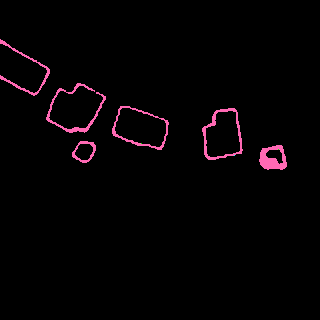}%
	}
	\subfloat{%
		\includegraphics[width=0.24\linewidth]{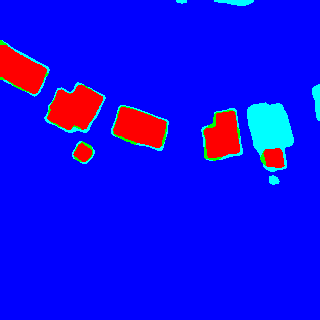}%
	}
	
	\vspace{-1.0em}
	\setcounter{subfigure}{0}
	
	\subfloat{%
		\includegraphics[width=0.24\linewidth]{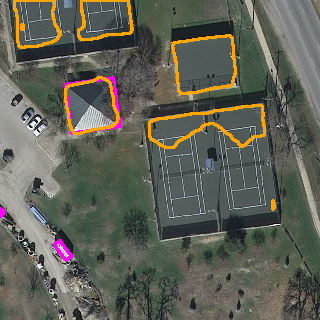}%
	} 
	\subfloat{%
		\includegraphics[width=0.24\linewidth]{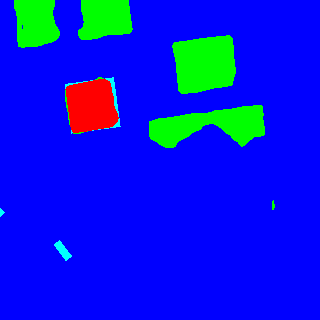}%
	}
	\subfloat{%
		\includegraphics[width=0.24\linewidth]{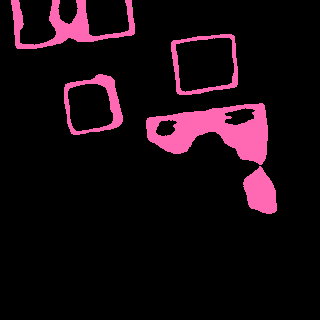}%
	}
	\subfloat{%
		\includegraphics[width=0.24\linewidth]{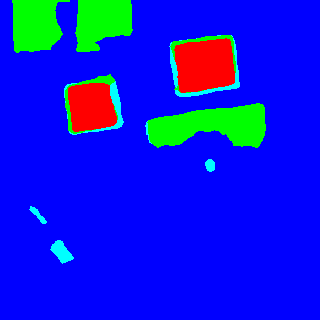}%
	}
	
	\vspace{-1.0em}
	\setcounter{subfigure}{0}
	
	\subfloat{%
		\includegraphics[width=0.24\linewidth]{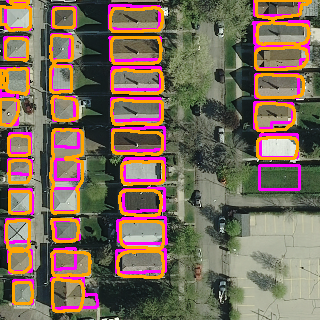}%
	} 
	\subfloat{%
		\includegraphics[width=0.24\linewidth]{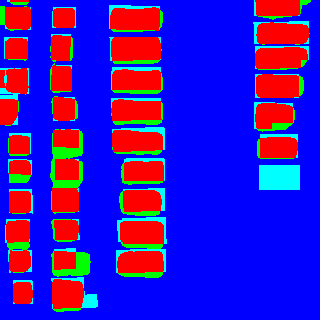}%
	}
	\subfloat{%
		\includegraphics[width=0.24\linewidth]{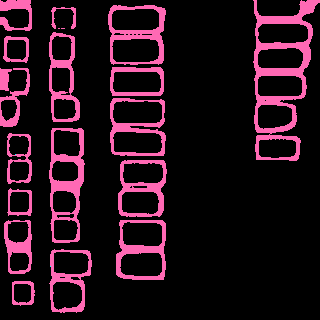}%
	}
	\subfloat{%
		\includegraphics[width=0.24\linewidth]{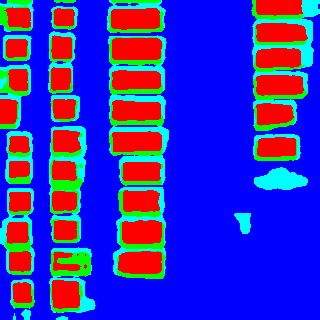}%
	}
	
	\vspace{-1.0em}
	\setcounter{subfigure}{0}
	
	\subfloat{%
		\includegraphics[width=0.24\linewidth]{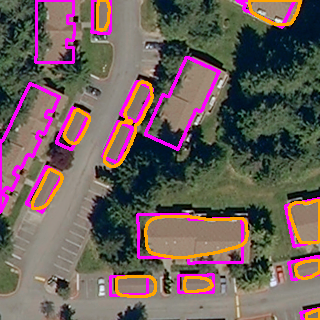}%
	} 
	\subfloat{%
		\includegraphics[width=0.24\linewidth]{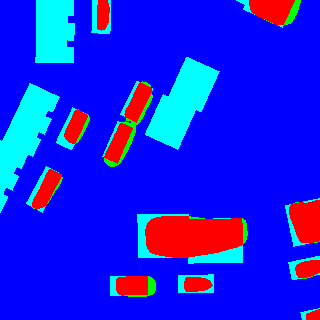}%
	}
	\subfloat{%
		\includegraphics[width=0.24\linewidth]{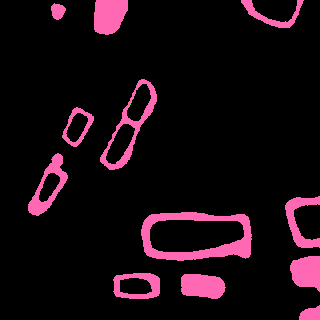}%
	}
	\subfloat{%
		\includegraphics[width=0.24\linewidth]{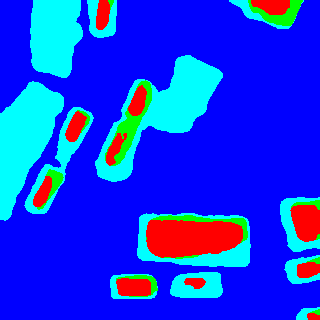}%
	}
	
	\vspace{-1.0em}
	\setcounter{subfigure}{0}
	
	\subfloat{%
		\includegraphics[width=0.24\linewidth]{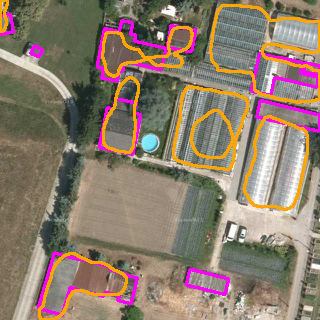}%
	} 
	\subfloat{%
		\includegraphics[width=0.24\linewidth]{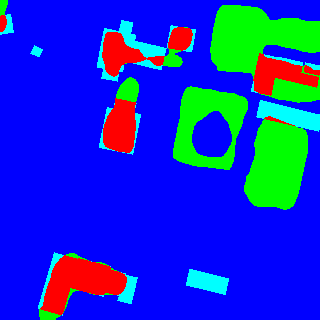}%
	}
	\subfloat{%
		\includegraphics[width=0.24\linewidth]{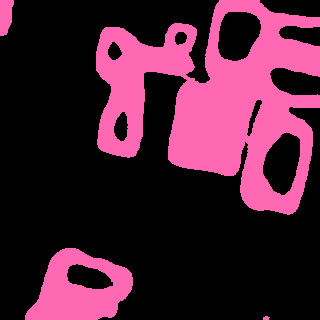}%
	}
	\subfloat{%
		\includegraphics[width=0.24\linewidth]{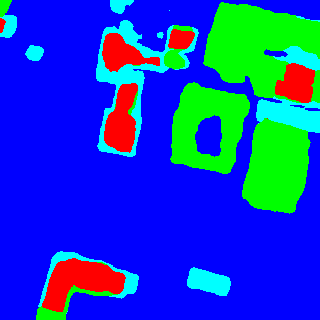}%
	}

	\caption{Zero-shot transfer comparison on Inria-OCRNet (ZS-MS scenario). Visual results comparing zero-shot SegAssess (4-class PQM) against trained AQSNet (FP/FN errors). Layout and color scheme as in Fig.~\ref{fig:compare}}  
	\label{fig:inria_ocrnet}
	\vspace{1.0em}
\end{figure*}

 \begin{figure*}[!t]
	\centering
	\vspace{-0.8em}
	\subfloat{%
		\includegraphics[width=0.24\linewidth]{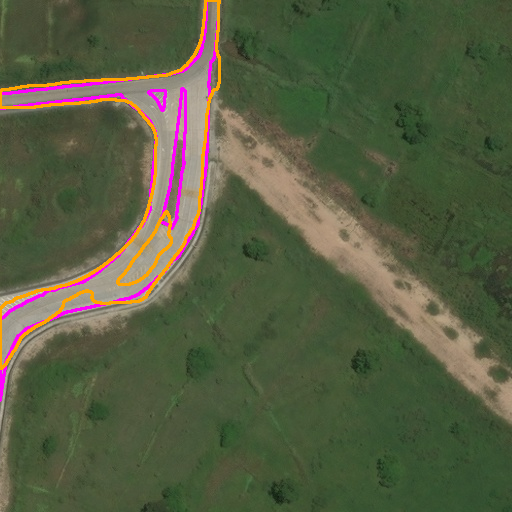}%
	}
	\subfloat{%
		\includegraphics[width=0.24\linewidth]{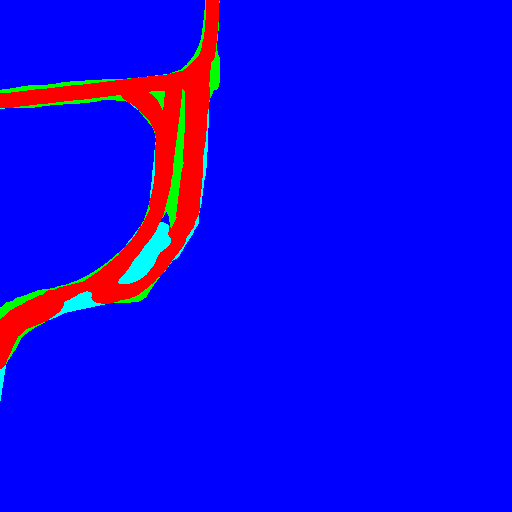}%
	}
	\subfloat{%
		\includegraphics[width=0.24\linewidth]{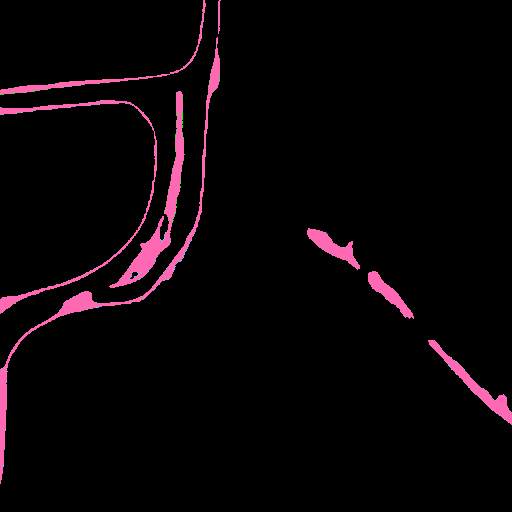}%
	}
	\subfloat{%
		\includegraphics[width=0.24\linewidth]{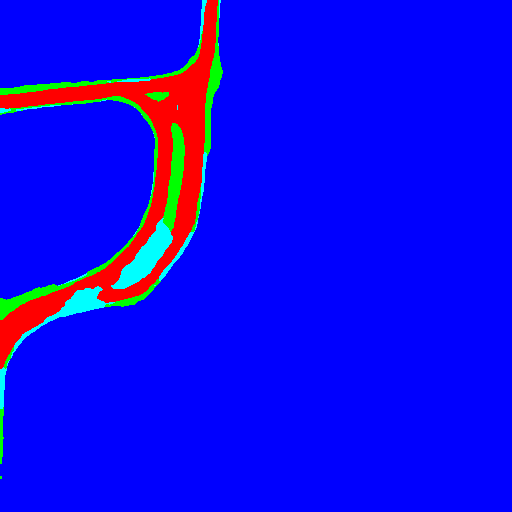}%
	}
	
	\vspace{-1.0em}
	\setcounter{subfigure}{0}
	
	\subfloat{%
		\includegraphics[width=0.24\linewidth]{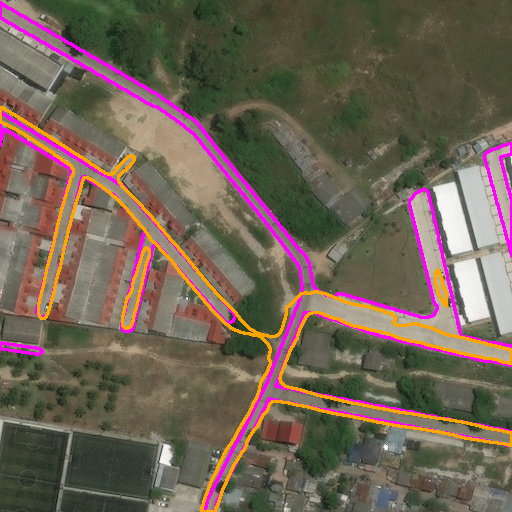}%
	} 
	\subfloat{%
		\includegraphics[width=0.24\linewidth]{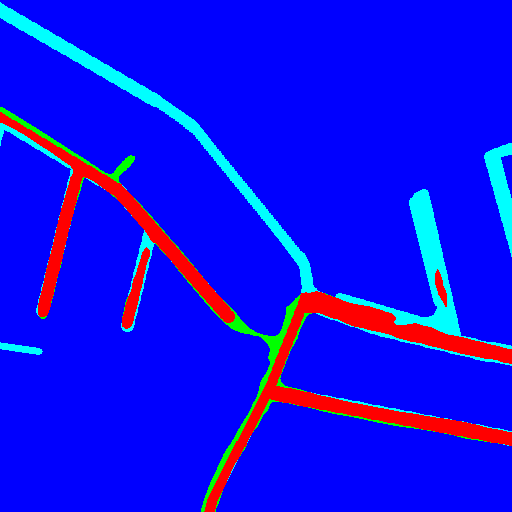}%
	}
	\subfloat{%
		\includegraphics[width=0.24\linewidth]{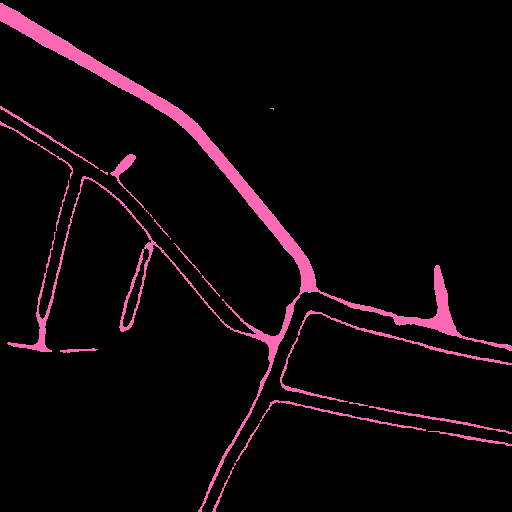}%
	}
	\subfloat{%
		\includegraphics[width=0.24\linewidth]{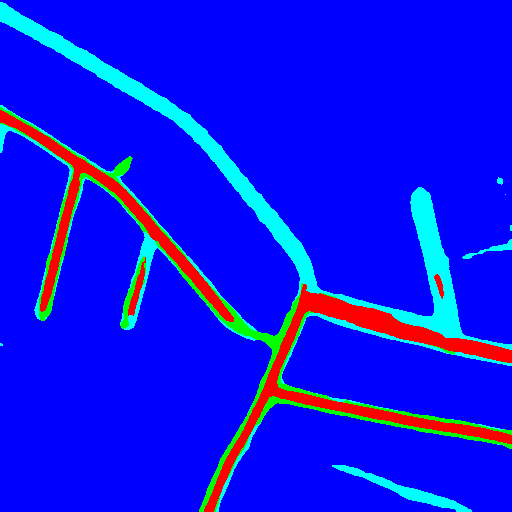}%
	}
	
	\vspace{-1.0em}
	\setcounter{subfigure}{0}
	
	\subfloat{%
		\includegraphics[width=0.24\linewidth]{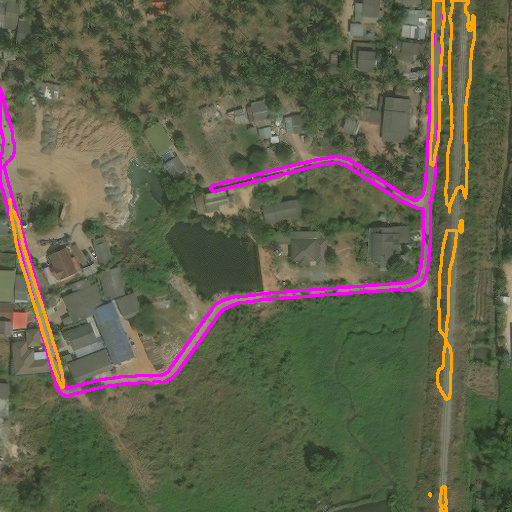}%
	} 
	\subfloat{%
		\includegraphics[width=0.24\linewidth]{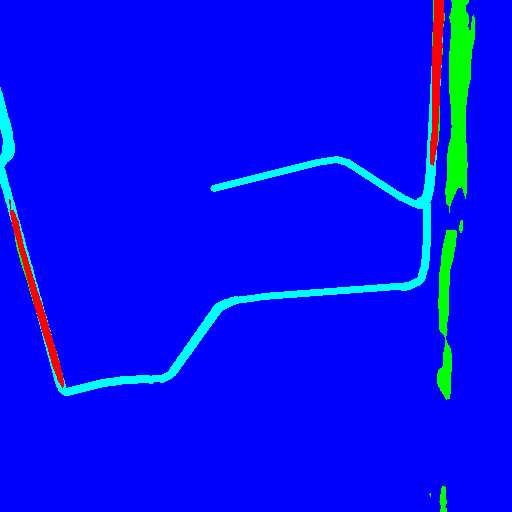}%
	}
	\subfloat{%
		\includegraphics[width=0.24\linewidth]{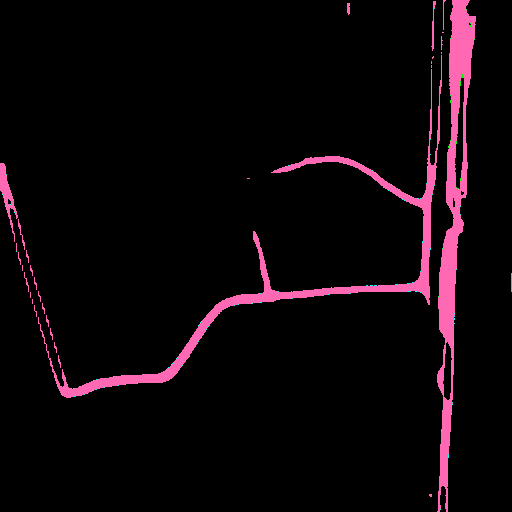}%
	}
	\subfloat{%
		\includegraphics[width=0.24\linewidth]{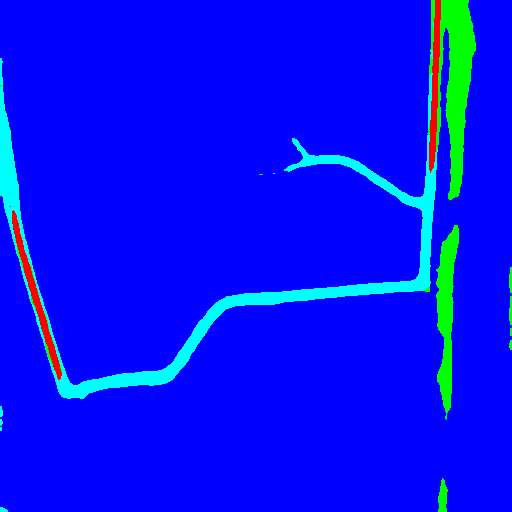}%
	}
	
	\vspace{-1.0em}
	\setcounter{subfigure}{0}
	
	\subfloat{%
		\includegraphics[width=0.24\linewidth]{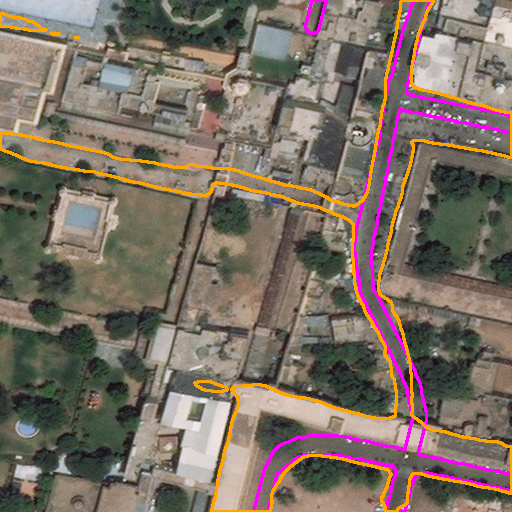}%
	} 
	\subfloat{%
		\includegraphics[width=0.24\linewidth]{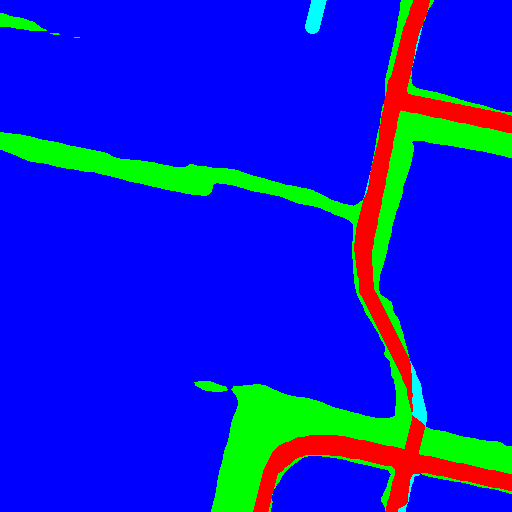}%
	}
	\subfloat{%
		\includegraphics[width=0.24\linewidth]{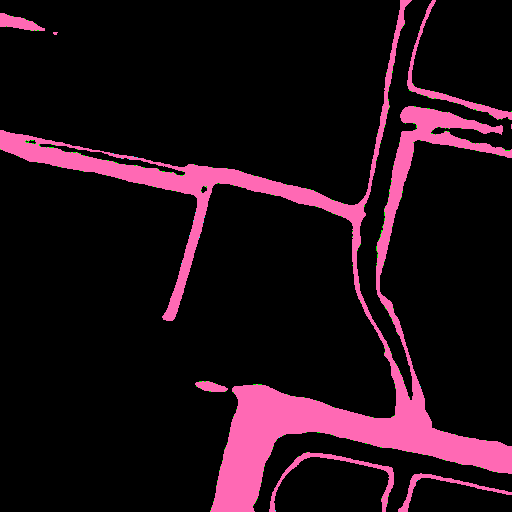}%
	}
	\subfloat{%
		\includegraphics[width=0.24\linewidth]{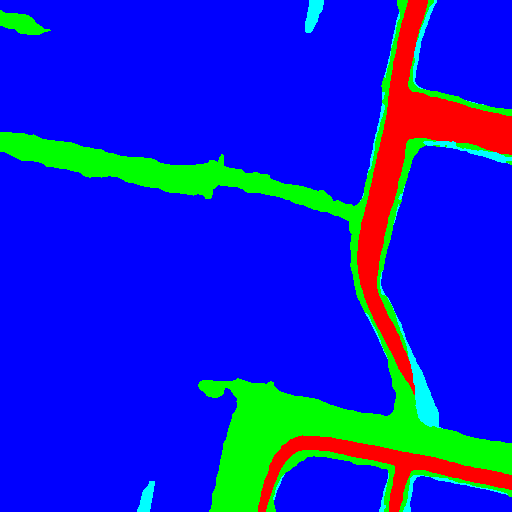}%
	}
	
	\vspace{-1.0em}
	\setcounter{subfigure}{0}
	
	\subfloat{%
		\includegraphics[width=0.24\linewidth]{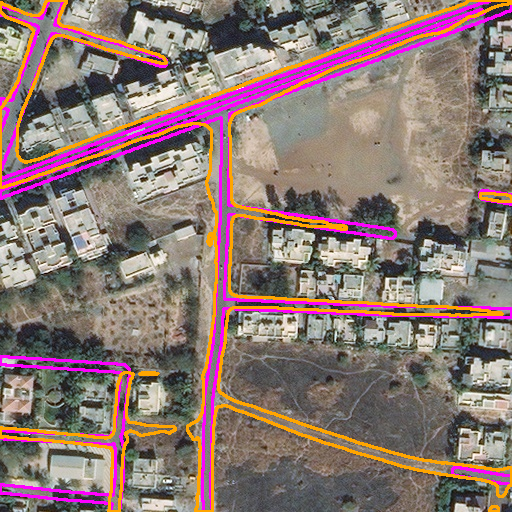}%
	} 
	\subfloat{%
		\includegraphics[width=0.24\linewidth]{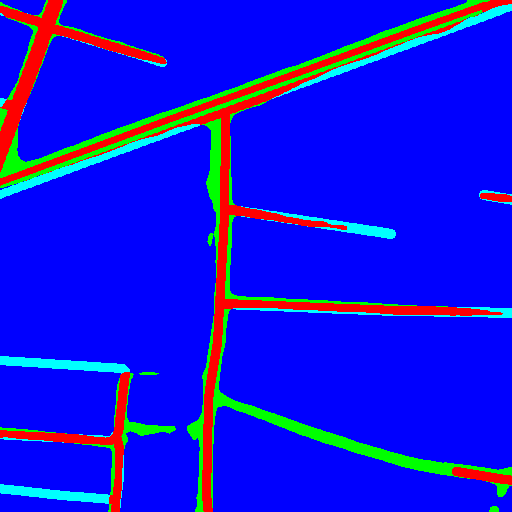}%
	}
	\subfloat{%
		\includegraphics[width=0.24\linewidth]{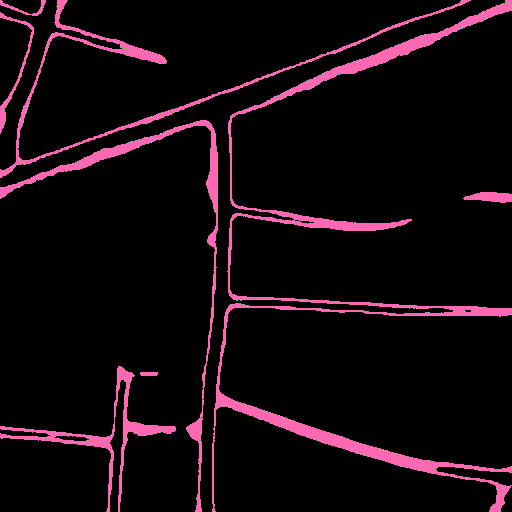}%
	}
	\subfloat{%
		\includegraphics[width=0.24\linewidth]{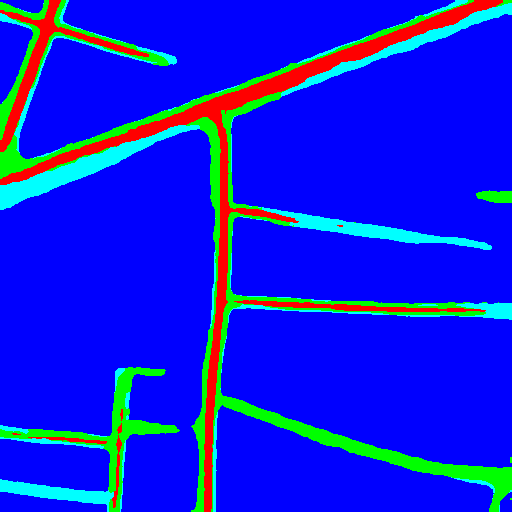}%
	}
	
	\caption{Zero-shot transfer comparison on DeepGlobe-OCRNet (ZS-MS scenario). Visual results comparing zero-shot SegAssess (4-class PQM) against trained AQSNet (FP/FN errors). Layout and color scheme as in Fig~\ref{fig:compare}.}  
	\label{fig:DeepGlobe_ocrnet}
	\vspace{1.0em}
\end{figure*}

\begin{figure*}[!t]
	\centering
	\vspace{-0.8em}
	\subfloat{%
		\includegraphics[width=0.24\linewidth]{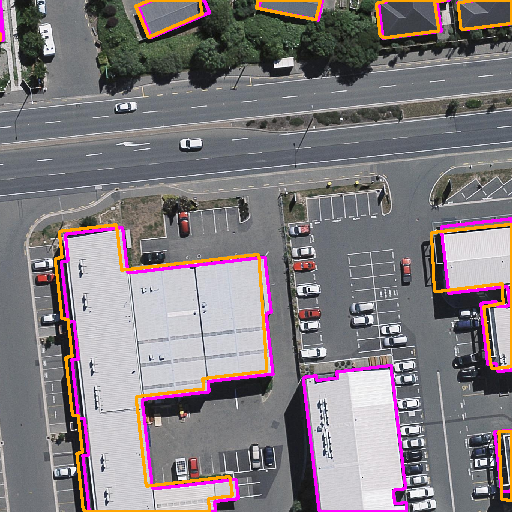}%
	}
	\subfloat{%
		\includegraphics[width=0.24\linewidth]{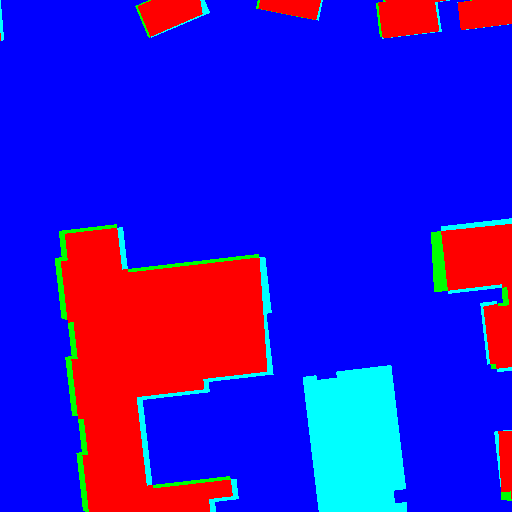}%
	}
	\subfloat{%
		\includegraphics[width=0.24\linewidth]{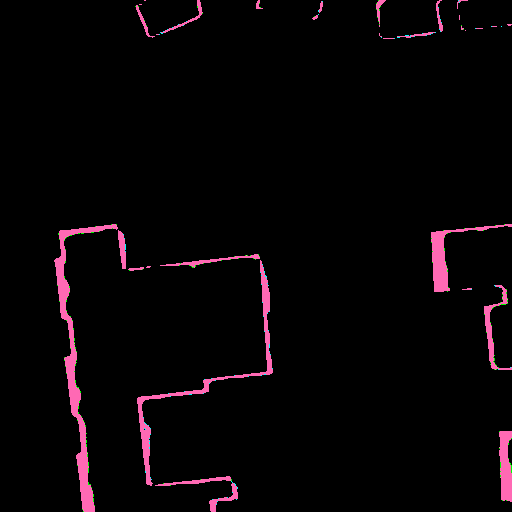}%
	}
	\subfloat{%
		\includegraphics[width=0.24\linewidth]{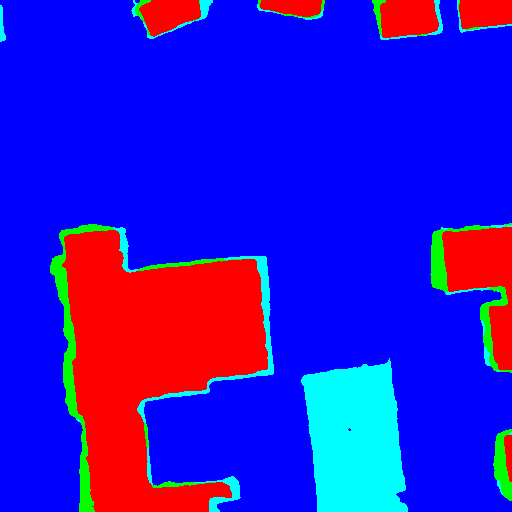}%
	}
	
	\vspace{-1.0em}
	\setcounter{subfigure}{0}
	
	\subfloat{%
		\includegraphics[width=0.24\linewidth]{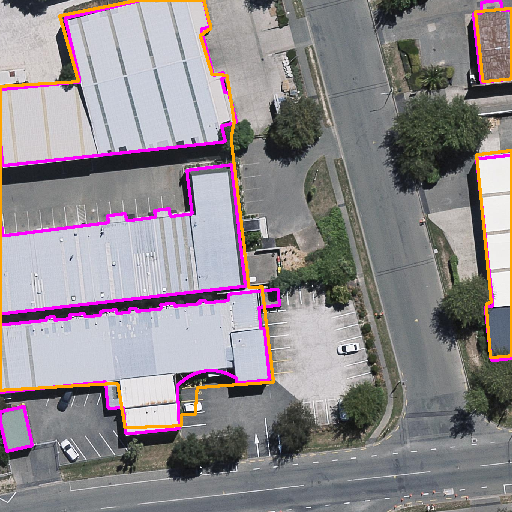}%
	} 
	\subfloat{%
		\includegraphics[width=0.24\linewidth]{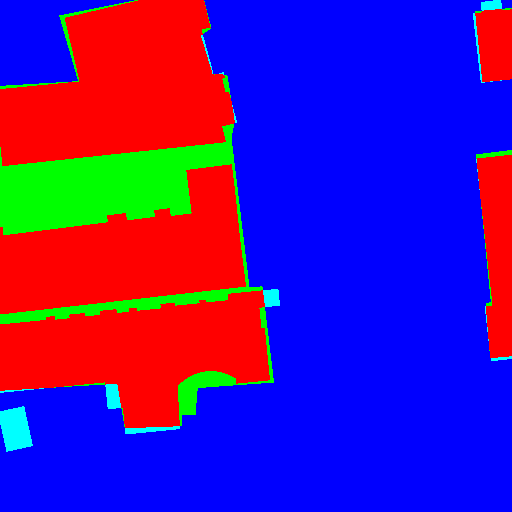}%
	}
	\subfloat{%
		\includegraphics[width=0.24\linewidth]{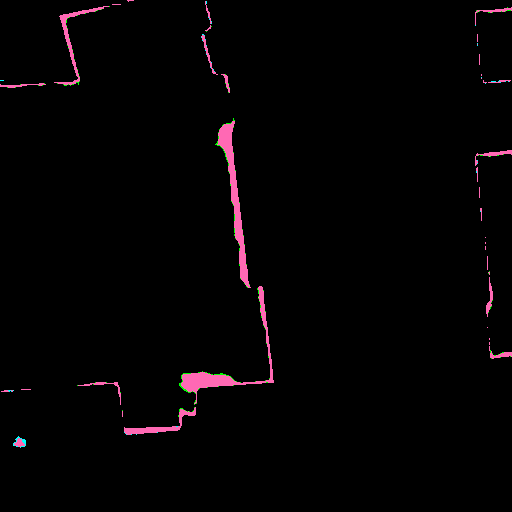}%
	}
	\subfloat{%
		\includegraphics[width=0.24\linewidth]{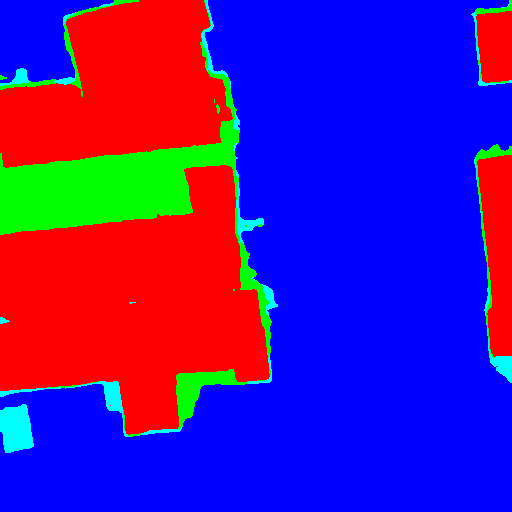}%
	}
	
	\vspace{-1.0em}
	\setcounter{subfigure}{0}
	
	\subfloat{%
		\includegraphics[width=0.24\linewidth]{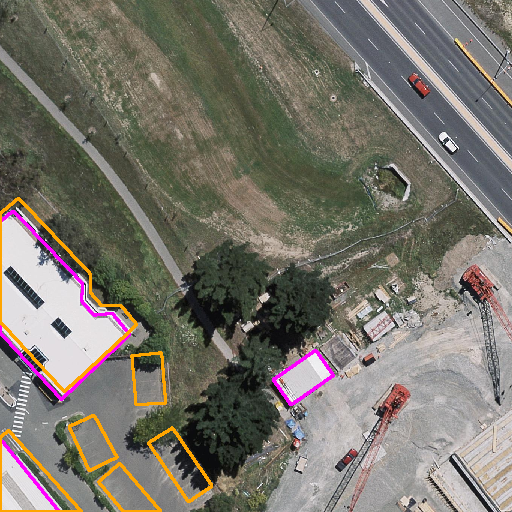}%
	} 
	\subfloat{%
		\includegraphics[width=0.24\linewidth]{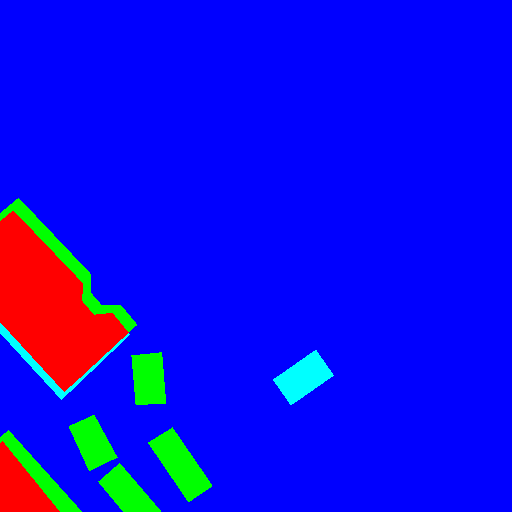}%
	}
	\subfloat{%
		\includegraphics[width=0.24\linewidth]{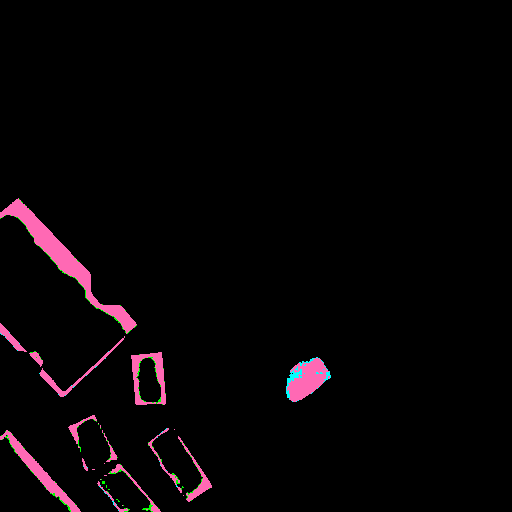}%
	}
	\subfloat{%
		\includegraphics[width=0.24\linewidth]{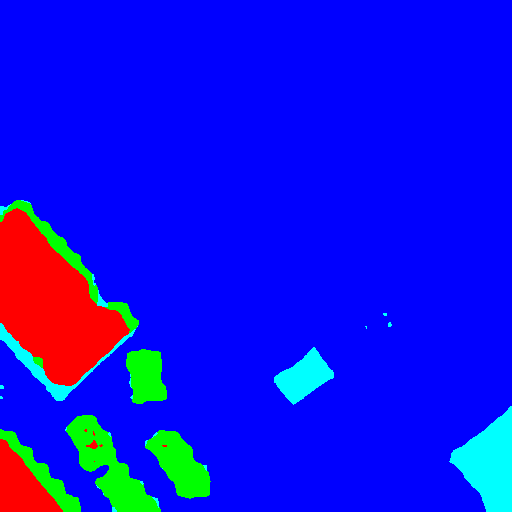}%
	}
	
	\vspace{-1.0em}
	\setcounter{subfigure}{0}
	
	\subfloat{%
		\includegraphics[width=0.24\linewidth]{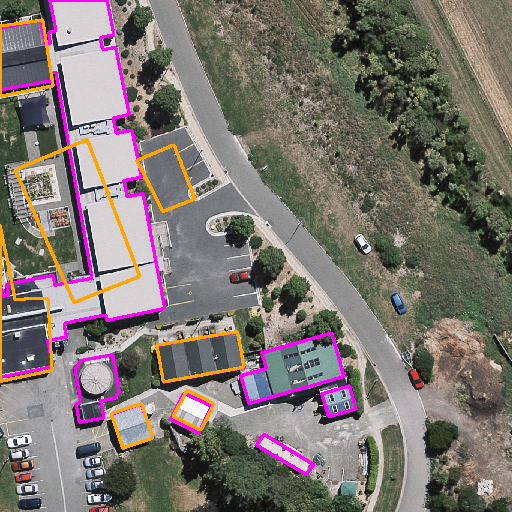}%
	} 
	\subfloat{%
		\includegraphics[width=0.24\linewidth]{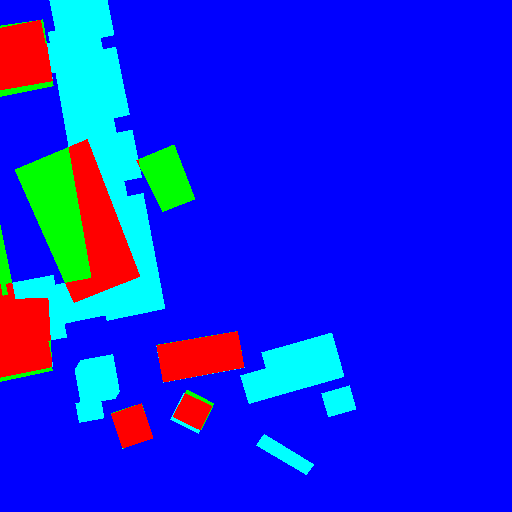}%
	}
	\subfloat{%
		\includegraphics[width=0.24\linewidth]{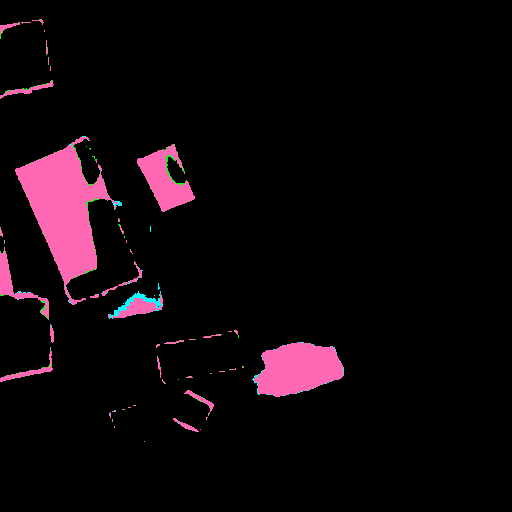}%
	}
	\subfloat{%
		\includegraphics[width=0.24\linewidth]{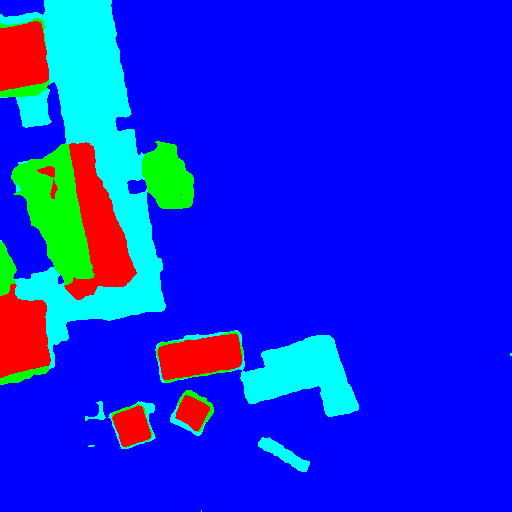}%
	}
	
	\vspace{-1.0em}
	\setcounter{subfigure}{0}
	
	\subfloat{%
		\includegraphics[width=0.24\linewidth]{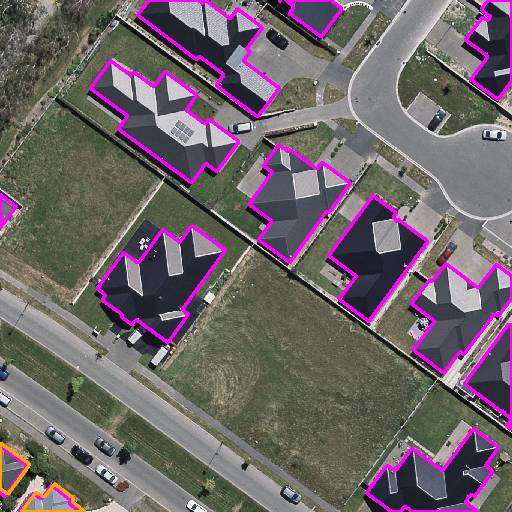}%
	} 
	\subfloat{%
		\includegraphics[width=0.24\linewidth]{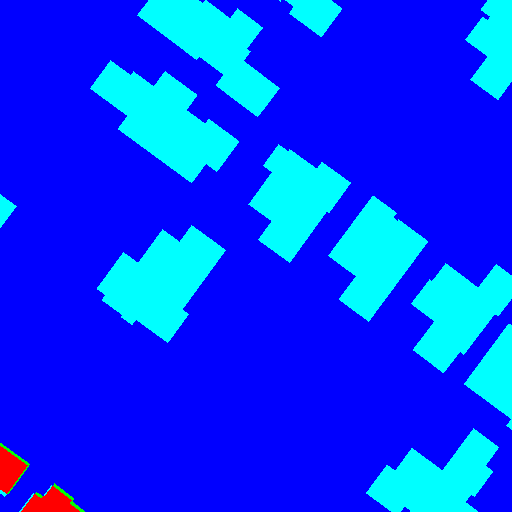}%
	}
	\subfloat{%
		\includegraphics[width=0.24\linewidth]{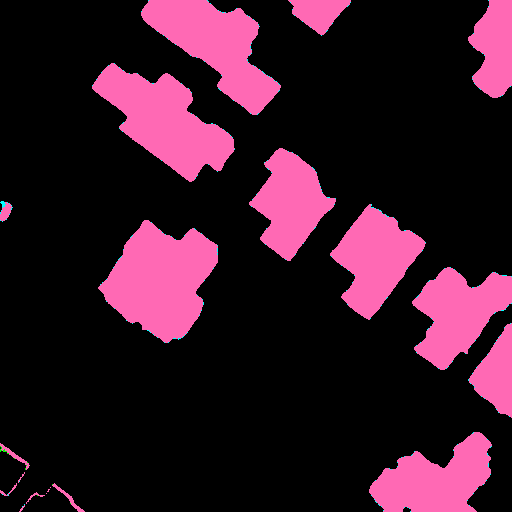}%
	}
	\subfloat{%
		\includegraphics[width=0.24\linewidth]{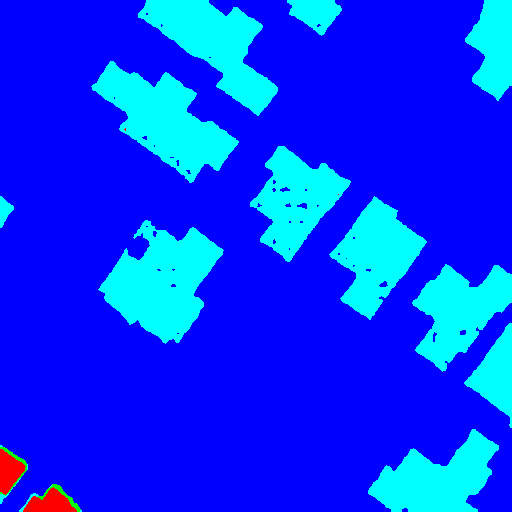}%
	}
	
	\caption{Zero-shot transfer comparison on BAQS-raw (ZS-HS scenario). Visual results comparing zero-shot SegAssess (4-class PQM) against trained AQSNet (FP/FN errors). Layout and color scheme as in Fig~\ref{fig:compare}}  
	\label{fig:BAQS_raw}
	\vspace{1.0em}
\end{figure*}

\begin{figure*}[!t]
	\centering
	\vspace{-0.8em}
	\subfloat{%
		\includegraphics[width=0.24\linewidth]{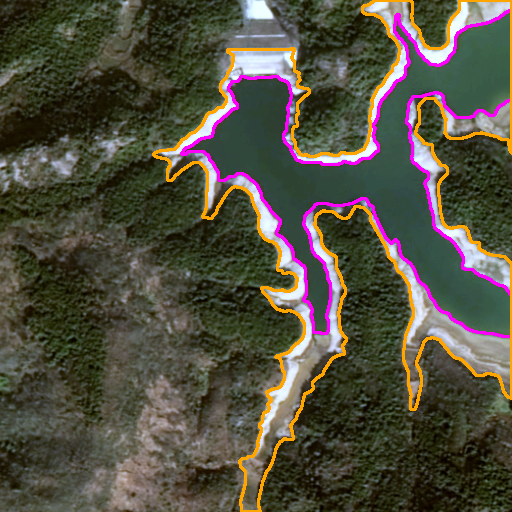}%
	}
	\subfloat{%
		\includegraphics[width=0.24\linewidth]{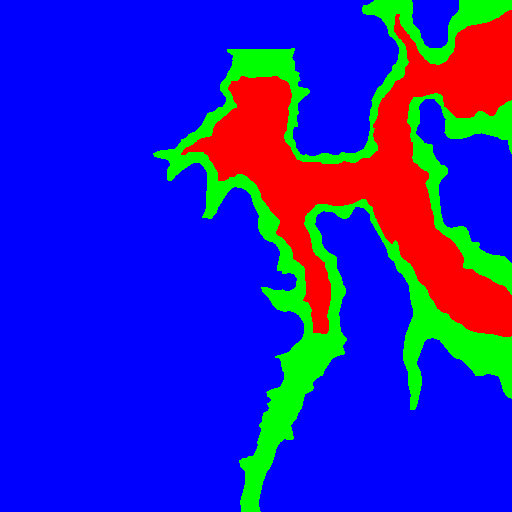}%
	}
	\subfloat{%
		\includegraphics[width=0.24\linewidth]{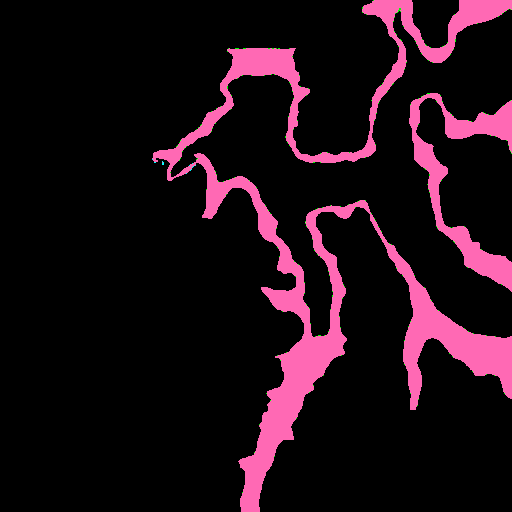}%
	}
	\subfloat{%
		\includegraphics[width=0.24\linewidth]{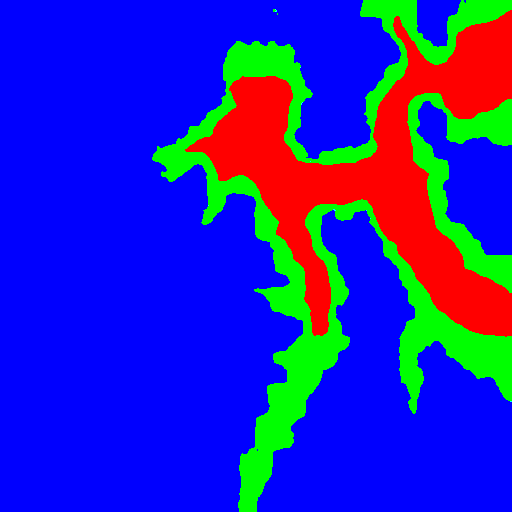}%
	}
	
	\vspace{-1.0em}
	\setcounter{subfigure}{0}
	
	\subfloat{%
		\includegraphics[width=0.24\linewidth]{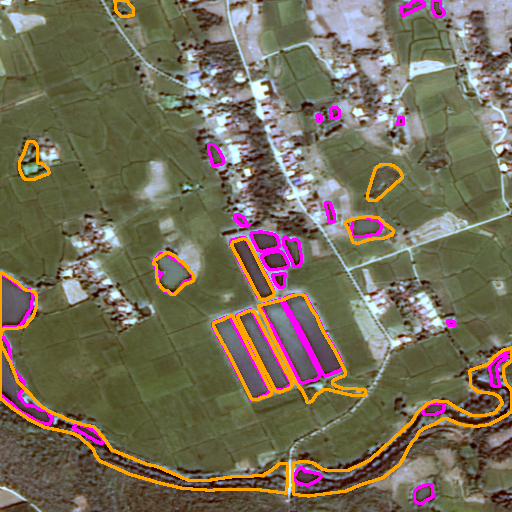}%
	} 
	\subfloat{%
		\includegraphics[width=0.24\linewidth]{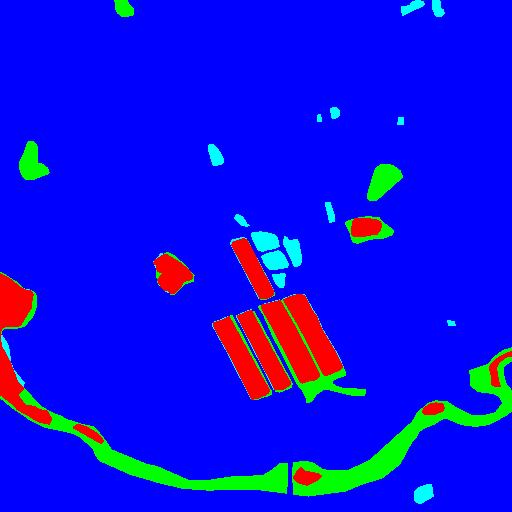}%
	}
	\subfloat{%
		\includegraphics[width=0.24\linewidth]{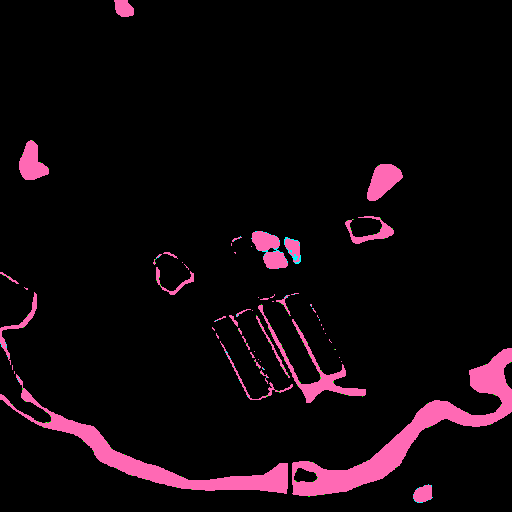}%
	}
	\subfloat{%
		\includegraphics[width=0.24\linewidth]{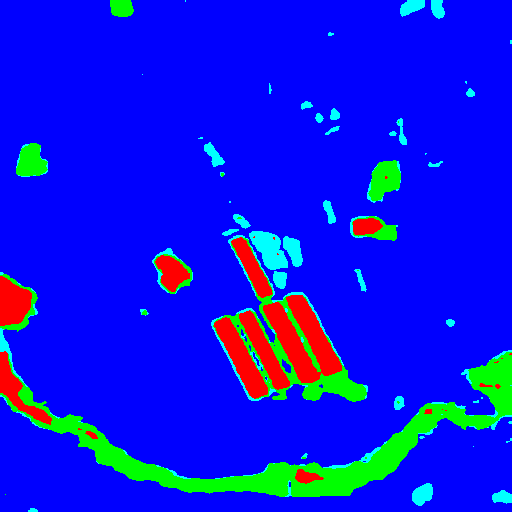}%
	}
	
	\vspace{-1.0em}
	\setcounter{subfigure}{0}
	
	\subfloat{%
		\includegraphics[width=0.24\linewidth]{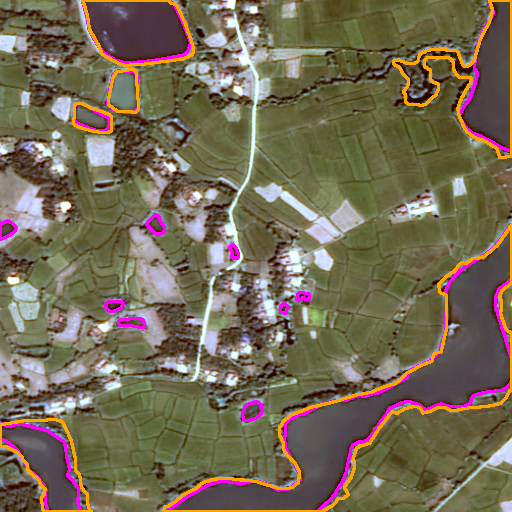}%
	} 
	\subfloat{%
		\includegraphics[width=0.24\linewidth]{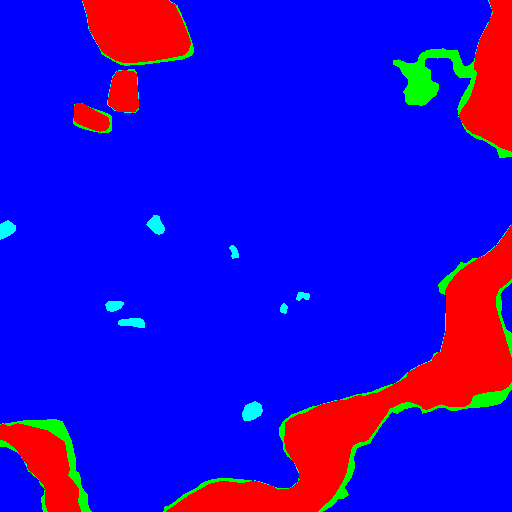}%
	}
	\subfloat{%
		\includegraphics[width=0.24\linewidth]{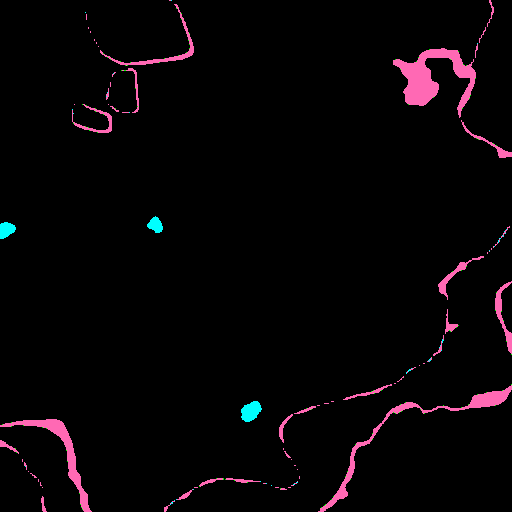}%
	}
	\subfloat{%
		\includegraphics[width=0.24\linewidth]{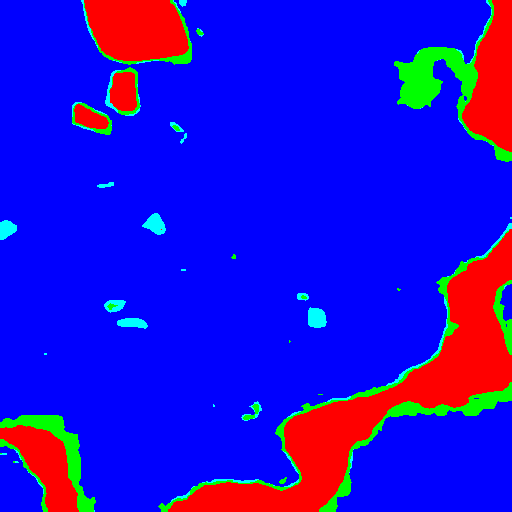}%
	}
	
	\vspace{-1.0em}
	\setcounter{subfigure}{0}
	
	\subfloat{%
		\includegraphics[width=0.24\linewidth]{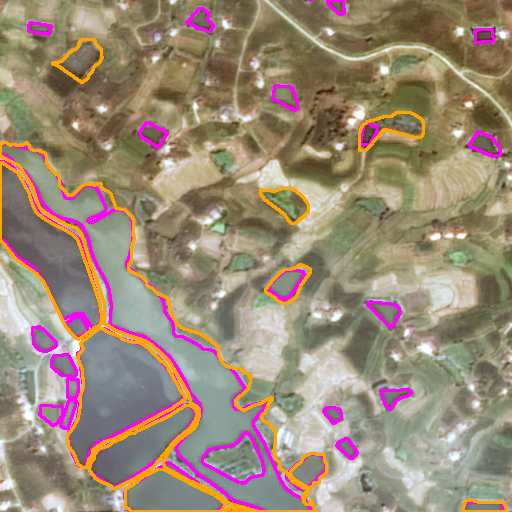}%
	} 
	\subfloat{%
		\includegraphics[width=0.24\linewidth]{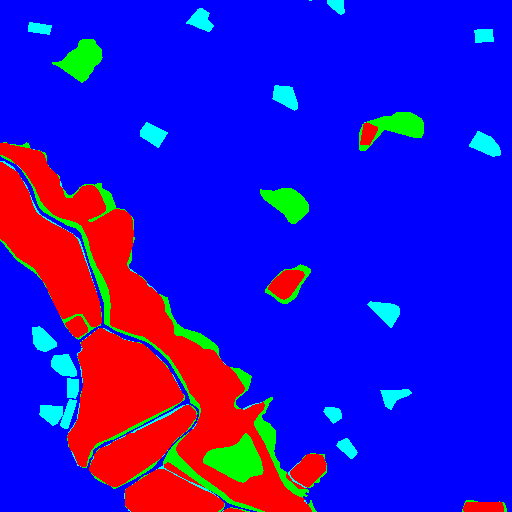}%
	}
	\subfloat{%
		\includegraphics[width=0.24\linewidth]{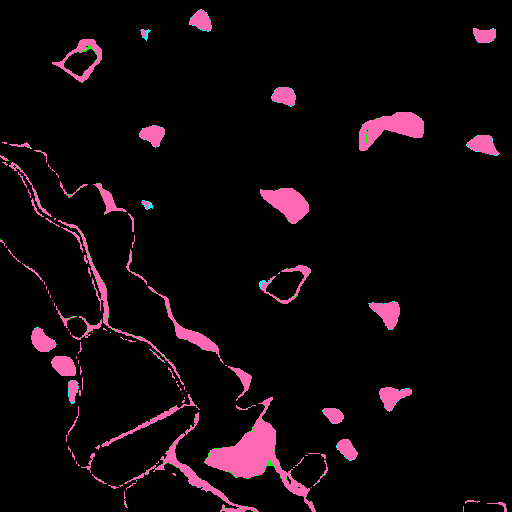}%
	}
	\subfloat{%
		\includegraphics[width=0.24\linewidth]{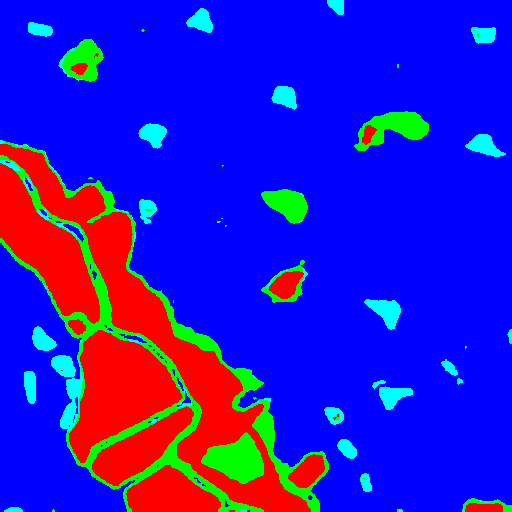}%
	}
	
	\vspace{-1.0em}
	\setcounter{subfigure}{0}
	
	\subfloat{%
		\includegraphics[width=0.24\linewidth]{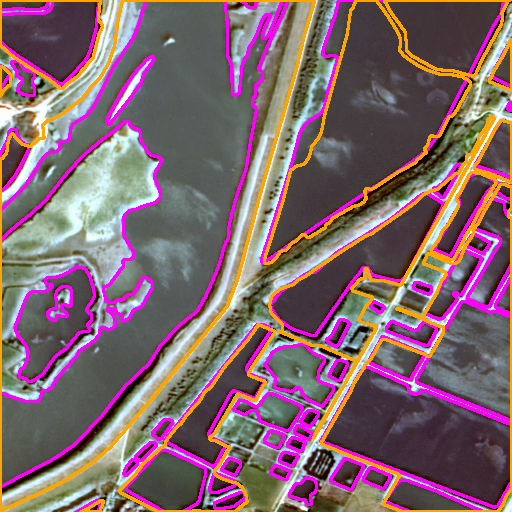}%
	} 
	\subfloat{%
		\includegraphics[width=0.24\linewidth]{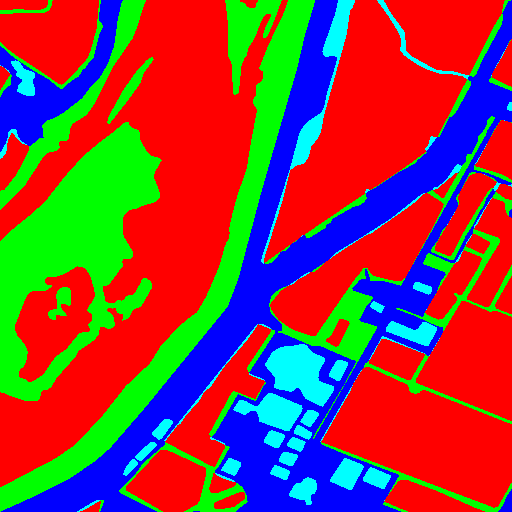}%
	}
	\subfloat{%
		\includegraphics[width=0.24\linewidth]{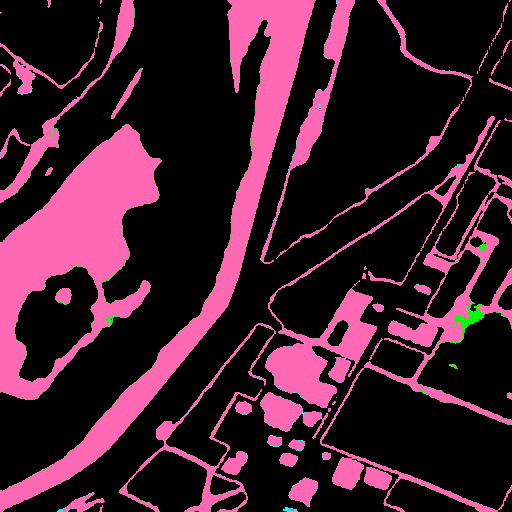}%
	}
	\subfloat{%
		\includegraphics[width=0.24\linewidth]{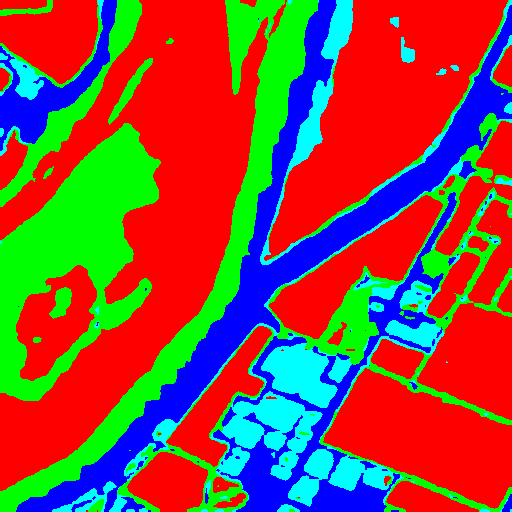}%
	}
	
	\caption{Zero-shot transfer comparison on WAQS-raw (ZS-HS scenario). Visual results comparing zero-shot SegAssess (4-class PQM) against trained AQSNet (FP/FN errors). Layout and color scheme as in Fig~\ref{fig:compare}}  
	\label{fig:WAQS_raw}
	\vspace{1.0em}
\end{figure*}

 \begin{figure*}[!t]
	\centering
	
	\subfloat{%
		\includegraphics[width=0.15\linewidth]{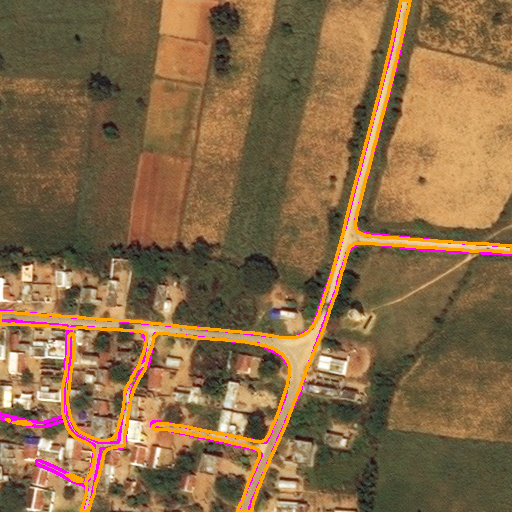}%
	}\hspace{0.3em}%
	\subfloat{%
		\includegraphics[width=0.15\linewidth]{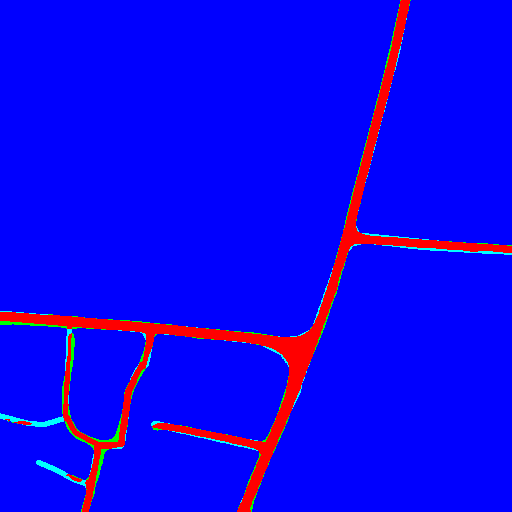}%
	}\hspace{0.3em}%
	\subfloat{%
		\includegraphics[width=0.15\linewidth]{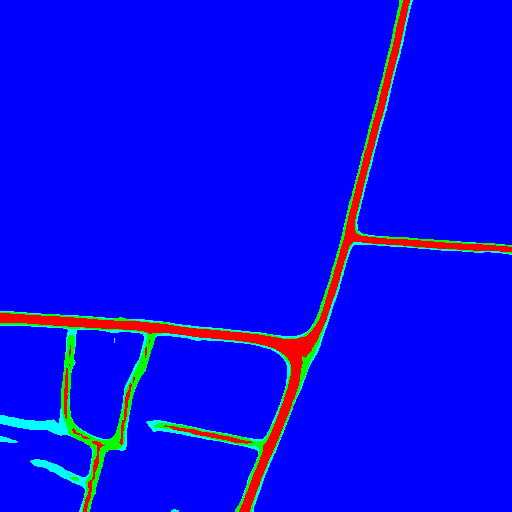}%
	}\hspace{0.3em}%
	\subfloat{%
		\includegraphics[width=0.15\linewidth]{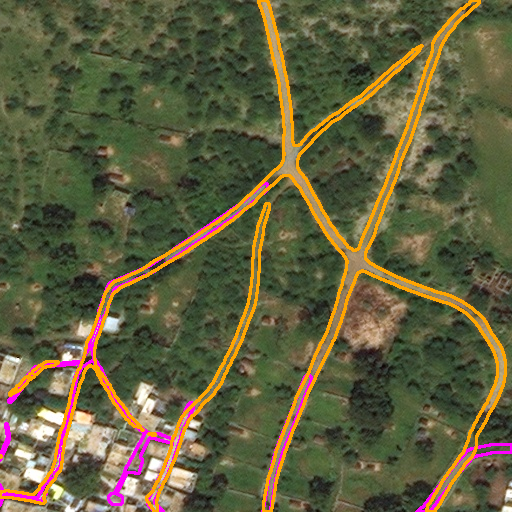}%
	}\hspace{0.3em}%
	\subfloat{%
		\includegraphics[width=0.15\linewidth]{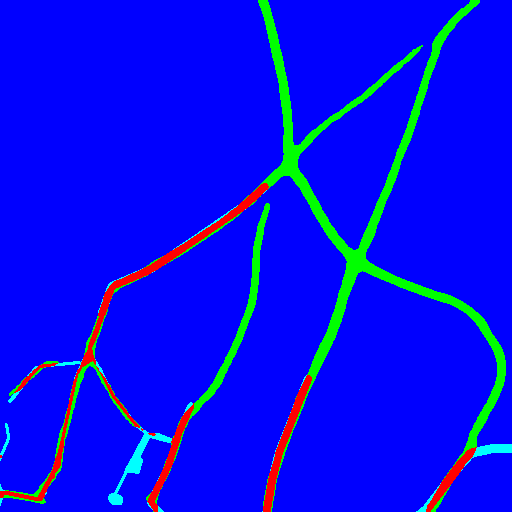}%
	}\hspace{0.3em}%
	\subfloat{%
		\includegraphics[width=0.15\linewidth]{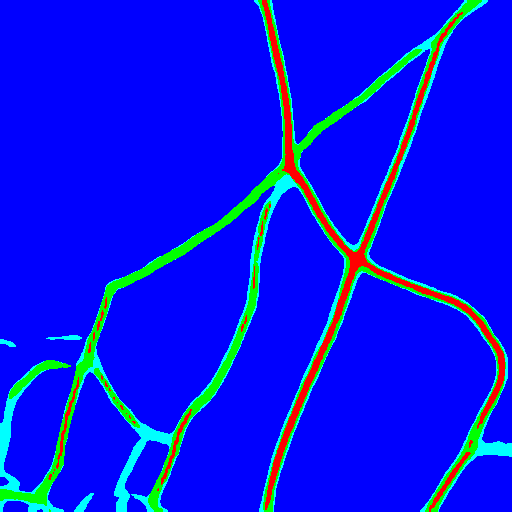}%
	}\hspace{0.3em}%
	
	\vspace{-0.7em}
	\setcounter{subfigure}{0}
	
	\subfloat{%
		\includegraphics[width=0.15\linewidth]{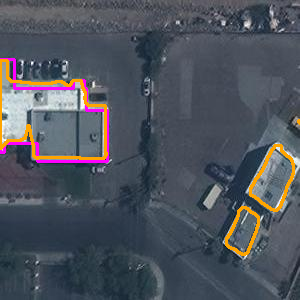}%
	}\hspace{0.3em}%
	\subfloat{%
		\includegraphics[width=0.15\linewidth]{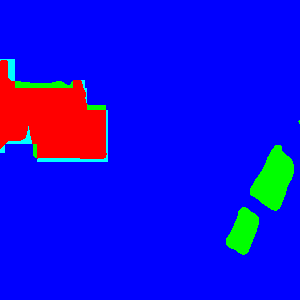}%
	}\hspace{0.3em}%
	\subfloat{%
		\includegraphics[width=0.15\linewidth]{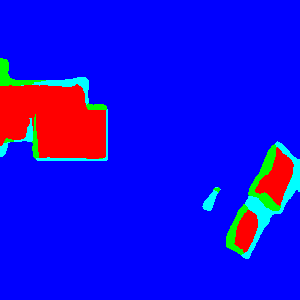}%
	}\hspace{0.3em}%
	\subfloat{%
		\includegraphics[width=0.15\linewidth]{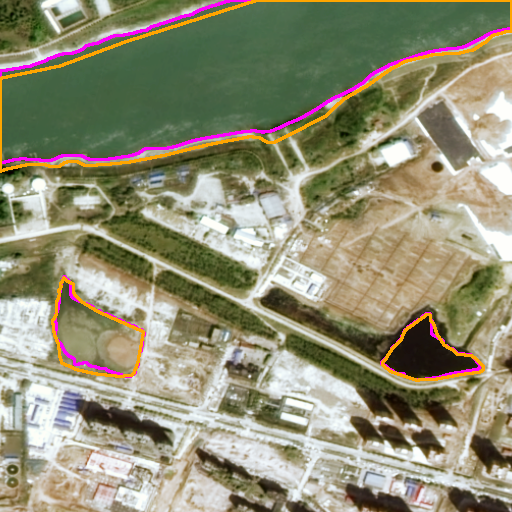}%
	}\hspace{0.3em}%
	\subfloat{%
		\includegraphics[width=0.15\linewidth]{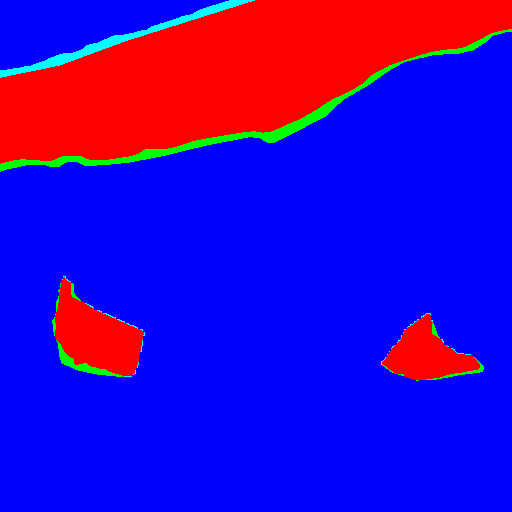}%
	}\hspace{0.3em}%
	\subfloat{%
		\includegraphics[width=0.15\linewidth]{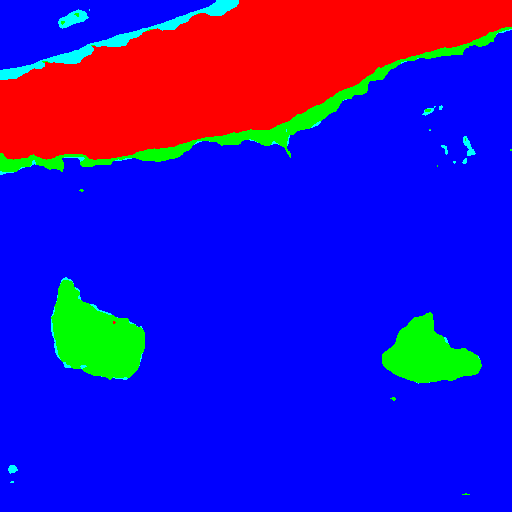}%
	}\hspace{0.3em}%
	
	\caption{Examples illustrating limitations of SegAssess, showing residual intra-class confusion (FP vs FN or TP vs TN). Four cases (arranged in two rows) each display the image with \textcolor{orange}{mask under evaluation}/\textcolor{magenta}{GT} contours, GT PQM and SegAssess 4-class PQM prediction. Colors represent \textcolor{red}{TP}, \textcolor{green}{FP}, \textcolor{blue}{TN} and \textcolor{cyan}{FN}.}  
	\label{fig:limitations}
\end{figure*}
A critical measure of a practical SQA framework is its ability to generalize to conditions unseen during training. In this section, we rigorously evaluate the transferability of SegAssess from two critical and progressively challenging perspectives. First, we assess its ability to handle segmentation masks from new sources (both model- and human-generated) while remaining within the learned image and object domains. Second, we probe the boundaries of its capabilities by testing its more demanding zero-shot performance when transferring to entirely new image domains and object categories. This two-tiered analysis provides a comprehensive picture of the model's robustness and practical deployment potential.

\subsubsection{Zero-shot transferring to unseen masks} \label{sec:ZS1}
A critical requirement for practical SQA frameworks is the ability to generalize to segmentation masks generated by unseen models or possessing characteristics different from the training data. We evaluated this crucial zero-shot (ZS) transferability of SegAssess using the reserved test sets: the six X-OCRNet variants representing unseen model-source (ZS-MS) masks, and the BAQS-raw and WAQS-raw datasets representing unseen human-source (ZS-HS) masks. SegAssess, trained using the AMS strategy (which did not include OCRNet or raw human masks), was applied directly to these datasets without any further fine-tuning. We compared its ZS performance against the performance achieved by AQSNet under its standard Train-Val protocol (see Section \ref{train_val1}), providing a challenging baseline.

As suggested in Tab.~\ref{Tab: zero_shot}, SegAssess is able to execute zero-shot transferring to both model-source and human-source unseen segmentation masks with impressive overall accuracy of at least 66.99$\%$ mF1 and 59.34$\%$ mIoU. When evaluating on the ZS-MS (X-OCRNet) datasets, SegAssess successfully performed inference across all six variants, consistently outperforming the AQSNet baseline even though AQSNet was trained specifically for each corresponding dataset source. This highlights SegAssess's robustness to unseen model architectures, contrasting with AQSNet's continued instability, including complete failures on CrowdAI-OCRNet and GID-OCRNet even under its Train-Val protocol. Visual comparisons in Fig.~\ref{fig:inria_ocrnet} and Fig.~\ref{fig:DeepGlobe_ocrnet} qualitatively demonstrate the transferability of SegAssess in real-world scenes. Applying to unseen model-generated segmentation masks, SegAssess generally segments TP, FP, TN, FN with patterns properly in line with GT.

Similarly, when assessing the unseen human-sourced masks in the ZS-HS scenario (BAQS-raw and WAQS-raw), zero-shot SegAssess again exhibited superior accuracy compared to the trained AQSNet baseline. While AQSNet achieved its best results on these specific datasets (potentially due to its original design focus), SegAssess's inherent four-class PQM approach provided a more comprehensive and accurate assessment even without prior exposure to these human annotation styles. Interestingly, SegAssess achieved higher overall scores on the ZS-HS datasets compared to the ZS-MS datasets. This could be partly attributed to the specific error distributions in the raw datasets (Fig.~\ref{fig:data}): BAQS-raw and WAQS-rawhas have a relatively high proportion of FN and FP pixels, respectively, which might slightly inflate the corresponding class metrics and the overall average scores.

Qualitative results further illustrate SegAssess's strong ZS capabilities. Visual comparisons for the ZS-MS scenario (Fig.~\ref{fig:inria_ocrnet} and Fig.~\ref{fig:DeepGlobe_ocrnet}) show SegAssess generating PQM maps that generally align well with the ground truth, capturing the TP, FP, TN, and FN patterns effectively despite minor imperfections, and often appearing superior to the trained AQSNet results. For the ZS-HS scenario ((Fig.~\ref{fig:BAQS_raw} and Fig.~\ref{fig:WAQS_raw}), while AQSNet shows improved FP/FN discrimination compared to its performance on model-source data, SegAssess consistently delivers finer classification detail due to its PQM formulation. Additionally, the manually derived masks in BAQS-raw and WAQS-raw tend to have more regular object outlines compared to some model outputs, potentially leading to more visually regular and quantitatively favorable PQM results from SegAssess.

In conclusion, these experiments demonstrate SegAssess's significant capability for zero-shot transfer to both unseen model-or human-source masks, validating its robustness and practical utility for deployment in real-world SQA applications where the source or quality characteristics of the segmentation mask may not be known a priori.

\subsubsection{Zero-shot transferring to unseen image/object domains} \label{sec:ZS2}
Building upon the analysis of mask-source transferability, we now investigate a more stringent test of generalization: the model's performance when transferred to entirely new image or object domains. This directly probes the boundaries of the model's learned representations. As detailed in Table~\ref{Tab: zero_shot2} we conducted two cross-domain experiments: one transferring from the CrowdAI (building) domain to the Inria (building) domain to assess cross-image-domain performance, and another transferring from the BAQS (building) domain to the WAQS (water) domain to evaluate cross-image-and-object-domain performance.

The results indicate that transferring to an unseen image domain, even with the same object class, leads to a noticeable performance degradation. While the accuracy for identifying correct regions (TP and TN) remains relatively robust, declining by approximately 10$\%$, the model's ability to precisely localize error categories (FP and FN) is more significantly impacted. The challenge is further amplified when crossing both image and object domains (BAQS building $\rightarrow$ WAQS water). In this scenario, SegAssess retains a strong capability to identify the prevalent TN category and a moderate, albeit reduced, ability to identify TP pixels. However, its capacity to recognize the much rarer error categories (FP and FN) is substantially diminished, approaching near-zero performance.

In summary, this two-tiered analysis clarifies the generalization capabilities and current limitations of SegAssess. It demonstrates remarkable robustness in zero-shot transfer to unseen mask sources within a consistent image and object domain (as shown in Section~\ref{sec:ZS1}). However, transferring across domains remains a significant challenge, particularly for the identification of minority error classes (FP/FN). This distinction defines the current operational boundaries of SegAssess and highlights critical directions for future work, as discussed in Section~\ref{limitation}.

\section{Limitations and future work}\label{limitation}

While SegAssess demonstrates impressive generalization across numerous datasets and achieves reliable zero-shot transferability to unseen segmentation masks from both model and human sources, certain limitations warrant further investigation. Although significantly improved compared to prior methods, SegAssess can still exhibit confusion between closely related PQM categories (i.e., intra-wrongness confusion between FP and FN, and intra-correctness confusion between TP and TN) in particularly challenging scenarios (Fig.~\ref{fig:limitations}). Furthermore, our ablation study on backbone sizes (see Section \ref{sec:designs}) indicated diminishing returns with larger models (SAM-L, SAM-H), suggesting that the current scale of available SQA training data may limit the effective utilization of these larger architectures.Finally, Section~\ref{ZS} clarify the scope of the demonstrated zero-shot capability: SegAssess excels at assessing unseen masks within the learned image and object domain (e.g., buildings, roads, water in high-resolution optical imagery), but its ability to generalize in a zero-shot manner to entirely new image domains or fundamentally different object types remains limited.
 
To address these limitations, we propose several avenues for future research. Firstly, to potentially improve fine-grained class discrimination and better leverage the scaling properties observed in foundation models, the development of a significantly larger-scale SQA benchmark dataset is needed. Such a dataset, ideally containing millions of image-mask-GT triplets across diverse domains, could unlock the potential of larger backbone architectures. Secondly, inspired by recent advances in foundation models \cite{kirillov2023segment, guo2024skysense,bountos2025fomo, zhao2025multi} exploring multi-modal and multi-task joint training appears promising. By training the SQA task simultaneously on data covering multiple object types and varied image domains (potentially including different sensor types or geographical settings), possibly alongside related objectives like segmentation itself, we could significantly improve the model's ability to generalize to unseen scenarios and achieve broader cross-domain, cross-object transferability.
\section{Conclusion}\label{conclusion}
In this paper, We introduced SegAssess, a novel deep learning framework founded on the Panoramic Quality Mapping (PQM) paradigm. By reformulating SQA as a comprehensive four-class segmentation task i.e. predicting True Positive (TP), False Positive (FP), True Negative (TN), and False Negative (FN) pixels, SegAssess provides a detailed, pixel-level quality map. The success of SegAssess stems from several key innovations built upon an enhanced Segment Anything Model (SAM) architecture. Uniquely leveraging the input segmentation mask as a prompt facilitates effective feature integration for the assessment task. Specific architectural advancements, including the Edge Guided Compaction (EGC) branch with its Aggregated Semantic Filter (ASF) module, effectively target challenging object boundaries and improve differentiation among quality categories. Furthermore, the Augmented Mixup Sampling (AMS) training strategy significantly enhances robustness and transferability by exposing the model to diverse mask characteristics representative of real-world variations. Comprehensive experiments demonstrated that SegAssess not only achieves state-of-the-art performance under standard training-validation protocols but, crucially, exhibits remarkable zero-shot transferability to entirely unseen segmentation masks, whether generated by different model architectures or sourced from human annotations. This validation underscores the effectiveness of the PQM paradigm implemented via SegAssess as a practical, robust, and transferable solution for unsupervised SQA. We hope this work paving the way for more reliable quality control in diverse remote sensing applications reliant on image segmentation.

\IEEEpeerreviewmaketitle

\section*{Acknowledgment}
This work was supported by the Key Research and Development Program of Hubei Province (No. 2023BAB173), funded by the Chinese National Natural Science Foundation Projects (No. 41901265), a Major Program of the National Natural Science Foundation of China (No. 92038301), and was supported in part by the Special Fund of Hubei Luojia Laboratory (No. 220100028).

\ifCLASSOPTIONcaptionsoff
  \newpage
\fi

\bibliographystyle{IEEEtran}
\bibliography{refer}
\end{document}